\documentclass[acmsmall]{acmart}
\AtBeginDocument{%
  \providecommand\BibTeX{{%
    \normalfont B\kern-0.5em{\scshape i\kern-0.25em b}\kern-0.8em\TeX}}}

\setcopyright{acmlicensed}
\copyrightyear{2024}
\acmYear{2024}
\acmDOI{XXXXXXX.XXXXXXX}

\usepackage{graphicx}
\usepackage{graphics}
\usepackage{multirow}%
\usepackage{amsthm}%
\usepackage{mathrsfs}%
\usepackage[title]{appendix}%
\usepackage{xcolor}%
\usepackage{textcomp}%
\usepackage{manyfoot}%
\usepackage{booktabs}%
\usepackage{algorithm}%
\usepackage{algorithmicx}%
\usepackage{algpseudocode}%
\usepackage{listings}%
\usepackage{multirow}
\usepackage{makecell}

\acmJournal{TIST}

\usepackage{makecell}
\usepackage{enumitem}
\usepackage{subcaption}
\usepackage{comment}
\usepackage{xspace}
\usepackage{diagbox}
\usepackage{rotating}
\usepackage{xcolor}%
\colorlet{RED}{red}
\colorlet{BLUE}{blue}
\colorlet{BLACK}{black}
\colorlet{PURPLE}{purple}

\newcommand{\smallsection}[1]{{\vspace{0.02in} \noindent {\bf{\underline{\smash{#1}}}}}}
\newcommand\red[1]{\textcolor{black}{#1}}

\newcommand{\numdata}{24\xspace}
\newcommand{\numscean}{35\xspace}
\newcommand{\numalgo}{12\xspace}

\newcommand{\uls}[1]{\underline{\smash{#1}}}

\newcommand{\ST}{\mathcal{T}}

\newcommand{\SG}{\mathcal{G}}
\newcommand{\SV}{\mathcal{V}}
\newcommand{\SE}{\mathcal{E}}
\newcommand{\SX}{\mathcal{X}}
\newcommand{\SC}{\mathcal{C}}

\newcommand{\STi}{\mathcal{T}_{i}}

\newcommand{\SGi}{\mathcal{G}_{i}}
\newcommand{\SVi}{\mathcal{V}_{i}}

\newcommand{\SEi}{\mathcal{E}_{i}}
\newcommand{\SXi}{\mathcal{X}_{i}}

\newcommand{\SCi}{\mathcal{C}_{i}}

\newcommand{\SSi}{\mathcal{S}_{i}}
\newcommand{\SSj}{\mathcal{S}_{j}}

\newcommand{\STj}{\mathcal{T}_{j}}
\newcommand{\STk}{\mathcal{T}_{k}}

\newcommand{\SGj}{\mathcal{G}_{j}}
\newcommand{\SCj}{\mathcal{C}_{j}}

\newcommand{\SQ}{\mathcal{Q}}

\newcommand{\bSVi}{\mathcal{V}'_{i}}
\newcommand{\bSEi}{\mathcal{E}'_{i}}

\newcommand{\MM}{\mathrm{\mathbf{M}}}

\newcommand{\cora}{$\mathsf{Cora}$\xspace}
\newcommand{\citeseer}{$\mathsf{Citeseer}$\xspace}
\newcommand{\arxiv}{$\mathsf{ogbn}$-$\mathsf{arxiv}$\xspace}

\newcommand{\proteins}{$\mathsf{ogbn}$-$\mathsf{proteins}$\xspace}
\newcommand{\mnist}{$\mathsf{MNIST}$\xspace}
\newcommand{\cifar}{$\mathsf{CIFAR10}$\xspace}
\newcommand{\aroma}{$\mathsf{Aromaticity}$\xspace}
\newcommand{\magdata}{$\mathsf{ogbn}$-$\mathsf{mag}$\xspace}
\newcommand{\twitch}{$\mathsf{Twitch}$\xspace}
\newcommand{\amazonphoto}{$\mathsf{Amazon}$-$\mathsf{photo}$\xspace}
\newcommand{\elliptic}{$\mathsf{Elliptic}$\xspace}
\newcommand{\paperm}{$\mathsf{Paper100M}$\xspace}

\newcommand{\corafull}{$\mathsf{CoraFull}$\xspace}
\newcommand{\products}{$\mathsf{ogbn}$-$\mathsf{products}$\xspace}
\newcommand{\facebook}{$\mathsf{Facebook}$\xspace}
\newcommand{\askubuntu}{$\mathsf{Ask}$-$\mathsf{Ubuntu}$\xspace}

\newcommand{\collab}{$\mathsf{ogbl}$-$\mathsf{collab}$\xspace}
\newcommand{\molhiv}{$\mathsf{ogbg}$-$\mathsf{molhiv}$\xspace}
\newcommand{\wikics}{$\mathsf{Wiki}$-$\mathsf{CS}$\xspace}
\newcommand{\nyctaxi}{$\mathsf{NYC}$-$\mathsf{Taxi}$\xspace}
\newcommand{\bitcoin}{$\mathsf{Bitcoin}$\xspace}

\newcommand{\ppa}{$\mathsf{ogbg}$-$\mathsf{ppa}$\xspace}
\newcommand{\sentiment}{$\mathsf{Sentiment140}$\xspace}

\newcommand{\actor}{$\mathsf{Actor}$\xspace}
\newcommand{\pubmed}{$\mathsf{Pubmed}$\xspace}
\newcommand{\dblp}{$\mathsf{DBLP}$\xspace}
\newcommand{\reddit}{$\mathsf{Reddit}$\xspace}
\newcommand{\pharmabio}{$\mathsf{PharmaBio}$\xspace}
\newcommand{\articles}{$\mathsf{Articles}$\xspace}
\newcommand{\ppi}{$\mathsf{PPI}$\xspace}
\newcommand{\amazon}{$\mathsf{Amazon}$\xspace}
\newcommand{\amazoncomputer}{$\mathsf{Amazon}$-$\mathsf{computer}$\xspace}
\newcommand{\amazonclothing}{$\mathsf{Amazon}$-$\mathsf{clothing}$\xspace}
\newcommand{\aromaticity}{$\mathsf{Aromaticity}$\xspace}
\newcommand{\sider}{$\mathsf{SIDER}$\xspace}
\newcommand{\tox}{$\mathsf{Tox21}$\xspace}
\newcommand{\politifact}{$\mathsf{PolitiFact}$\xspace}
\newcommand{\gossipcop}{$\mathsf{GossipCop}$\xspace}

\newcommand{\zinc}{$\mathsf{ZINC}$\xspace}
\newcommand{\movie}{$\mathsf{MovieLens}$\xspace}
\newcommand{\gowalla}{$\mathsf{Gowalla}$\xspace}
\newcommand{\aqsol}{$\mathsf{AqSol}$\xspace}

\newcommand{\ourfw}{\textsc{BeGin}\xspace}
\newcommand{\method}{\ourfw}

\newcommand{\checkimg}{$\includegraphics[scale=0.02]{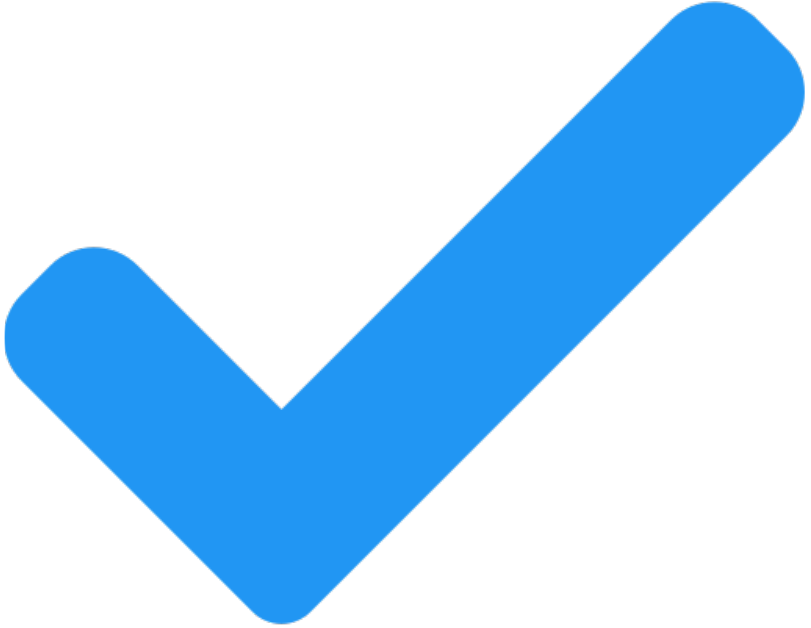}$}
\newcommand{\ximg}{$\includegraphics[scale=0.02]{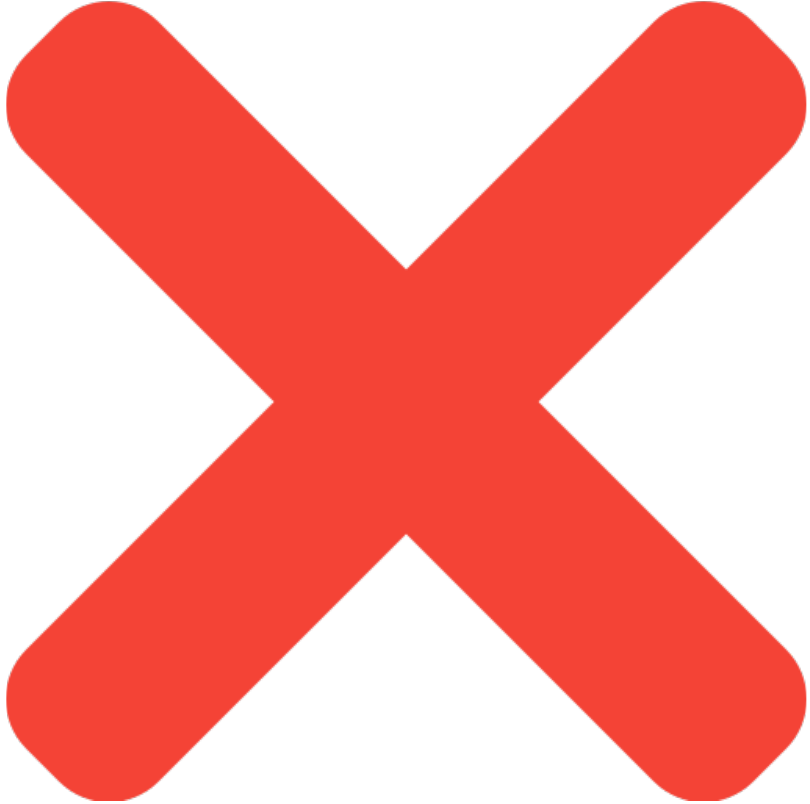}$}
\newcommand{\bbf}[1]{\scalebox{0.9}[1.0]{\textbf{#1}}}
\definecolor{ForestGreen}{RGB}{34,139,34}

\begin{document}
\title{BeGin: Extensive Benchmark Scenarios and An Easy-to-use Framework for Graph Continual Learning}

\author{Jihoon Ko}
\authornote{Both authors contributed equally to this research.}
\email{jihoonko@kaist.ac.kr}
\author{Shinhwan Kang}
\authornotemark[1]
\email{shinhwan.kang@kaist.ac.kr}
\author{Taehyung Kwon}
\email{taehyung.kwon@kaist.ac.kr}
\author{Heechan Moon}
\email{heechan9801@kaist.ac.kr}
\author{Kijung Shin}
\authornote{Corresponding author.}
\affiliation{%
  \institution{Kim Jaechul Graduate School of AI, KAIST}
  \city{Seoul}
  \country{South Korea}
}

\renewcommand{\shortauthors}{Ko, Kang, et al.}

\begin{abstract}
    Continual Learning (CL) is the process of learning ceaselessly a sequence of tasks.
Most existing CL methods deal with independent data (e.g., images and text) for which many benchmark frameworks and results under standard experimental settings are available.
Compared to them, however, CL methods for graph data (graph CL) are relatively underexplored because of (a) the lack of standard experimental settings, especially regarding how to deal with the
dependency between instances, (b) the lack of benchmark datasets and scenarios, and (c) high complexity in implementation and evaluation due to the dependency.
In this paper, regarding (a) we define four standard incremental settings (task-, class-, domain-, and time-incremental) for node-, link-, and graph-level problems, extending the previously explored scope. 
Regarding (b), we provide \numscean benchmark scenarios based on \numdata real-world graphs.
Regarding (c), we develop \method, an easy and fool-proof framework for graph CL. \method is easily extended since it is modularized with reusable modules for data processing, algorithm design, and evaluation.
Especially, the evaluation module is completely separated from user code to eliminate potential mistakes.
Regarding benchmark results, we cover $3\times$ more combinations of incremental settings and levels of problems than the latest benchmark. All assets for the benchmark framework are publicly available at \url{https://github.com/ShinhwanKang/BeGin}.
\end{abstract}




\maketitle

\begin{figure}[ht]
    \centering
    \includegraphics[width=0.97\linewidth]{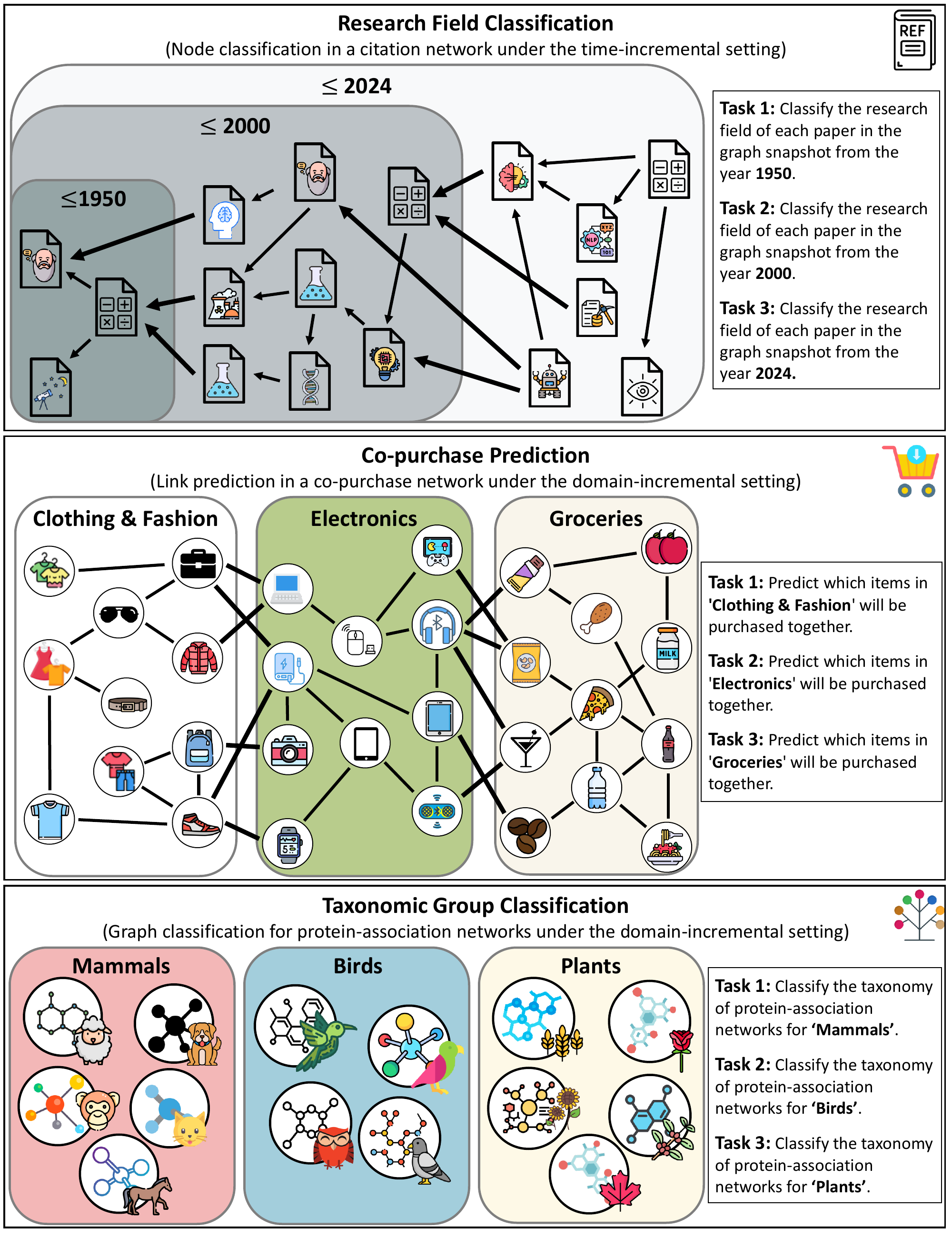}
    \caption{\textbf{Examples of graph continual learning problems.}}
    \label{fig:int:examples}
\end{figure}

\begin{figure}[ht]
\vspace{-2mm}
    \centering
    \begin{subfigure}[b]{0.27\textwidth}
         \centering
         \includegraphics[width=\linewidth]{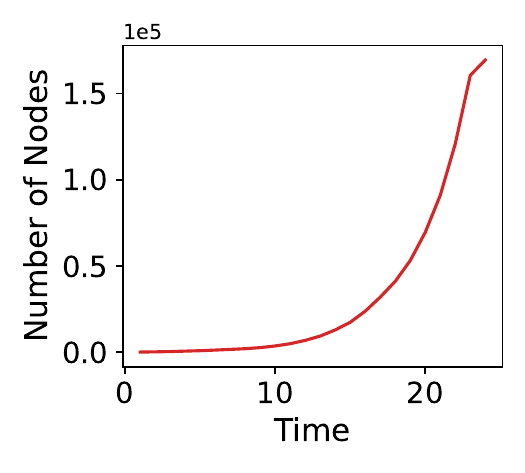} \\
         \vspace{-3mm}
         \caption{Number of Nodes}
         \label{fig:stat_numnode}
    \end{subfigure}
    \begin{subfigure}[b]{0.27\textwidth}
         \centering
         \includegraphics[width=\linewidth]{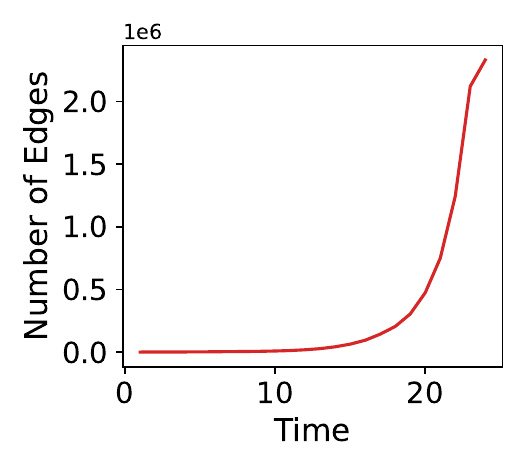} \\
         \vspace{-3mm}
         \caption{Number of Edges}
         \label{fig:stat_numedge}
    \end{subfigure}
    \begin{subfigure}[b]{0.27\textwidth}
         \centering
         \includegraphics[width=\linewidth]{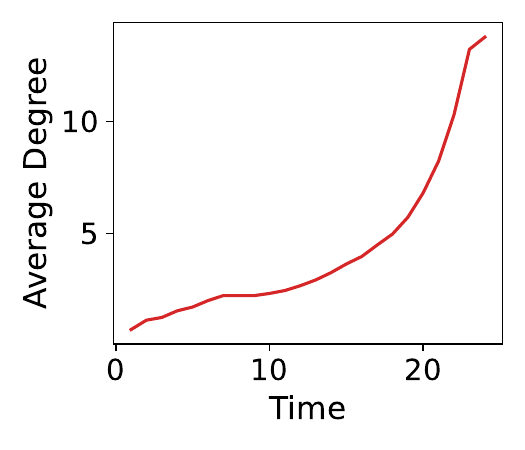} \\
         \vspace{-3mm}
         \caption{Average Degree}
         \label{fig:stat_degree}
    \end{subfigure}
    \begin{subfigure}[b]{0.27\textwidth}
         \centering
         \includegraphics[width=\linewidth]{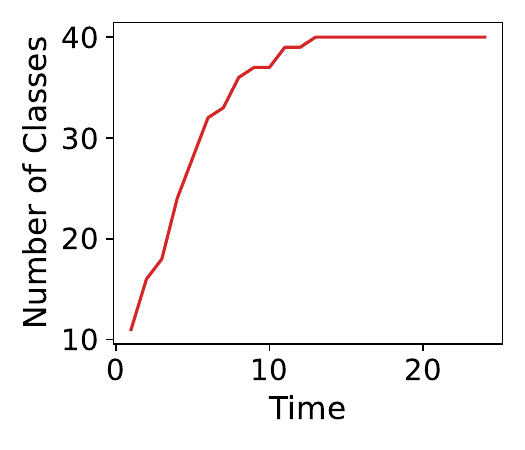} \\
         \vspace{-3mm}
         \caption{Number of Classes}
         \label{fig:stat_numclass}
    \end{subfigure}
    \begin{subfigure}[b]{0.27\textwidth}
         \centering
         \includegraphics[width=\linewidth]{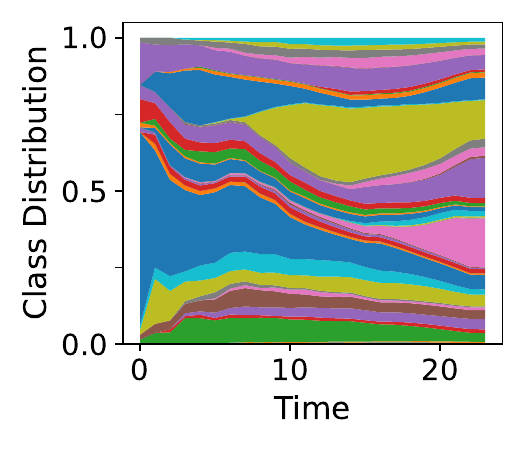} \\
         \vspace{-3mm}
         \caption{Shift in Class Distributions}
         \label{fig:stat_classdist}
    \end{subfigure}
    \begin{subfigure}[b]{0.27\textwidth}
         \centering
         \includegraphics[width=\linewidth]{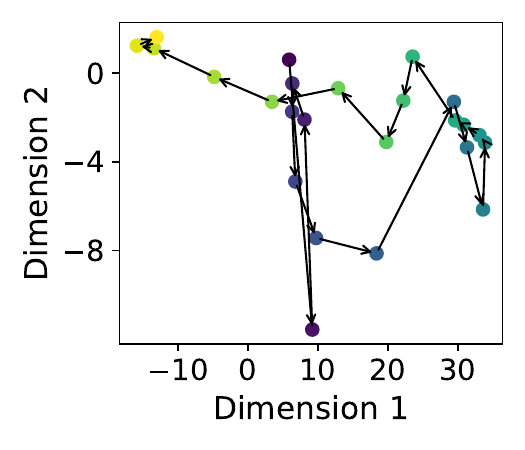} \\
         \vspace{-3mm}
         \caption{Shift in Average Features}
         \label{fig:stat_avgfeat}
    \end{subfigure}
    \caption{\textbf{Motivating data analysis: temporal dynamics in real-world graphs.} In the \arxiv dataset (see Section~\ref{sec:scenarios:examples} for details), we observe a gradual increase over time in four aspects (a)-(d). 
    Additionally, the class distribution varies over time, accompanied by the emergence of new classes, as depicted in (e). 
    Moreover, in (f), where t-SNE~\citep{van2008visualizing} is used for dimensionality reduction, the average node features exhibit shifts over time, as indicated by the directional arrows.}
    \label{fig:int:increasing}
\end{figure}

\section{Introduction}
\label{sec:intro}

Continual Learning (CL), which is also known as lifelong learning and incremental learning, is the process of learning continuously a sequence of tasks.
CL aims to retain knowledge from previous tasks (i.e., knowledge consolidation) to overcome the degradation of performance (i.e., catastrophic forgetting) on previous tasks.
Recently, CL has received considerable attention because of its similarity to the development of human intelligence. 

Most of the existing works for CL deal with independent data, such as images and text.
For example, Shin et al. \cite{shin2017continual} aim to learn a sequence of image-classification tasks where the domains of images vary with tasks. 
Ermis et al. \cite{ermis2022memory} aim to learn a sequence of text-classification tasks where the possible classes of text grow over tasks.
For CL with independent data, many datasets \citep{goodfellow2013empirical,hsu2018re,lomonaco2017core50} and benchmarks \citep{he2021clear,lin2021clear,lomonaco2021avalanche,pham2021dualnet,shin2017continual} have been provided.

CL naturally provides significant benefits for graph data in real-world applications since the data grows in size and diversity across various domains accompanied by the emergence of new tasks, as illustrated in Figures~\ref{fig:int:examples} and \ref{fig:int:increasing}.
The majority of existing graph representation learning methods, however, are designed for fixed tasks, assuming that the input graph remains unchanged~\citep{kipf2016semi,velivckovic2017graph}. While a number of methods exist to handle graph evolution over time \citep{xu2020inductive,rossi2020temporal}, the ability to incorporate new tasks, classes, and/or domains is still relatively rare \citep{wang2020streaming,zhou2021overcoming}.

Despite its necessity, graph CL \citep{febrinanto2022graph,zhou2021overcoming} has been relatively underexplored compared to CL with independent data, mainly due to the complexity caused by the dependency between instances.
For example, in node classification, the class of a node is correlated not only with its features but also with its connections to other nodes, which may belong to other tasks, and their features.
Major reasons for the relative unpopularity of graph CL include the scarcity of standard experimental settings. 
For example, how to deal with changes in various dimensions (e.g., domains and classes) and the dependency between instances needs to be standardized. Moreover, the relative scarcity of benchmarking datasets and scenarios also limits its popularity.

In this paper, we focus on resolving these issues. 
Our first contribution is to define four incremental settings for graph data, enriching those established in general CL but partially explored for Graph CL.
To this end, we identify and decouple four possible dimensions of change, which are task, class, domain, and time.
It is worth noting that previous studies often couple these dimensions of change complicatedly without sufficient justification~\citep{zhang2022cglb,zhou2021overcoming} (see Section~\ref{sec:scenarios:settings} for details). Thus, there has been a lack of opportunities to consider each dimension on its own for effective algorithm design, evaluation, and other purposes.
Each of the settings is defined so that the dependency between instances, which is a unique property of graph data, can be utilized. Simultaneously, each setting may lead to catastrophic forgetting without careful consideration of it.
Then, we show that the settings can be applied to node-, link-, and graph-level problems, including node classification, link prediction, and graph classification.
After that, we provide \numscean benchmark scenarios for graphs from \numdata real-world datasets, which cover $12$ combinations of the incremental settings and the levels of problems, as summarized in Table~\ref{tab:coverage}. 

\begin{table*}[t]
\centering
\caption{\label{tab:coverage} \textbf{Comparison of graph CL benchmarks.} Our \ourfw framework includes more incremental settings, levels of problems, graph CL methods, and evaluation metrics, compared to the others.} 
\setlength{\tabcolsep}{5pt}
\renewcommand{\arraystretch}{1.1}
\scalebox{0.8}{
\begin{tabular}{c|cc|ccc|ccc|ccc}
\toprule
\multicolumn{3}{c|}{}  & \multicolumn{3}{c|}{\method (Proposed)} & \multicolumn{3}{c|}{\citeauthor{zhang2022cglb} \citep{zhang2022cglb}} & \multicolumn{3}{c}{\citeauthor{carta2021catastrophic} \citep{carta2021catastrophic}} \\
\midrule 
\multicolumn{1}{c|}{\multirow{5}{*}{Scenarios}} & \multicolumn{2}{c|}{Problem Level} & Node & Link & Graph & Node & Link & Graph & Node & Link & Graph \\ 
\cmidrule(lr){2-12}
 & \multicolumn{1}{c|}{\multirow{4}{*}{\rotatebox[origin=c]{90}{\begin{tabular}[c]{@{}c@{}}Incremental\\ Setting\end{tabular}}}}
 & Task & \checkimg & \checkimg & \checkimg & \checkimg & \ximg & \checkimg & \ximg & \ximg & \ximg \\
 & \multicolumn{1}{c|}{} & Class & \checkimg & \checkimg & \checkimg & \checkimg & \ximg & \checkimg & \ximg & \ximg & \checkimg \\
 & \multicolumn{1}{c|}{} & Domain & \checkimg & \checkimg & \checkimg & \ximg & \ximg & \ximg & \ximg & \ximg & \ximg \\
 & \multicolumn{1}{c|}{} & Time & \checkimg & \checkimg & \checkimg & \ximg & \ximg & \ximg & \ximg & \ximg & \ximg \\ 
\midrule
\multicolumn{3}{c|}{\# of Graph CL Methods}  & \multicolumn{3}{c|}{\textbf{\numalgo}} & \multicolumn{3}{c|}{6} & \multicolumn{3}{c}{3} \\
\midrule
\multicolumn{3}{c|}{\# of Evaluation Metrics}  & \multicolumn{3}{c|}{\textbf{4}} & \multicolumn{3}{c|}{2} & \multicolumn{3}{c}{1} \\
\bottomrule
\end{tabular}
}
\end{table*}

Our second contribution is \ourfw (\textbf{Be}nchmarking \textbf{G}raph Cont\textbf{i}\textbf{n}ual Learning), which is a framework for implementation and evaluation of graph CL methods.
Evaluation is complicated for graph CL due to additional dependency between instances.
Specifically, it is common to utilize instances of one task for other tasks in order to leverage their dependencies, and thus without careful consideration, information that should not be shared among tasks can be leaked among tasks.
In order to eliminate potential mistakes in evaluation,
\ourfw is \textbf{fool-proof} by completely separating the evaluation module from the learning part, where users implement their own graph CL methods.
The learning part only has to answer queries provided by the evaluation module after each task is processed.  
\ourfw is \textbf{easy-to-use}, and especially it is easily extended since it is modularized with reusable modules for data processing, algorithm design, training, and evaluation.

Our last contribution is to provide extensive benchmark results of \numalgo graph CL methods, in terms of four evaluation metrics, based on our proposed scenarios and the framework. 
Refer to Table~\ref{tab:coverage} for comparison with existing ones.

For \textbf{reproducibility}, we provide all source code required for reproducing the benchmark results and documents for users at \url{https://github.com/ShinhwanKang/BeGin}.

\section{Related Works}
\label{sec:related}

\smallsection{Continual Learning with Independent Data.}
Continual learning (CL) methods have been developed mostly for independent data (e.g., images and text), being largely categorized as follows:
\begin{itemize}[leftmargin=*]
    \item \textbf{Replay-based methods}~\citep{chaudhry2019tiny,rebuffi2017icarl,lopez2017gradient}: Replay-based methods store a sample of data for previous tasks. Then, they re-use the data while learning a new task to mitigate forgetting.  For example, Gradient Episodic Memory (GEM) \citep{lopez2017gradient} stores data from previous tasks and prevents the increase of losses on them while learning a new task.
    \item \textbf{Regularization-based methods}~\citep{kirkpatrick2017overcoming,aljundi2018memory,li2017learning}: Regularization-based methods seek to consolidate knowledge from previous tasks by introducing regularization terms in the loss function. Elastic Weight Consolidation (EWC) \citep{kirkpatrick2017overcoming} weights to parameters according to the diagonal of the Fisher information matrix, and 
    Memory Aware Synapses (MAS) \citep{aljundi2018memory} computes the importance of parameters according to how sensitive the parameters are. 
    Learning without Forgetting (LwF) \citep{li2017learning} minimizes the difference between the outputs of a previous model and a new model. 
    \item \textbf{Parameter-isolation-based methods}~\citep{mallya2018packnet,mallya2018piggyback,serra2018overcoming}: Parameter-isolation-based methods learn binary or real-valued masks for weighting parameters or model outputs for each task. For instance, PackNet~\citep{mallya2018packnet} generates binary masks based on the magnitude of learned parameters and re-trains only unmasked ones for new tasks. Piggyback~\citep{mallya2018piggyback} learns real-valued masks and binarizes them to mask parameters for each task. HAT~\citep{serra2018overcoming} learns real-valued masks, which play a role similar to attention modules, for weighting each layer output for each task.
\end{itemize}
Moreover, there exist several frameworks \citep{lomonaco2021avalanche,lin2021clear} designed for implementing and evaluating CL methods for independent data.
However, none of them currently supports CL with graph data.

\smallsection{Continual Learning with Graph Data (Graph CL).}
Due to their expressiveness, graphs are widely used to model various types of data, and considerable attention has been paid to machine learning with graph-structured data.
Since many such graphs (e.g., online social networks) evolve over time, continual learning is naturally desirable for them, and thus several graph CL methods have been developed \citep{febrinanto2022graph,galke2021lifelong, sun2023self}. They are mainly categorized as follows:
\begin{itemize}[leftmargin=*]
    \item \textbf{Replay-based methods}~\citep{zhou2021overcoming,wang2022lifelong}: For example, ER-GNN \citep{zhou2021overcoming} samples nodes and uses them for re-training.
FGN~\citep{wang2022lifelong} transforms each original node into a form of a feature graph, which takes original features as nodes, and then applies a replay-based method to them \citep{aljundi2019gradient}.
CaT~\citep{liu2023cat} utilizes graph condensation (i.e., graph dataset distillation) techniques to effectively capture historical distributions with limited resources and address distributional imbalances across tasks, such as variations in the number of nodes.
Specifically, for each task, CaT first generates a condensed graph and then performs learning on this condensed graph, together with the condensed graphs from previous tasks, rather than learning directly from the original, non-condensed graphs.
    \item \textbf{Regularization-based methods}~\citep{liu2021overcoming, cai2022multimodal}: For example, TWP \citep{liu2021overcoming} stabilizes parameters important in topological aggregation by graph neural networks through regularization.
    \item \textbf{Parameter-isolation-based methods~\citep{zhang2023continual}}: For example, for each new task, PI-GNN~\citep{zhang2023continual} expands a graph neural network with new parameters that are trained for the current task while it freezes existing parameters to maintain performance on previous tasks.
    \item \textbf{Hybrid-based methods}~\citep{daruna2021continual, wang2020streaming}: 
    For example, CGNN \citep{wang2020streaming} combines replay- and regularization-based approaches.
\end{itemize}
Apart from these primary categories, HPNs~\citep{zhang2022hierarchical} retain and expand abstract knowledge of different levels in the form of prototypes (i.e., embeddings). Given a node, the prototypes relevant to the node are retrieved and used as input for prediction.

Despite these efforts, graph CL is still largely underexplored, especially compared to CL with independent data, and the lack of benchmark frameworks and scenarios is one major reason.
To the best of our knowledge, there exist only two benchmarks for graph CL \citep{carta2021catastrophic,zhang2022cglb}.
However, \cite{carta2021catastrophic} supports only graph-level tasks under one incremental setting, and \cite{zhang2022cglb} supports node- and graph-level tasks under only two settings. 
Compared to them, our benchmark and framework are more extensive, as summarized in Table~\ref{tab:coverage}.

\smallsection{Comparison with Other Graph Learning Methods.}
Methods have been developed to handle shifts in data distribution between training and test datasets~\citep{li2022ood,wu2021towards, ko2021learning,baek2020learning}.
They, however, differ from CL methods, which typically focus on managing data distributional shifts across different tasks.
Moreover, transfer learning (TL) on graph data has been studied for various downstream tasks (e.g., link prediction \citep{tang2016transfer}). While CL aims to train a single model for a sequence of tasks, TL aims to adapt a separate model to a new task. 
In addition, many works on incremental/dynamic graph learning \citep{rossi2020temporal,you2022roland} focus on the latest snapshot of a dynamic graph to maximize performance on it.
However, graph CL 
aims to preserve performance not only on the current snapshot but also on past ones, which can resemble future snapshots due to seasonality, etc.

\section{Benchmark Scenarios}

\label{sec:scenarios}

We introduce \numscean benchmark scenarios with $5$ graph \textbf{problems}, $4$ incremental \textbf{settings}, and \numdata real-world \textbf{datasets}.

\smallsection{Common Notations.} 
We denote each $i$-th task in a sequence by $\ST_i$
and use $\SG=(\SV, \SE, \SX)$ to denote a graph that consists of a set of nodes $\SV$, a set of edges $\SE$, and node features $\SX: \SV \rightarrow \mathbb{R}^{d}$, where $d$ is the number of node features. In some of the considered datasets, edge features are given in addition to or instead of node features, and they can be treated similarly to node features. 
Lastly, we use $\SQ$ to indicate the set of queries used for evaluation.

\subsection{Graph Learning Problems of Three Levels}

\label{sec:scenarios:probs}

Our benchmark scenarios are based on various node-, link-, and graph-level problems.
Below, we describe node classification, link prediction, and graph classification, as examples.

\smallsection{Node Classification (NC).} 
For a node classification (NC) task $\ST_i$, the input consists of (a) a graph $\SGi=(\SVi,\SEi,\SXi)$, (b) a labeled set $\bSVi\in \SVi$ of nodes, (c) a set of classes $\SCi$, and (d) the class $f(v)\in \SCi$ for each node $v\in \bSVi$.
A query $q$ on a NC task $\ST_i$ is a node $v_{q}\notin \bSVi$ where $f(v_{q})\in \SCi$, and its ground-truth answer is $f(v_q)$.

\smallsection{Link Prediction (LP).} 
For a link prediction (LP) task $\ST_i$, the input consists of a graph $\SGi=(\SVi,\SEi\setminus\bSEi,\SXi)$, where $\SEi$ is the ground-truth set of edges and $\bSEi\in \SEi$ is the set of missing edges among them.
A query on an LP task $\ST_i$ is a node pair $(u,v) \notin (\SEi\setminus\bSEi)$, and its ground-truth answer is $\textbf{1}((u,v)\in \bSEi)$, i.e., whether there exists a missing edge between $u$ and $v$ or not.

\smallsection{Graph Classification (GC).} 
For a graph classification (GC) task $\ST_i$, the input includes (a) a labeled set of graphs $\SSi$, (b) a set of classes $\SCi$, and (c) the class $f(\SG)\in \SCi$ for each graph $\SG\in \SSi$.
A query $q$ on a GC task $\ST_i$ is a graph $\SG_{q}\notin \SSi$ where $f(\SG_{q})\in \SCi$, and its ground-truth  answer is $f(\SG_{q})$.

\smallsection{Graph Regression (GR).} 
For a graph regression (GR) task $\ST_i$, the input includes (a) a set of graphs $\SSi$ and (b) a real value $f(\SG)\in \mathbb{R}$ assigned to each graph $\SG\in \SSi$.
A query $q$ on a GR task $\ST_i$ is a graph $\SG_{q}\notin \SSi$, and its ground-truth answer is $f(\SG_{q})$.

The problem definition of \textbf{Link Classification (LC)} problem, which is also used for our benchmark, is extended straightforwardly from that of node classification.

\subsection{Four Incremental Settings}

\label{sec:scenarios:settings}

We introduce four incremental settings for graph CL and describe how they can be applied to the above three problems. 
When designing them, we aim to decouple changes in different dimensions (tasks, classes, domains, and time) if it does not reflect actual graph dynamics. Moreover, we aim to make the dependency between instances (e.g., connections between nodes) exploitable. Lastly, we make the input for a task (partially) lost in later tasks so that catastrophic forgetting may happen without careful attention to it.

\smallsection{Task-Incremental (Task-IL).}
The set of classes varies with tasks (i.e., $\forall i\neq j, \SCi\neq \SCj$), and they are often disjoint (i.e., $\forall i\neq j, \SCi\cap \SCj=\emptyset$).
In addition, for each query in $\SQ$, the corresponding task, which we denote by $\ST_i$, is provided, and thus its answer is predicted among $\SCi$.

\smallsection{Class-Incremental (Class-IL).}
The set of classes grows over tasks (i.e., $\forall i<j, \SCi \subsetneq \SCj$). In addition, for each query in $\SQ$, the corresponding task is \textbf{NOT} provided, and thus its answer is predicted among all classes seen so far (i.e., $\bigcup_{j\leq i}\SCj=\SCi$ for a current task $\ST_i$).

\smallsection{Domain-Incremental (Domain-IL).}
We divided entities (i.e., nodes, edges, and graphs) over tasks according to their domains, which are additionally given (see Section~\ref{sec:scenarios:examples} for real-world examples of domains).
Note that, as domains, we use information not directly related to labels, which we aim to infer.
The details for each problem are as follows:

\begin{itemize}[leftmargin=*]
    \item NC \& LC: The labeled nodes (or edges) of the input graph are divided into tasks according to their domains. The input graph is fixed (i.e., $\forall i\neq j$,  $\SGi=\SGj$) for all tasks.
    \item LP: The ground-truth edges are partitioned into (a) the set $\bar{\SE}$ of base edges and (b) the set $\tilde{\SE}$ of additional edges.
The base edges are provided commonly for all tasks, and they are especially useful when answering queries on past tasks.\footnote{Without base edges, past-task queries need to be answered restrictively using only other-domain edges.} 
The additional edges $\tilde{\SE}$ are provided further for each task according to their domains.
For each task $\ST_i$, $\bar{\SE}\cup \tilde{\SE}_i$, where $\tilde{\SE}_i$ is the additional edges assigned to $\ST_i$, is 
used as the ground-truth edges (i.e., $\SEi=\bar{\SE}\cup \tilde{\SE}_i$),
and missing ground-truth edges are chosen among $\tilde{\SE}_i$, i.e., $\bSEi\subset  \tilde{\SE}_i$.
These edge sets $\SEi$ and $\bSEi$ are used 
for each LP task $\ST_i$,
as described in Section~\ref{sec:scenarios:probs}.
    \item GC \& GR: 
    The labeled graphs are divided into GC or GR tasks according to their domains.
\end{itemize}

\smallsection{Time-Incremental (Time-IL).}
We consider a dynamic graph evolving over time.
We denote its $i$-th snapshot by $\SG^{(i)}=(\SV^{(i)}, \SE^{(i)}, \SX^{(i)})$. 
The class set may or may not vary with tasks, and the details of each problem are as follows:

\begin{itemize}[leftmargin=*]
    \item NC \& LC: The input graph for each task $\ST_i$ is the $i$-th snapshot $\SG^{(i)}$ of the dynamic graph, and labeled nodes (or edges) are given among new nodes (or edges) added to the snapshot (i.e.,  $\bSVi\subset \SV^{(i)}\setminus\SV^{(i-1)}$, where $\SV^{(0)}=\emptyset$). 
    The label and features of each node (or edge) remain constant over time, while its connections may vary.
    \item LP: 
    As in the Domain-IL setting, base edges are used.
    For each task $\ST_i$, the set $\bar{\SE}_i$ of base edges so far and the new edges added to the $i$-th snapshot $\SG^{(i)}$ of the dynamic graph are used as the ground-truth edges (i.e., $\SEi = \bar{\SE}_i \cup (\SE^{(i)}\setminus\SE^{(i-1)})$, where $\SE^{(0)}=\emptyset$).
    After each task is processed, a subset of  $\SE^{(i)}\setminus\SE^{(i-1)}\setminus\bSEi$ (i.e., new edges that are not used as missing edges) are added as base edges.
    The features of each node are fixed over time and thus for all tasks.
    \item GC: 
    The snapshots of the dynamic graph are grouped and assigned to tasks in chronological order. 
    Specifically, for any $i$ and $j$ where $i<j$, every snapshot in the labeled set $\SSi$ of the task $\ST_i$ is older than every snapshot in $\SSj$ of $\ST_j$.
    The features of nodes may change over time. 
\end{itemize}

\smallsection{Remarks on Domain-IL and Time-IL.}
While time can be considered as a domain, we deliberately distinguish the Time-IL and Domain-IL settings. Many real-world graphs undergo temporal evolution, establishing an intrinsic connection between time and graph dynamics (see Figure~\ref{fig:int:increasing} for an example). However, such a relationship is not apparent for (non-temporal) domains. We assume that graphs evolve under time-IL settings but not under Domain-IL settings.

\smallsection{Remarks on Difference from \citep{zhang2022cglb}.}
For NC and LC, we fix the input graph for all tasks (i.e., $\forall i\neq j$, $\SGi=\SGj$)  enabling us to study the changes in each dimension (i.e., task, class, and domain) on its own, and we separately consider the dynamics of the input graph in Time-IL settings.
It is important to note that the Task-IL and Class-IL settings in \citep{zhang2022cglb} differ from our approach. In their settings, the timing of when edges are added to the input graph is determined solely by the tasks (or classes) that the endpoints belong to.
That is, changes in tasks (or classes) and the dynamics of the input graph are perfectly coupled without sufficient justification. This coupling does not reflect actual graph dynamics, which are not solely or predominantly determined by tasks.

\begin{table*}[t]
    \caption{\textbf{Summary of the considered real-world datasets and benchmark scenarios.}}
    \label{tab:datasets}
    \centering
    \scalebox{0.53}{
    \begin{tabular}{c|c|c|c|c|c|c|c|c|c}
        \toprule
        Problem & Dataset & \# Nodes & \# Edges & \makecell{\# Node (Edge) \\ Features} & \makecell{Incremental \\ Setting} & \# Tasks & Task Split Type & \makecell{Train/Val/Test \\ Split Type} & \makecell{Performance \\ Metric} \\
        \midrule
        & \multirow{2}{*}{\cora} & \multirow{2}{*}{2,708} & \multirow{2}{*}{10,556} & \multirow{2}{*}{1,433 (0)} & \multirow{2}{*}{Task-IL, Class-IL} & \multirow{2}{*}{3} & Class label & \multirow{2}{*}{Original Split} & \\
         & & & & & & & (2 classes per task) & & \\
         \cmidrule{2-9}
          & \multirow{2}{*}{\citeseer} & \multirow{2}{*}{3,327} & \multirow{2}{*}{9,104} & \multirow{2}{*}{3,703 (0)} &  \multirow{2}{*}{Task-IL, Class-IL} & \multirow{2}{*}{3} & Class label & \multirow{2}{*}{Original Split} & \\
         & & & & & & & (2 classes per task) & & \\
         \cmidrule{2-9}
          & \multirow{3}{*}{\arxiv} & \multirow{3}{*}{19,793} & \multirow{3}{*}{126,842} & \multirow{3}{*}{8,710 (0)} & \multirow{2}{*}{Task-IL, Class-IL} & \multirow{2}{*}{8} & Class label & \multirow{2}{*}{Original Split} & \\
         & & & & & & & (5 classes per task) & & \\
         \cmidrule{6-9}
          & &  &  &  & Time-IL & 24 & Time (Publication year) & Random (4:1:5) & Accuracy \\
         \cmidrule{2-9}
         \multirow{2}{*}{\makecell{Node \\Classification \\ (NC)}} & 
         \multirow{3}{*}{\magdata} & \multirow{3}{*}{736,389} &  \multirow{3}{*}{10,832,542} & \multirow{3}{*}{128 (0)} & \multirow{2}{*}{Task-IL, Class-IL} & \multirow{2}{*}{128} & Class label & \multirow{2}{*}{Original Split} & \\
         & & & & & & & (2 classes per task) & & \\ 
         \cmidrule{6-9}
         & &  &  &  & Time-IL & 10 & Time (Publication year) & Random (4:1:5) & \\
         \cmidrule{2-9}
          & \multirow{2}{*}{\corafull} & \multirow{2}{*}{19,793} & \multirow{2}{*}{126,842} & \multirow{2}{*}{8,710 (0)} & \multirow{2}{*}{Task-IL} & \multirow{2}{*}{35} & Class label & \multirow{2}{*}{Original Split} & \\
         & & & & & & & (2 classes per task) & & \\
         \cmidrule{2-9}
          & \multirow{2}{*}{\products} & \multirow{2}{*}{2,449,029} & \multirow{2}{*}{123,718,024} & \multirow{2}{*}{100 (0)} &  \multirow{2}{*}{Class-IL} & \multirow{2}{*}{9} & Class label & \multirow{2}{*}{Original Split} & \\
         & & & & & & & (5 classes per task) & & \\
         \cmidrule{2-10}
         & \proteins & 132,534 & 79,122,504 & 8 (0) & Domain-IL & 8 & Domain (Species) & Random (4:1:5) & ROC-AUC \\ 
         \cmidrule{2-10}
         & \twitch & 168,114 & 6,797,557 & 4 (0) & Domain-IL & 21 & Domain (Language) & Random (4:1:5) & Accuracy \\
         \midrule
         \multirow{3}{*}{\makecell{Link \\ Classification \\ (LC)}} & 
         \multirow{3}{*}{\bitcoin} & \multirow{3}{*}{5,858} &  \multirow{3}{*}{35,592} & \multirow{3}{*}{0 (0)} & \multirow{2}{*}{Task-IL, Class-IL} & \multirow{2}{*}{3} & Class label &  & \multirow{2}{*}{Accuracy} \\
         & & & & & & & (2 classes per task) & & \\
         \cmidrule{6-8}\cmidrule{10-10}
         & & & &  & Time-IL & 7 & Time & & ROC-AUC\\
         \cmidrule{1-8}\cmidrule{10-10}
          & \collab & 235,868 & 1,285,465 & 128 (0) & Time-IL & 50 & Time (Publication year) & & \\
         \cmidrule{2-8}
         
         & \wikics & 11,701 & 431,726 & 300 (0) & Domain-IL & 54 & Domain (labels of endpoints) & Random (8:1:1) & \multirow{3}{*}{Hits@50}  \\ 
         \cmidrule{2-8}
         \multirow{2}{*}{\makecell{Link \\ Prediction \\ (LP)}} & 
         \askubuntu & 159,313 & 507,988 & 1 (0) & Time-IL & 69 & Time (Interaction timestamp) & & \\
         \cmidrule{2-8}
         & \facebook & 134,833 & 1,380,293 & 1 (0) & Domain-IL & 8 & Domain (Categories of pages) & \\
         \cmidrule{2-8}\cmidrule{10-10}
         & \gowalla & 70,839 & 1,027,370 &2 (0) & Time-IL &10 & Time (Check-in timestamp) & & \multirow{2}{*}{Hits@100} \\
         \cmidrule{2-8}
         & \movie & 9,992 & 575,281 & 42 (0) & Time-IL & 10 & Time (Rating timestamp) & \\
         \midrule
         \midrule
         Problem & Dataset & \makecell{\# Graphs \\ (Avg. \# Nodes)} & \makecell{\# Avg.\\\# Edges} & \makecell{\# Node (Edge) \\ Features} & \makecell{Incremental \\ Setting} & \# Tasks & Task Split Type & \makecell{Train/Val/Test \\ Split Type} & \makecell{Performance \\ Metric} \\
         \midrule
          & \multirow{2}{*}{\mnist} & \multirow{2}{*}{55,000 (70.6)} & \multirow{2}{*}{564.5} & \multirow{2}{*}{3 (0)} & \multirow{2}{*}{Task-IL, Class-IL} & \multirow{2}{*}{5} & Class label & \multirow{2}{*}{Original Split} & \\
         & & & & & & & (2 classes per task) & & \\
         \cmidrule{2-9}
          & \multirow{2}{*}{\cifar} & \multirow{2}{*}{45,000 (117.6)} & \multirow{2}{*}{941.2} & \multirow{2}{*}{5 (0)} & \multirow{2}{*}{Task-IL, Class-IL} & \multirow{2}{*}{5} & Class label & \multirow{2}{*}{Original Split} & \\
          & & & & & & & (2 classes per task) & &  \\
         \cmidrule{2-9}
         \multirow{3}{*}{\makecell{Graph \\ Classification \\ (GC)}} & 
         \multirow{2}{*}{\aroma} & \multirow{2}{*}{3,868 (29.7)} & \multirow{2}{*}{65.4} & \multirow{2}{*}{2 (0)} & \multirow{2}{*}{Task-IL, Class-IL} & \multirow{2}{*}{10} & Class label & \multirow{2}{*}{Original Split} & \\
         & & & & & & & (3 classes per task) & & Accuracy \\
         \cmidrule{2-9}
         & \nyctaxi & 35,064 (265.0) & 2100.5 & 7 (1) & Time-IL & 16 & Time (Month) & \multirow{2}{*}{Random (6:2:2)} & \\
         \cmidrule{2-8}
         & \sentiment & 5,500 (13.43) & 23.71 & 300 (0) &  Time-IL & 11 & Time (Day) &  & \\
         \cmidrule{2-9}
         & \ppa & 40,700 (243.1) & 4603.4 & 2 (7) &  Domain-IL & 11 & Domain (Species) & Random (8:1:1) & \\
         \cmidrule{2-10}
         & \molhiv & 41,127 (25.5) & 27.5 & 9 (3) &  Domain-IL & 20 & Domain (Scaffold) & Random (8:1:1) & ROC-AUC \\ 
         \midrule
         \multirow{3}{*}{\makecell{Graph \\ Regression \\ (GR) }} & \multirow{2}{*}{\zinc} & \multirow{2}{*}{12,000 (23.16)} & \multirow{2}{*}{49.83} & \multirow{2}{*}{28 (4)} &  \multirow{2}{*}{Domain-IL} & \multirow{2}{*}{11} & \multirow{2}{*}{Domain (Molecule size)} & \multirow{2}{*}{Random (8:1:1)} & \multirow{3}{*}{\makecell{Mean \\ absolute \\ error (MAE)}} \\ 
         & & & & & & & & & \\ 
         \cmidrule{2-9}
         & \aqsol & 9,823 (17.57) & 35.76 & 65 (6) & Domain-IL & 5 & Domain (Scaffold) & Random (8:1:1) & \\ 
        \bottomrule
    \end{tabular}
    }
\end{table*}

\subsection{Real-world Datasets and Benchmark Scenarios}
\label{sec:scenarios:examples}

We describe \numdata real-world datasets and \numscean benchmark scenarios based on them under various incremental settings. We summarize the datasets and the scenarios in Table~\ref{tab:datasets}.

\subsubsection{Datasets for Node-Level Problems.}
\begin{itemize}[leftmargin=*]
\item \cora, \citeseer \citep{sen2008collective}, and \corafull \citep{bojchevski2017deep}
are citation networks. Each node is a scientific publication, and its class is the field of the publication.
For \cora and \citeseer,
we formulate three binary classification tasks for Task-IL and three tasks with $2$, $4$, and $6$ classes for Class-IL.
Similarly, for \corafull, we formulate $35$ binary classification tasks. 
Note that, one class is left unused in \cora.

\item \arxiv and \magdata \citep{hu2020open,wang2020microsoft} are citation networks, where 
each node is a research paper. For \arxiv, its class belongs to $40$ subject areas, which are divided into $8$ groups for Task-IL, and the number of classes increases by $5$ for each task in Class-IL. 
For \magdata, For Task-IL and Class-IL, among $349$ classes indicating fields of studies, we use the $257$ classes with at least $10$ nodes, and they are divided into $128$ groups for Task-IL.
For Class-IL, the number of classes grows by $2$ in each task.
For Time-IL, we formulate $24$ and $10$ tasks chronologically using publication years for \arxiv and \magdata, respectively.
Specifically, in \arxiv, we construct the first task with the paper published in the year $1997$ or earlier. For each subsequent $i$-th task ($i\geq 2)$, we include the papers published in the year $(1996 + i)$. In \magdata, each $i$-th task comprises papers published in the year $(2009 + i)$.
Note that these scenarios and previous ones (i.e., \cora, \citeseer, and \corafull) are directly related to the real-world task of categorizing publications by research fields (e.g., Computing Classification System (CCS) concepts), which is important for enhancing research organization and retrieval.

\item Nodes in \proteins \citep{hu2020open,szklarczyk2019string} are proteins, and edges indicate meaningful associations between proteins.
For each protein, $112$ binary classes, which indicate the presence of $112$ functions, are available. 
Each protein belongs to one among $8$ species, which are used as domains in Domain-IL.
Each of the $8$ tasks consists of $112$ binary classification problems. 
Note that this scenario is directly related to the real-world application of protein function prediction, which is vital for understanding biological processes, contributing to drug discovery, the development of new therapies, and more.

\item \products \citep{hu2020open,chiang2019cluster} is a co-purchase network, where each node is a product, and its class belongs to $47$ categories, which are divided into $9$ groups for Class-IL. 
The number of classes increases by $5$ for each task, excluding two categories.

\item Nodes in \twitch \citep{rozemberczki2021twitch} are users of a video-streaming platform, and edges indicate mutual follower relationships between users. For each user, its class indicates whether the user is joining an affiliate program or not, which 
can be important in real-world company decision-making.
Each user belongs to one among $21$ broadcasting language groups, and they are used as domains for Domain-IL. 

\end{itemize}

\subsubsection{Dataset for Link-Level Problems.}
\begin{itemize}[leftmargin=*]

\item \bitcoin \citep{kumar2016edge,kumar2018rev2} is a who-trust-whom network, where nodes are users of a bitcoin-trading platform.
Each directed edge has an integer rating between $-10$ to $10$ and a timestamp.
For Task-IL and Class-IL settings, we divide the directed edges into six rating-based groups (i.e., $[-10, -9)$, $[-9, 0)$, $[3, 4)$, $[4, 5)$, $[5, 6)$, and $[6, 10]$) to minimize class imbalance.
The groups are divided into pairs, which are employed separately for each task in Task-IL and cumulatively for each task in Class-IL.
For Time-IL, we formulate $7$ tasks chronologically using the timestamps, where the signs of the ratings are used as binary classes.

\item \collab \citep{hu2020open,wang2020microsoft} is a co-authorship network between authors with publication years.
For Time-IL, the first task consists of the co-authorships established in the year $1970$ or earlier; and each subsequent $i$-th task ($i\geq 2)$ consists of those established in the year $(1969 + i)$.

\item \wikics \citep{mernyei2020wiki} is a hyperlink network between computer science articles.
Each article has a label indicating one of the $10$ subfields it belongs to.
The node labels are used as domains, and the edges are divided into $54$ groups, according to the labels of their endpoints.\footnote{Specifically, there are $\binom{10}{2}+\binom{10}{1}=55$ different combinations of the classes (or domains) of links' endpoints, and we ignored one combination since there is no link with that specific combination.}

\item Nodes in \askubuntu \citep{paranjape2017motifs} are users of an online Q/A platform, and edges indicate interactions between them.
The edges are chronologically divided into $69$ groups for Time-IL based on timestamps, ensuring that interactions within each group occur within the same month.

\item In \facebook \citep{rozemberczki2019gemsec}, nodes are pages on Facebook, and edges indicate mutual ``likes'' between the pages that fall within the same category. Each node is assigned to one of eight categories, and we leverage these categories as domains for Domain-IL. 
\item \gowalla~\citep{liang2016modeling,wang2019neural} consists of check-in history from a location-based social networking platform where users share their locations through check-ins. Each node represents either a user or a location, and each edge represents a user's check-in at a location. For Time-IL, we organize 10 tasks chronologically based on check-in timestamps.
\item \movie~\citep{harper2015movielens} is a movie-rating dataset. We convert it into a graph where nodes represent either users or movies. An edge is created between a user and a movie if and only if the user gives the movie a rating of 4 or higher. For Time-IL, we organize 10 tasks chronologically based on the rating timestamps.
This scenario and the previous one (i.e., \gowalla) are directly related to personalized recommendation systems (spec., movie and POI recommendations), which are essential for helping users find relevant options among numerous candidates.
\end{itemize}

\subsubsection{Datasets for Graph-Level Problems.}
\begin{itemize}[leftmargin=*]
\item Images in \mnist and \cifar~\citep{dwivedi2020benchmarking,achanta2012slic} are converted to graphs of super-pixels.
There are $10$ classes of graphs, and they are partitioned into $5$ groups, which are used separately for each task in Task-IL and cumulatively for each task in Class-IL.
\item Molecules in \molhiv~\citep{hu2020open,wu2018moleculenet,landrum2006rdkit}, \aroma~\citep{xiong2019pushing,wu2018moleculenet}, \zinc~\citep{gomez2018automatic}, and \aqsol~\citep{sorkun2019aqsoldb} are represented as graphs, where nodes correspond to atoms and edges correspond to chemical bonds.
For \molhiv, the binary class of each graph indicates whether molecules inhibit HIV replication or not, and we divide the molecules into $20$ groups based on structural similarity by the scaffold splitting procedure \citep{landrum2006rdkit}. 
\aroma contains labels representing the number of aromatic atoms in each molecule, and in this paper, we divide the molecules into $30$ groups based on the labels and formulate Task- and Class-IL settings with $10$ tasks.
For \molhiv and \aroma, we only consider the $30$ classes with at least 20 graphs, as in \citep{zhang2022cglb}.
\zinc and \aqsol contain real values representing aqueous solubility, which are used as ground-truth values for graph regression.
For Domain-IL, we divide the molecules in \zinc into 11 groups based on molecular size, and those in \aqsol into 5 groups based on structural similarity, using the scaffold splitting procedure \citep{landrum2006rdkit}.
Note that predicting molecular properties (such as the number of aromatic atoms in \aroma, HIV inhibition in \molhiv, and aqueous solubility in \zinc and \aqsol) has significant real-world applications, including accelerating drug discovery, improving treatment efficacy, and guiding the development of targeted therapies.

\item Each graph in \nyctaxi~\citep{nyctaxi} shows the amount of taxi traffic between locations in New York City over each hourly period from $2018$ to $2021$.
Specifically, nodes are locations, and there exists a directed edge between two nodes if there exists any taxi customer between them during an hour. 
The number of such customers is used as the edge weight.
We split the locations into seven groups based on borough information provided in \citep{nyctaxi} and use one-hot indicators as the node features.
The date and time of the corresponding taxi traffic are used to partition the graphs chronologically into intervals of three months, resulting in $16$ groups, for Time-IL.
The binary class of each graph indicates whether it indicates taxi traffic on weekdays or weekends. 

\item Graphs in \ppa \citep{hu2020open, szklarczyk2019string, hug2016new, zitnik2019evolution} are protein-protein interactions. We formulate a multi-class (spec., $37$-class) classification problem of predicting which taxonomic groups of species each graph comes from. The dataset is preprocessed to contain $11$ species for each taxonomic group and $100$ graphs for each species. We categorize the species so that each category contains exactly one species from each taxonomic group, and the categories are used as domains for Domain-IL.
Note that the taxonomic classification of living organisms is important for understanding biodiversity, studying evolutionary relationships, and discovering new species.

\item Graphs in \sentiment \citep{go2009twitter} are dependency trees parsed from posts on Twitter, a social media platform. 
To generate the graphs, we randomly sample $250$ posts with positive sentiment and $250$ posts with negative sentiment for each day from May $26$ to July $7$, 2009. 
Each sampled post is processed using the \texttt{en\_core\_web\_lg} pipeline\footnote{\texttt{en\_core\_web\_lg} contains $514$k unique word vectors of 300 dimensions.} from the SpaCy library \citep{honnibal2020spacy} to derive graphs (spec., dependency trees) from the online posts.
Each graph captures the dependency between the tokens (i.e., nodes) in the post and token embeddings (i.e., node features).
The class of each graph indicates whether the sentiment of the corresponding post is positive or negative. 
For Time-IL, we formulate $11$ tasks chronologically using the timestamps of posts.
This scenario is directly related to the real-world task of sentiment classification, which is used to understand public opinion, enabling businesses and organizations to make informed decisions by analyzing customer feedback, social media trends, and market sentiments.

\end{itemize}

\section{\method: Proposed Benchmark Framework}
\label{sec:framework}

\begin{figure*}[t]
\centering
\includegraphics[width=\linewidth]{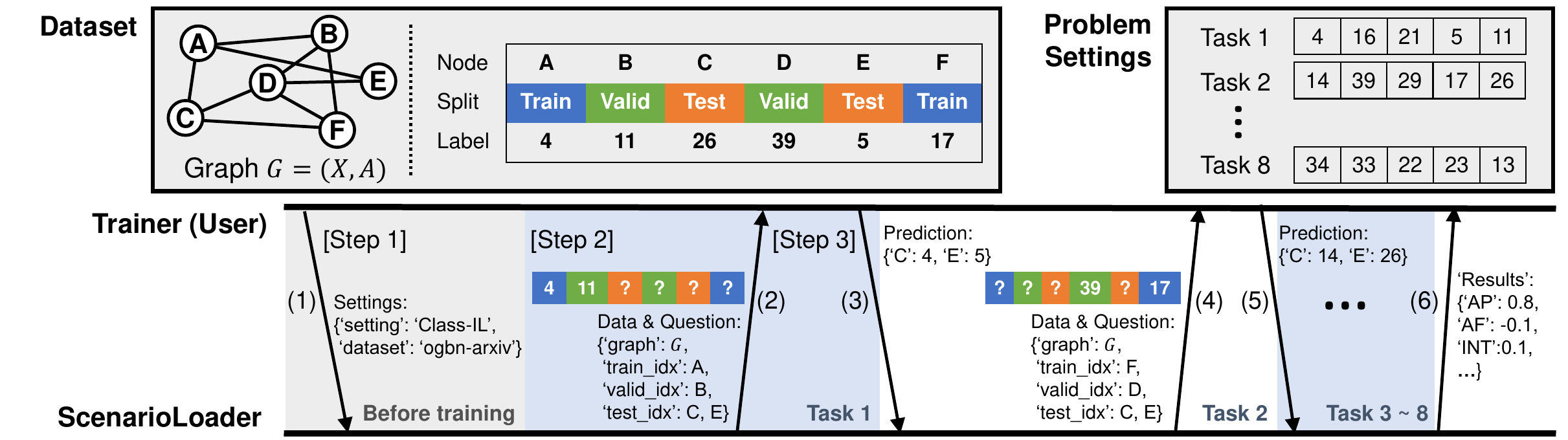}
\caption{\label{fig:loader_overview} Example communications between the trainer (user code) and the loader.}
\end{figure*}

\begin{figure}[t]

\centering
\includegraphics[width=\linewidth]{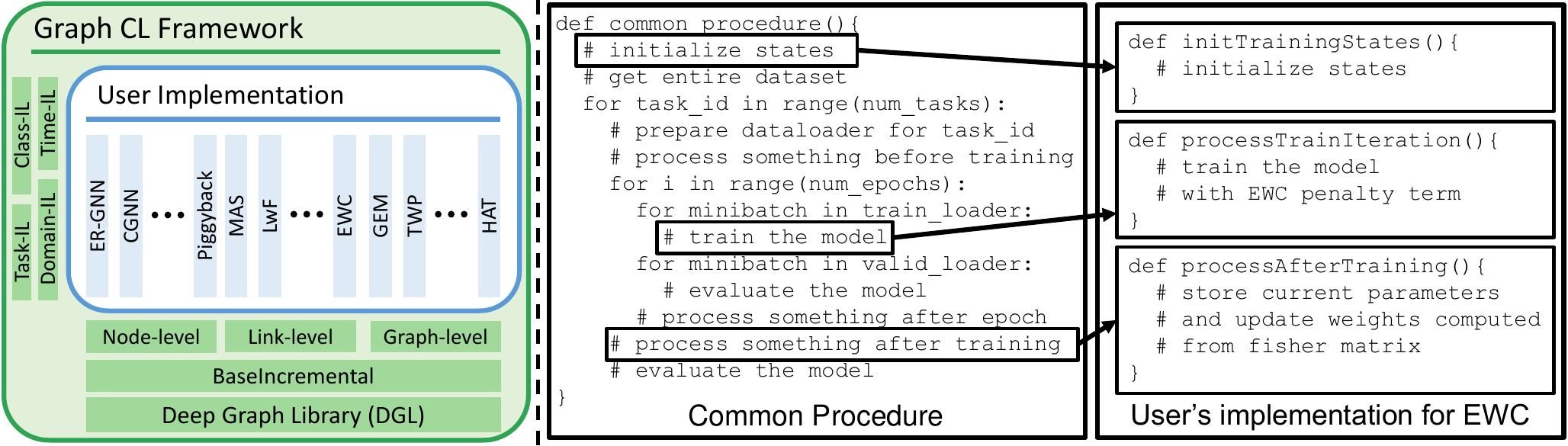}
\caption{\textbf{(Left) Modularized structure of \method}, our proposed benchmark framework for implementation and evaluation of continual learning methods for graph data.
\textbf{(Right) An example implementation of EWC with \ourfw.} To implement and benchmark new graph CL methods, users only need to fill out the modularized event functions in the trainer, which then proceeds the training procedure with the event functions. We provide detailed explanations for implementing EWC with \ourfw in Appendix~\ref{sec:app:ewc}.}
\label{fig:overview}
\end{figure}

In this section, we present \method (\textbf{Be}nchmarking \textbf{G}raph Cont\textbf{i}\textbf{n}ual Learning), our proposed benchmark framework for making graph continual learning (graph CL) \textbf{(a) Easy}: assisting users so that they can implement new graph CL methods with little effort, \textbf{(b) Fool-proof:} preventing potential mistakes of users in evaluation, which is complicated for graph CL, and \textbf{(c) Extensive}: supporting various benchmark scenarios, including those described in Section~\ref{sec:scenarios}.
The modularized structure of \method is illustrated on the left side of Figure~\ref{fig:overview},

\subsection{ScenarioLoader (Loader)}

The ScenarioLoader (loader in short) of \ourfw is responsible for communicating with user code (i.e., the training part) to perform a benchmark under a desired incremental setting.

First, the loader receives the entire dataset and the desired incremental setting as inputs. Then, according to the inputs, it processes the dataset into a sequence of tasks, as described in Section \ref{sec:scenarios}.
Before each task starts, the loader provides (a) the input for the task and
(b) the set $\SQ$ of queries for evaluation to the user code.
Once the user code is done with the current task, the loader receives the predicted answers for the queries in $\SQ$.
Lastly, if there is no more task to be performed, the loader returns the evaluation results, which are computed by the evaluator module, to the user code.
In Figure~\ref{fig:loader_overview}, we provide an example of such communications.

\smallsection{Remarks on the fool-proofness of \method.}
The evaluation part is intentionally concealed from user code, even after all tasks are processed. By hiding the ground-truth answers to the queries, we aim to prevent potential mistakes and misuse by users.
It should be noticed that, compared to continual learning with independent data, unintentional information leak is easier to happen for node- and link-level problems due to the dependency between tasks.
For example, for an NC task, not only nodes assigned to the current task but also those assigned to previous or future tasks can be used (e.g., for graph convolution) because they are connected by edges.
Moreover, our framework additionally restricts information given for each task to prevent potential information leaks. For example, all queries in $\SQ$ are asked by the loader and answered by the user code at each evaluation step, even when some of the queries are on unseen tasks. Otherwise, information about the tasks that queries are on can be revealed to the user code and exploited for answering the questions, which is prohibited in the Class-IL setting.
Note that the objective of our design is to reduce potential mistakes (i.e., unintentional information leaks), but not to prevent adversarial users from attempting to cheat by any means.

\subsection{Evaluator}
\label{sec:framework:evaluator}

\ourfw provides the evaluator to compute basic metrics (spec., accuracy, AUCROC, and Hits@K) based on the ground truth and predicted answers for the queries in $\SQ$ provided by the loader.
The basic evaluator can easily be extended by users for additional basic metrics. The basic metrics are sent to 
the loader, and for each basic metric, the basic performance matrix $\MM\in \mathbb{R}^{N \times N}$, where $N$ is the number of tasks, is computed.  
The $(i,j)$-th entry $\MM_{i,j}$ indicates the performance on $j$-th task $\STj$ after the $i$-th task $\STi$ is processed.
Based on $\MM$, the following evaluation metrics are computed.

\begin{itemize}[leftmargin=*]
    \item \textbf{Average Performance (AP):} Average performance on $k$ tasks after learning the $k$-th task $\STk$.
    
    \item \textbf{Average Forgetting (AF):} Average forgetting on ($k-1$) tasks after learning the $k$-th task $\STk$ ($2\leq k\leq N$). We measure the forgetting on $\STi$ by the difference between the performance on $\STi$ after learning $\STk$ and the performance on $\STi$ right after learning $\STi$. 

    \item \textbf{Intransigence (INT)} \citep{chaudhry2018riemannian} averaged over $k$ tasks: We measure the intransigence on $\STi$ by the difference between the performances of the Joint model (see Section~\ref{sec:exp:setting}) and the target model on $\STi$ after learning $\STi$.
    
    \item \textbf{Forward Transfer (FWT)} \citep{lopez2017gradient} averaged over ($k-1$) tasks ($2\leq k\leq N$): We measure the forward transfer on $\STi$ by the difference between the performance on $\STi$ after learning $\mathcal{T}_{i-1}$ and the performance on $\STi$ without any learning. 
\end{itemize}

Note that \textbf{AF} quantifies forgetting of previous tasks, and \textbf{INT} measures performance on the current task.
\textbf{FWT} measures performance on future tasks, to quantify generalizable knowledge retained from previous tasks.
Formally, the evaluation metrics are defined as follows:

\begin{align*}
\mathrm{\mathbf{AP}}=\sum\limits_{i=1}^{k}\frac{\MM_{k,i}}{k}, \
\mathrm{\mathbf{AF}}=\sum\limits^{k-1}_{i=1}\frac{\MM_{i,i}-\MM_{k,i}}{k-1}, \
\mathrm{\mathbf{INT}}=\sum\limits_{i=1}^{k}\frac{\MM^{\text{Joint}}_{i, i} - \MM_{i, i}}{k}, \
\mathrm{\mathbf{FWT}}=\sum\limits^{k}_{i=2}\frac{\MM_{i-1,i}-r_i}{k-1},
\end{align*}
where $\MM^{\text{Joint}}$ is a basic performance matrix of the Joint model,
and $r_i$ denotes the performance of a randomly initialized model on $\STi$.

\subsection{Trainer}
For usability, \ourfw provides the trainer, which users can extend when implementing new methods.
It manages the overall training procedure, including data loading, training, and validation, so that users only have to implement novel parts of their methods. 
As in \citep{lomonaco2021avalanche}, the trainer divides a training procedure of continual learning as a series of events. 
For example, the subprocesses in a training procedure contain events called when the trainer (a) receives an input for the current task, (b) trains a model for one iteration for the current task, and (c) handles any necessary pre- and post-processing for the training procedure.
Each event is modularized as a function, which users can fill out, and the trainer proceeds with the training procedure with the event functions.

One thing we need to consider is that there can be cases where intermediate results generated in each event must be stored to be used in other events. 
For example, in the EWC method, as described comprehensively in Appendix~\ref{sec:app:ewc},
a penalty term for preventing catastrophic forgetting should be additionally considered to compute the training loss. 
To compute the term, the learned parameters and the weights computed from the fisher information matrix on the previous tasks are needed, but they cannot be obtained on the current task. In order to resolve this issue, the trainer provides a dictionary where intermediate results can be stored and shared by events. For the aforementioned EWC method, the learned parameters and computed weights on a task are stored in the dictionary and used for computing the training loss on the following tasks.
The right side of Figure~\ref{fig:overview} shows how the EWC method for node classification can be implemented with \ourfw.
In this example, users only need to fill out three event functions: (a) \texttt{initTrainingStates} for initializing the dictionary of training states, (2) \texttt{processTrainIteration} (or \texttt{afterInference}, which is called inside its default implementation) for computing the penalty term in the loss function,
and (3) \texttt{processAfterTraining} for storing the learned parameters and computing the weights of the parameters for the penalty term.
In Appendix~\ref{sec:app:ewc}, we present a code-level implementation of EWC using \ourfw to demonstrate the usability of \ourfw in detail.

\smallsection{Event Functions}.
In Appendix~\ref{sec:app:eventfunctions}, we list the event functions supported by \ourfw along with brief descriptions. 
Notably, \ourfw offers event functions for self-supervised learning (SSL) methods. Using them, users can apply four SSL methods provided by \ourfw (GraphCL~\cite{you2020graph}, InfoGraph~\cite{sun2019infograph}, Deep Graph Infomax~\cite{velivckovic2018deep}, and LightGCL~\cite{cai2023lightgcl}) for pre-training or create and apply their own custom SSL methods (refer to Appendix~\ref{app:ssl} for the experimental results on the impact of applying SSL techniques).
Further details about the event functions, including their arguments, can be found in the official document of \ourfw, which is available at \citep{code}.

\subsection{Continual Learning Methods Implemented and Available on \ourfw}
\label{sec:framework:method}

Thanks to the flexibility of \ourfw, \numalgo continual learning methods have been implemented seamlessly on \ourfw, and they are accessible at \citep{code}. 
These methods fall into the following categories:
\begin{itemize}[leftmargin=*]
    \item General \& replay-based methods: GEM~\citep{lopez2017gradient}
    \item General \& regularization-based methods: EWC~\citep{kirkpatrick2017overcoming}, LwF\citep{li2017learning}, and MAS~\citep{aljundi2018memory} 
    \item General \& parameter-isolation-based methods: PackNet~\citep{mallya2018packnet}, Piggyback~\citep{mallya2018piggyback}, and HAT~\citep{serra2018overcoming}
    \item Graph-specific \& replay-based methods: ERGNN~\citep{zhou2021overcoming} and CaT~\citep{liu2023cat}
    \item Graph-specific \& regularization-based methods: TWP~\citep{liu2021overcoming}
    \item Graph-specific \& parameter-isolation-based methods: PI-GNN~\citep{zhang2023continual}
    \item Graph-specific \& hybrid-based methods: CGNN~\citep{wang2020streaming}
\end{itemize}
Refer to Section~\ref{sec:related} for brief descriptions of these methods.

\section{Benchmark Results}
\label{sec:experiments}
In this section, we provide the details of experimental settings and benchmark results of $10$ graph CL methods implemented by \ourfw.

\subsection{Experimental Settings}
\label{sec:exp:setting}

\smallsection{Machines.} 
We performed all experiments on a Linux server with Quadro RTX 8000 GPUs.

\smallsection{Methods.}
For general CL methods, we used LwF~\citep{li2017learning}, EWC~\citep{kirkpatrick2017overcoming}, MAS~\citep{aljundi2018memory}, GEM~\citep{lopez2017gradient}, PackNet~\citep{mallya2018packnet}, Piggyback~\citep{mallya2018piggyback}, and HAT~\citep{serra2018overcoming}.
The general parameter-isolation-based methods for independent data (i.e., PackNet, Piggyback, and HAT) were adapted for graph CL, as described in Appendix~\ref{sec:app:pim}, and they were applied only to Task-IL settings since they require knowledge of the specific task associated with each query.
For CL methods designed for graph-structured data, we used TWP~\citep{liu2021overcoming}, ERGNN~\citep{zhou2021overcoming}, CGNN~\citep{wang2020streaming}, CaT~\citep{liu2023cat}, and PI-GNN~\citep{zhang2023continual}.
ERGNN and CGNN, which were designed for node-level problems, were not applied to link- and graph-level problems. 
See Section~\ref{sec:related} for brief descriptions of the above methods.
For the baseline methods without CL techniques, we used the Bare and Joint models used in \citep{zhang2022cglb}. 
The Bare model trains a backbone model (e.g., graph neural networks) incrementally without employing specific strategies for mitigating catastrophic forgetting.
Specifically, it updates the model parameters obtained from the previous task, (e.g., through gradient descent), based on the label information in the current task (i.e., node labels in NC, ground-truth edges in LP, graph labels in GC, and ground-truth values in GR).
The Joint model trains a backbone model using the entire dataset, incorporating label information from all tasks, while ignoring the CL procure with a sequence of tasks.
 
\smallsection{Models.}
For all experiments, we used GCN \citep{kipf2016semi} as the backbone model to compute node embeddings, and we used the Adam \citep{kingma2014adam} optimizer to train the model. 
For NC, we used a fully-connected layer right after the backbone model to compute the final output.
For LC and LP, we additionally used a $3$-layer MLP that takes a pair of node embeddings and outputs the final embeddings of the pair. 
For GC, we used mean pooling to obtain graph embeddings from the output of the backbone model and fed the embeddings into a 3-layer MLP.
For NC and LC, we used 3 GCN layers with a hidden dimension of 256.
For GC and GR, we used 4 GCN layers with a hidden dimension of 146, as in \citep{dwivedi2020benchmarking}. For experiments with PI-GNN, where the model progressively expands, we adjusted the hidden dimensions of the models so that the number of parameters in the model at the final task is similar to those of other competitors.

\smallsection{Training Protocol.}
We set the number of training epochs to $1,000$ for \cora, \citeseer, \arxiv, and \bitcoin; 200 for \proteins and \red{\twitch}, \wikics, \collab, \red{\facebook, and \askubuntu}; and $100$ for \magdata, \products, and all GC datasets.
For all datasets except for \products, we performed full-batch training. 
For \products, we trained GCN with the neighborhood sampler provided by DGL for mini-batch training.
For all experiments, we used early stopping. Specifically, for NC and LC, we reduced the learning rate by a factor of $10$, if the performance did not improve after $20$ epochs, and stopped the experiment if the learning rate became over $1,000$ times smaller than the initial learning rate. For \citeseer, we used the same stopping criteria, with a patience of $50$ epochs. For LP, GC, and GR, we reduced the learning rate by a factor of $10$, if performance did not improve after $10$ epochs, and stopped the experiment if the learning rate became $100$ times ($1,000$ times in \wikics) smaller the initial learning rate.

\smallsection{Hyperparameter Settings.} 
We performed a grid search to find the best hyperparameter (e.g., the learning rate and the dropout ratio) settings for the backbone model for each method.
Specifically, we chose the setting where AP on the validation set was maximized and reported the results on the test set in the selected setting. For evaluation, we conducted $10$ experiments with different random seeds and reported the mean performance.
For the CL methods, we set the initial learning rate among $\{$1e-3, 5e-3, 1e-2$\}$, the dropout ratio among $\{$0, 0.25, 0.5$\}$, and the weight decay coefficient between $\{$0, 5e-4$\}$.
For the experiments on \proteins, \magdata (for both task-IL and class-IL settings), \products, and \ppa, we fixed the learning rate to 1e-3. For \citeseer, we additionally considered the learning rate 5e-4.
For \magdata (for both task-IL and class-IL settings), \products, and \ppa, we set the dropout ratio between $\{0, 0.25\}$. In addition, for \products, we restricted the number of maximum neighbors from which messages are aggregated to $5$, $10$, and $10$ on the first layer, the second layer, and the third layer, respectively, of GCN. 

For replay-based methods (e.g., GEM, ERGNN, CGNN, and CaT), 
we set the maximum size of memory the same for a fair comparison. 
Specifically, we set it to $12$ for (a) all experiments on \cora and \citeseer, (b) $210$ for \corafull, (c) $2,000$ for \arxiv, \proteins, \twitch, and \askubuntu, (d) $8,000$ for \magdata, (e) $25,000$ for \products, (f) $500$ for \mnist, \cifar, \molhiv, and \ppa, (g) $50$ for \aroma, (h) $800$ for \nyctaxi, (i) $4,000$ for \wikics, (j) $20,000$ for \collab and \facebook, and (k) $60$ for \sentiment.
Following the original papers, we set the margin for quadratic programming to $0.5$ for GEM, and we used the Coverage Maximization (CM) sampler and set the distance threshold to $0.5$ for ERGNN.

For regularization methods (e.g., LwF, EWC, MAS, TWP, and CGNN), 
we set the regularization coefficient $\lambda$ to $1.0$, $10000.0$, $1.0$, and $80.0$ for LwF, EWC, MAS, and CGNN, respectively.
On \cora, \citeseer, \arxiv, \proteins, \mnist, \cifar, \nyctaxi, \twitch, \facebook, and \sentiment, we additionally considered $0.1$ for LwF, $100.0$ for EWC, and $0.1$ for MAS as a potential value of $\lambda$.
For TWP, we set $\beta$ to $0.01$ and $\lambda_{l}$ to $10,000$, and we chose $\lambda_{t}$ between $\{100, 1000\}$.
Note that CGNN combines regularization- and replay-based approaches, and thus we need to consider both the maximum size of memory and the regularization coefficients.

For parameter-isolation-based methods (e.g. PackNet, Piggyback, and HAT), we set their hyperparameters as follows. For all parameter-isolation-based methods, we set the weight decay ratio to $0$ to maintain the parameters for previous tasks.
For PackNet, we set the pruning ratio $p$ to $\exp(\frac{1}{N-1} \log \frac{1}{N})$, where $N$ is the number of tasks, so that the model can learn the parameters of the ratio of $1/N$ at the last task.
In addition, we set the number of pre-training epochs to $10\%$ of the number of training epochs. 
For Piggyback, we set the threshold $\tau$, which is for determining the binary mask for the parameters of each task, between $\{$1e-1, 1e-2$\}$. 
For HAT, we set the compressibility $c$ to $0.75$ and set the stability $s_{\text{max}}$ to $400$. For PI-GNN, we used the same memory budget with the replay-based methods and set the coefficient $\lambda$, which handles the size imbalance problem in the parameter-isolation phase, between \{1e-1, 1e-0\}. In addition, for Time-IL, we set the number of epochs for knowledge rectification stage to 10\% of the number of training epochs.\footnote{Note that, for Task-IL, Class-IL, and Domain-IL settings, the knowledge rectification stage is not needed since the topological structure of the input graph does not change.} Due to their large volume, the hyperparameter settings used for each considered scenario are available at ~\citep{code}.

\begin{table*}
\caption{\textbf{Results of Average Performance (AP, the higher, the better), Average Forgetting (AF, the lower, the better), and Intransigence (INT, the lower, the better)}. The best score is in bold, and the second-best score is underlined. Each number is rescaled to a range of $100$. O.O.M: out of memory. N/A: the considered methods are not applicable to the problems or scenarios. We report the full results in Appendix~\ref{sec:app:add_results}.}
\begin{subtable}[h]{\textwidth}
\setlength{\tabcolsep}{1pt}
\renewcommand{\arraystretch}{1.02}
\centering
\scalebox{0.68}{
\begin{tabular}{c|cccc|cc|cc|cccc}
\toprule
\multirow{3}{*}{Methods} & \multicolumn{4}{c|}{Node Classification} & \multicolumn{2}{c|}{Link Classification} & \multicolumn{2}{c|}{Link Prediction} & \multicolumn{4}{c}{Graph Classification} \\
& \makecell{\cora \\ (Task-IL)} & \makecell{\citeseer \\ (Class-IL)} & \makecell{\proteins \\ (Domain-IL)} & \makecell{\arxiv \\ (Time-IL)} & \makecell{\bitcoin \\ (Task-IL)} & \makecell{\bitcoin \\ (Class-IL)} & \makecell{\wikics \\ (Domain-IL)} & \makecell{\collab \\ (Time-IL)} & \makecell{\cifar \\ (Task-IL)} & \makecell{\mnist \\ (Class-IL)} & \makecell{\molhiv \\ (Domain-IL)} & \makecell{\nyctaxi \\ (Time-IL)} \\
\midrule
Bare & 90.3±1.8 & 44.7±4.0 & 69.0±2.2 & 62.8±0.4 & 64.8±7.1 & 24.3±3.0 & 6.8±3.5 & \uls{34.7±2.5} & 64.6±7.4 & \uls{19.4±0.5} & 68.6±3.9 & 70.9±0.6 \\
LwF & 91.5±1.2 & 46.4±3.9 & 71.4±2.0 & 63.9±0.4 & 70.4±3.2 & 24.2±2.9 & 6.5±2.8 & 34.3±2.8 & 84.0±3.0 & \uls{19.4±0.5} & 70.2±4.8 & 71.1±0.2 \\
EWC & 91.2±1.3 & 45.2±3.7 & 76.1±1.1 & 61.3±0.4 & 68.2±4.6 & 24.2±3.0 & 11.3±2.6 & 23.2±1.4 & 78.4±4.2 & 19.3±0.6 & \bbf{74.1±3.2} & \bbf{72.0±1.4} \\
MAS & 91.8±1.7 & \bbf{56.0±2.4} & 69.4±1.6 & 58.8±0.7 & \uls{70.6±3.3} & \uls{24.4±2.3} & 10.0±3.5 & 22.7±7.5 & 76.2±4.6 & 19.2±0.6 & 68.0±3.0 & \uls{71.4±0.5} \\
GEM & 88.2±3.1 & 48.2±3.3 & \uls{81.0±0.3} & 65.4±0.3 & 70.0±3.5 & \bbf{28.7±4.2} & \bbf{18.0±2.9} & \bbf{38.8±6.3} & 76.9±3.0 & \bbf{19.9±2.0} & 70.7±3.3 & 70.6±6.6 \\
TWP & 91.0±1.5 & 45.0±3.7 & O.O.M & 59.3±1.0 & 67.3±4.7 & 24.3±2.8 & \uls{13.3±4.3} & 30.8±3.0 & 78.8±4.4 & 19.3±0.5 & \uls{73.9±3.4} & 71.2±0.3 \\
ERGNN & 89.0±3.1 & 45.7±4.3 & N/A & \bbf{69.6±0.4} & N/A & N/A & N/A & N/A & N/A & N/A & N/A & N/A \\
CGNN & 91.1±1.5 & \uls{53.1±3.5} & N/A & \uls{67.6±0.5} & N/A & N/A & N/A & N/A & N/A & N/A & N/A & N/A \\
CaT & 56.9±10.7 & 19.1±3.1 & N/A & 58.8±0.4 & N/A & N/A & N/A & N/A & N/A & N/A & N/A & N/A \\
PI-GNN & 90.8±2.8 & 50.2±6.8 & \bbf{82.0±0.3} & 64.6±0.5 & N/A & N/A & N/A & N/A & N/A & N/A & N/A & N/A \\
PackNet & \uls{93.3±1.8} & N/A & N/A & N/A & \bbf{71.8±3.0} & N/A & N/A & N/A & \bbf{85.5±2.3} & N/A & N/A & N/A \\
Piggyback & \bbf{93.8±1.6} & N/A & N/A & N/A & 68.1±2.3 & N/A & N/A & N/A & \uls{85.2±2.1} & N/A & N/A & N/A \\
HAT & 92.0±2.0 & N/A & N/A & N/A & 62.7±6.1 & N/A & N/A & N/A & 64.3±7.8 & N/A & N/A & N/A \\
\midrule
Joint & 92.4±1.5 & 55.6±4.0 & 73.2±0.2 & 73.4±0.2 & 73.5±3.5 & 37.7±3.1 & 22.9±1.8 & 58.8±2.5 & 86.8±2.1 & 90.0±0.4 & 77.9±1.2 & 86.3±0.4 \\
\bottomrule
\end{tabular}
}
\caption{\label{tab:results_ap}Average Performance (AP)}
\end{subtable}
\begin{subtable}[h]{\textwidth}

\setlength{\tabcolsep}{1pt}
\renewcommand{\arraystretch}{1.02}
\centering
\scalebox{0.65}{
\begin{tabular}{c|cccc|cc|cc|cccc}
\toprule
\multirow{3}{*}{Methods} & \multicolumn{4}{c|}{Node Classification} & \multicolumn{2}{c|}{Link Classification} & \multicolumn{2}{c|}{Link Prediction} & \multicolumn{4}{c}{Graph Classification} \\
& \makecell{\cora \\ (Task-IL)} & \makecell{\citeseer \\ (Class-IL)} & \makecell{\proteins \\ (Domain-IL)} & \makecell{\arxiv \\ (Time-IL)} & \makecell{\bitcoin \\ (Task-IL)} & \makecell{\bitcoin \\ (Class-IL)} & \makecell{\wikics \\ (Domain-IL)} & \makecell{\collab \\ (Time-IL)} & \makecell{\cifar \\ (Task-IL)} & \makecell{\mnist \\ (Class-IL)} & \makecell{\molhiv \\ (Domain-IL)} & \makecell{\nyctaxi \\ (Time-IL)} \\
\midrule
Bare & 2.6±2.3 & 55.0±6.6 & 13.1±3.4 & -4.9±0.6 & 11.4±8.7 & 72.2±4.7 & 29.9±4.3 & 24.5±2.2 & 27.0±8.2 & 97.8±0.8 & 10.7±5.3 & \uls{2.8±1.3} \\
LwF & 1.2±1.9 & 53.9±6.0 & 5.4±2.6 & -4.4±0.6 & 3.5±2.1 & 72.6±4.7 & 31.1±3.1 & 35.1±3.3 & \uls{3.0±1.5} & 97.6±0.9 & 9.1±5.7 & \bbf{1.4±0.7} \\

EWC & 2.0±1.4 & 54.2±6.5 & 7.4±2.5 & -8.4±1.2 & 6.3±4.6 & 72.4±6.0 & 14.6±4.6 & \bbf{5.5±3.7} & 5.5±2.7 & 97.7±0.8 & \uls{3.8±2.8} & 7.1±2.4 \\
MAS & 0.5±0.7 & \bbf{28.3±5.0} & 1.2±2.7 & -11.6±1.6 & 2.9±2.5 & 72.6±4.9 & \uls{13.4±4.7} & \uls{7.3±6.6} & 7.7±3.9 & 97.3±0.8 & \bbf{0.7±3.1} & 5.6±0.6 \\
GEM & 6.0±5.7 & 50.7±5.3 & \uls{0.3±2.8} & \uls{-13.6±0.6} & \uls{2.6±2.1} & \bbf{57.9±10.0} & \bbf{4.4±5.7} & 29.9±6.5 & 10.6±2.3 & \bbf{86.6±8.0} & 6.5±4.0 & 6.5±6.5 \\
TWP & 2.5±1.8 & 54.5±6.3 & O.O.M & -3.9±1.1 & 8.2±6.0 & \uls{72.1±5.5} & 18.3±2.8 & 17.6±1.7 & 5.6±3.0 & 97.4±0.8 & 5.0±3.4 & 5.5±0.9 \\
ERGNN & 5.2±5.9 & 51.8±5.8 & N/A & -9.5±0.5 & N/A & N/A & N/A & N/A & N/A & N/A & N/A & N/A \\
CGNN & 2.3±1.6 & \uls{38.4±5.5} & N/A & -7.6±0.7 & N/A & N/A & N/A & N/A & N/A & N/A & N/A & N/A \\
CaT & 6.3±15.3 & 53.4±8.1 & N/A & -4.0±0.6 & N/A & N/A & N/A & N/A & N/A & N/A & N/A & N/A \\
PI-GNN & \uls{0.0±0.0} & 39.7±13.5 & \bbf{-0.8±0.4} & \bbf{-15.7±0.6} & N/A & N/A & N/A & N/A & N/A & N/A & N/A & N/A \\
PackNet & \uls{0.0±0.0} & N/A & N/A & N/A & \bbf{0.0±0.0} & N/A & N/A & N/A & \bbf{0.0±0.0} & N/A & N/A & N/A \\
Piggyback & \uls{0.0±0.0} & N/A & N/A & N/A & \bbf{0.0±0.0} & N/A & N/A & N/A & \bbf{0.0±0.0} & N/A & N/A & N/A \\
HAT & \bbf{-2.1±1.9} & N/A & N/A & N/A & 13.4±8.2 & N/A & N/A & N/A & 27.2±8.9 & N/A & N/A & N/A \\
\bottomrule
\end{tabular}
}
\caption{\label{tab:results_af}Average Forgetting (AF)}
\end{subtable}
\begin{subtable}[h]{\textwidth}
\setlength{\tabcolsep}{1pt}
\renewcommand{\arraystretch}{1.02}
\centering
\scalebox{0.65}{
\begin{tabular}{c|cccc|cc|cc|cccc}
\toprule
\multirow{3}{*}{Methods} & \multicolumn{4}{c|}{Node Classification} & \multicolumn{2}{c|}{Link Classification} & \multicolumn{2}{c|}{Link Prediction} & \multicolumn{4}{c}{Graph Classification} \\
& \makecell{\cora \\ (Task-IL)} & \makecell{\citeseer \\ (Class-IL)} & \makecell{\proteins \\ (Domain-IL)} & \makecell{\arxiv \\ (Time-IL)} & \makecell{\bitcoin \\ (Task-IL)} & \makecell{\bitcoin \\ (Class-IL)} & \makecell{\wikics \\ (Domain-IL)} & \makecell{\collab \\ (Time-IL)} & \makecell{\cifar \\ (Task-IL)} & \makecell{\mnist \\ (Class-IL)} & \makecell{\molhiv \\ (Domain-IL)} & \makecell{\nyctaxi \\ (Time-IL)} \\
\midrule
Bare&	0.1±1.3&	-12.9±5.3&	-6.7±0.6&	-5.3±1.3&	0.5±1.4&	-21.2±6.7&	\uls{-14.7±2.2}&	3.1±3.7&	\bbf{0.0±0.3}&	\bbf{-4.5±0.5}&	\uls{-0.4±2.5}&	9.8±1.0 \\ 
LwF&	-0.1±0.5&	\bbf{-13.9±5.2}&	-2.3±2.1&	-6.9±1.7&	\uls{0.1±1.6} &	\uls{-21.3±6.4}&	\bbf{-15.5±3.4}&	\bbf{-6.8±4.6}&	\uls{0.1±0.1}&	\uls{-4.3±0.6}&	\bbf{-0.5±1.9}&	10.9±0.8 \\
EWC&	-0.5±1.1&	-12.9±5.2&	\bbf{-8.9±0.9}&	-0.5±2.1&	0.4±1.0&	-21.2±6.6&	-4.2±6.1&	33.3±5.1&	2.1±0.6&	-4.2±0.6&	0.7±1.4&	\bbf{4.7±1.8} \\
MAS&	-0.1±1.0&	-6.4±5.0&	3.4±2.6&	5.1±2.0&	0.3±1.8 &	\bbf{-21.5±7.0}&	-1.7±3.1&	32.0±5.1&	5.1±1.1&	-3.9±0.6&	9.8±2.7&	6.7±0.8 \\
GEM&	-0.1±0.7&	\uls{-13.5±5.3}&	\uls{-8.3±2.5}&	0.4±1.8&	1.1±1.3&	-16.0±8.0&	-0.9±4.1&	\uls{-6.2±3.8}&	1.1±0.5&	4.0±5.2&	1.5±1.9&	\uls{6.6±2.3} \\
TWP&	-0.6±0.9&	-12.8±5.3&	O.O.M&	-2.8±1.5&	\bbf{0.0±1.2}&	-21.0±6.0&	-9.8±6.0 &	13.9±5.0 &	1.6±0.4&	-4.0±0.6&	-0.3±0.9&	7.0±1.0 \\
ERGNN&	-0.3±1.1&	-11.8±5.6&	N/A&	\bbf{-7.7±1.6}&	N/A&	N/A&	N/A&	N/A&	N/A&	N/A&	N/A&	N/A \\
CGNN&	-0.5±0.9&	-10.2±5.1&	N/A&	\uls{-7.6±1.4}&	N/A&	N/A&	N/A&	N/A&	N/A&	N/A&	N/A&	N/A \\
CaT & 31.0±10.0 & 13.8±10.2 & N/A & -2.0±1.6 & N/A & N/A & N/A & N/A & N/A & N/A & N/A & N/A \\
PI-GNN & 1.3±1.5 & -8.2±6.2 & -7.6±3.2 & 3.5±1.5 & N/A & N/A & N/A & N/A & N/A & N/A & N/A & N/A \\
PackNet&	\uls{-1.2±0.9}&	N/A&	N/A&	N/A&	\uls{0.1±1.9}&	N/A&	N/A&	N/A&	0.4±0.3&	N/A&	N/A&	N/A \\
Piggyback&	\bbf{-1.7±1.2}&	N/A&	N/A&	N/A&	3.1±3.3&	N/A&	N/A&	N/A&	1.4±0.7&	N/A&	N/A&	N/A \\
HAT&	1.5±1.6&	N/A&	N/A&	N/A&	0.3±1.3 &	N/A&	N/A&	N/A&	\bbf{0.0±0.3}&	N/A&	N/A&	N/A \\
\bottomrule
\end{tabular}
}
\caption{\label{tab:results_int}Intransigence (INT)}
\end{subtable}

\end{table*}

\begin{figure*}[t]
    \centering
    \includegraphics[width=0.8\textwidth]{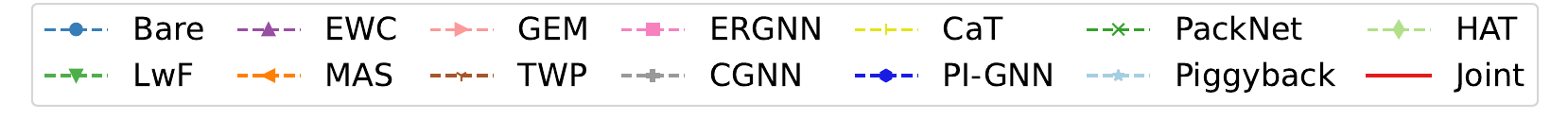}
    
    \begin{subfigure}[b]{0.242\textwidth}
         \centering
         \includegraphics[width=\textwidth]{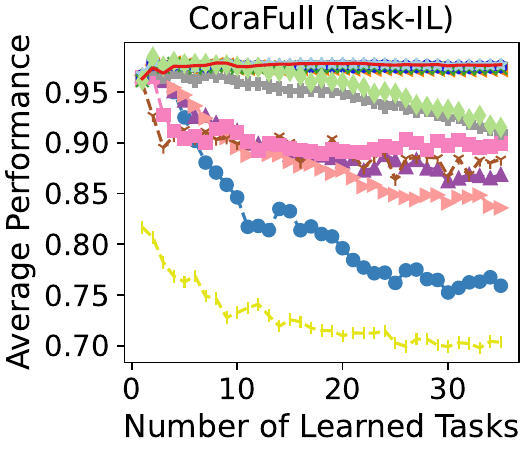}
         \caption{\makecell{\corafull \\ (NC, Task-IL)}}
         \label{fig:node_task_corafull}
     \end{subfigure}
     \begin{subfigure}[b]{0.242\textwidth}
         \centering
         \includegraphics[width=\textwidth]{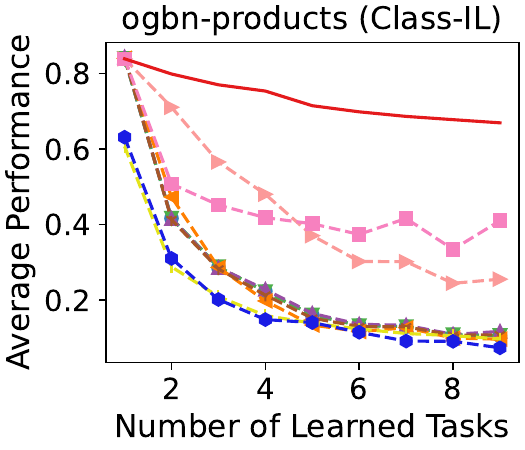}
         \caption{\makecell{\products \\ (NC, Class-IL)}}
         \label{fig:node_class_products}
     \end{subfigure}
     \begin{subfigure}[b]{0.242\textwidth}
         \centering
         \includegraphics[width=\textwidth]{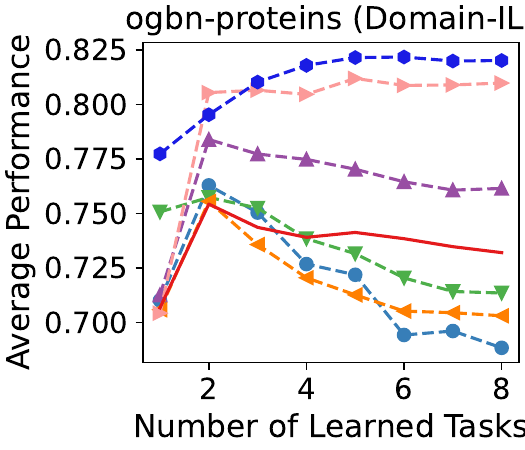}
         \caption{\makecell{\proteins \\ (NC, Domain-IL)}}
         \label{fig:node_domain_proteins}
     \end{subfigure}
     \begin{subfigure}[b]{0.242\textwidth}
         \centering
         \includegraphics[width=\textwidth]{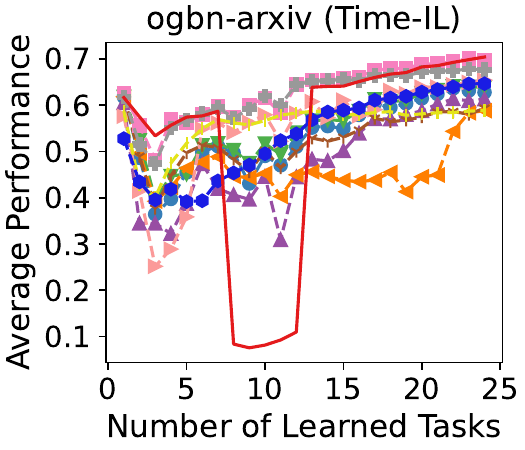}
         \caption{\makecell{\arxiv \\ (NC, Time-IL)}}
         \label{fig:note_time_arxiv}
     \end{subfigure}
     
     \begin{subfigure}[b]{0.242\textwidth}
         \centering
         \includegraphics[width=\textwidth]{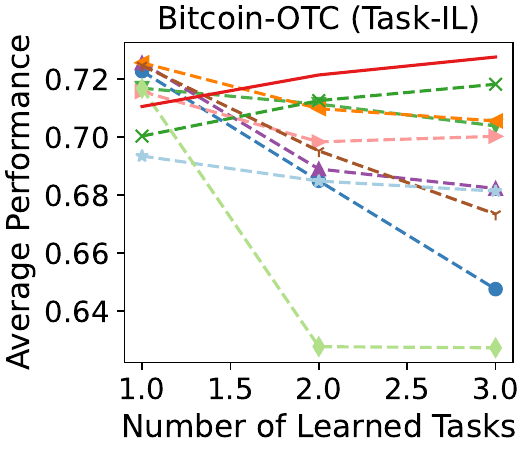}
         \caption{\makecell{\bitcoin \\ (LC, Task-IL)}}
         \label{fig:link_task_bitcoin}
     \end{subfigure}
     \begin{subfigure}[b]{0.242\textwidth}
         \centering
         \includegraphics[width=\textwidth]{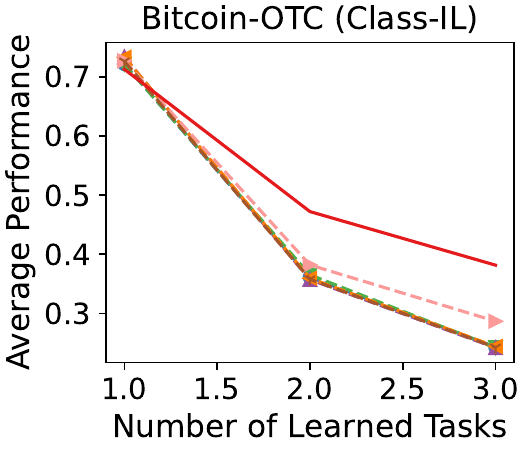}
         \caption{\makecell{\bitcoin \\ (LC, Class-IL)}}
         \label{fig:link_class_bitcoin}
     \end{subfigure}
     \begin{subfigure}[b]{0.242\textwidth}
         \centering
         \includegraphics[width=\textwidth]{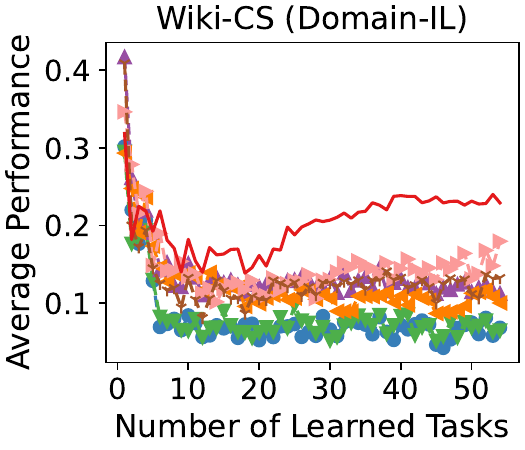}
         \caption{\makecell{\wikics \\ (LP, Domain-IL)}}
         \label{fig:link_domain_wikics}
     \end{subfigure}
     \begin{subfigure}[b]{0.242\textwidth}
         \centering
         \includegraphics[width=\textwidth]{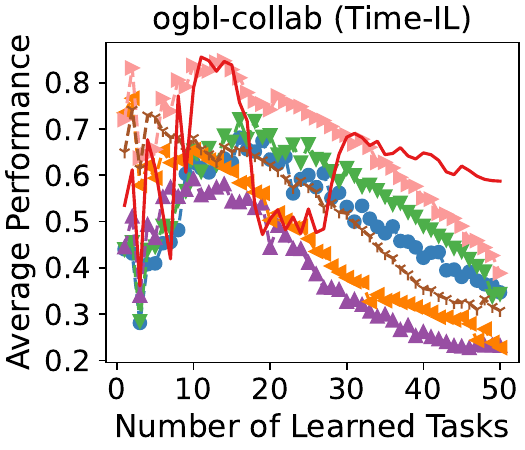}
         \caption{\makecell{\collab \\ (LP, Time-IL)}}
         \label{fig:link_time_ogbl-collab}
     \end{subfigure}
     \begin{subfigure}[b]{0.242\textwidth}
         \centering
         \includegraphics[width=\textwidth]{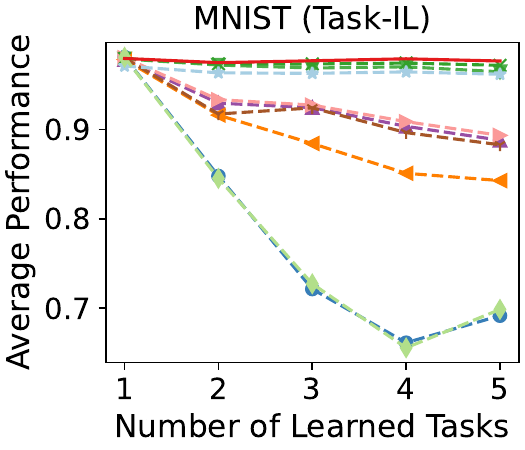}
         \caption{\makecell{\mnist \\ (GC, Task-IL)}}
         \label{fig:graph_task_mnist}
     \end{subfigure}
     \begin{subfigure}[b]{0.242\textwidth}
         \centering
         \includegraphics[width=\textwidth]{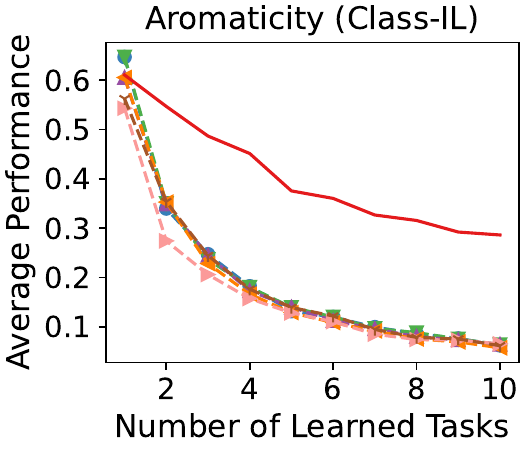}
         \caption{\makecell{\aroma \\ (GC, Class-IL)}}
         \label{fig:graph_class_aromaticity}
     \end{subfigure}
     \begin{subfigure}[b]{0.242\textwidth}
         \centering
         \includegraphics[width=\textwidth]{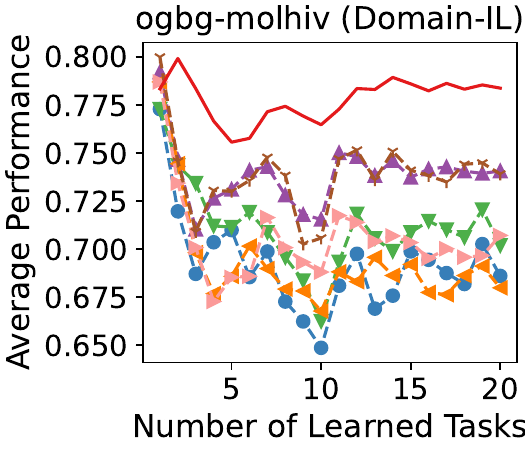}
         \caption{\makecell{\molhiv \\ (GC, Domain-IL)}}
         \label{fig:graph_domain_molhiv}
     \end{subfigure}
     \begin{subfigure}[b]{0.242\textwidth}
         \centering
         \includegraphics[width=\textwidth]{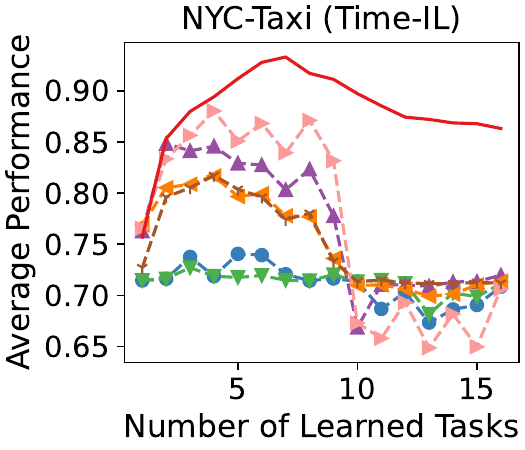}
         \caption{\makecell{\nyctaxi \\ (GC, Time-IL)}}
         \label{fig:graph_time_nyctaxi}
     \end{subfigure}
     \caption{\textbf{Change of Average Performance (AP) during continual learning.} 
     NC: Node Classification. LC: Link Classification. LP: Link Prediction. GC: Graph Classification.
     Note that the Joint model, which is trained using the entire dataset together, sometimes (e.g., in (d)) suffers from instability, especially when training samples for some classes are very limited in Time-IL.
     The full results for all considered scenarios are available in Appendix~\ref{sec:app:add_results}.}
     \label{fig:app_perfchange}
\end{figure*}

\subsection{Average Performance (AP), Average Forgetting (AF), and Intransigence (INT)}
\label{sec:exp:results}

Our benchmark results in terms of final\footnote{When $k$ is equal to $N$ in the equations in Section~\ref{sec:framework:evaluator}.} AP, AF, and INT are shown in Tables~\ref{tab:results_ap}-\ref{tab:results_int}, respectively.

\smallsection{AP \& AF.}
In Task-IL settings, Piggyback, PackNet, and PI-GNN, which are parameter-isolation-based methods, perform best overall in terms of both AP and AF. 
When we average ranks over all the Task-IL results (refer to Appendix~\ref{sec:app:add_results}), the average ranks of Piggyback, PackNet, and PI-GNN in terms of AP are $2.6$, $3$, and $3.5$, respectively. The method with the next highest rank is LwF, with an average rank of $4.25$.
\footnote{They perfectly retain knowledge from previous tasks (i.e., AF is 0) as they use parameters disjointly for different tasks.}
Among other methods, the regularization-based methods (i.e., LwF, EWC, MAS, and TWP) tend to perform better than the replay-based ones (i.e., GEM, ERGNN, and CaT) on Node Classification (NC). 

For the link-level and graph-level problems in the Class-IL setting, most CL methods perform only comparably with the Bare model.
Only GEM consistently outperforms the Bare model in both link-level and graph-level problems. 
For the node-level problem in the Class-IL setting, most CL methods outperform the Bare model.
Among them, CGNN and GEM perform best in terms of AP, while PI-GNN performs best in terms of AF. 

In Domain-IL settings, GEM, a replay-based method, outperforms all other methods in terms of AP and AF, only except for PI-GNN, which is limited only to two scenarios.
In Time-IL, no model outperforms the Bare model consistently on all datasets in terms of AP.
That is, it is challenging for current CL methods to deal with temporal dynamics in real-world graphs.

\smallsection{INT.}
In Task-IL, LwF shows overall the best performance among $12$ graph CL methods.
Specifically, when we average ranks over all the Task-IL results, LwF takes the first place with an average rank of $2.6$, and 
CaT takes the last place with an average rank of $12.2$.
Interestingly, the parameter-isolation-based methods, which are best in terms of AP and AF, are not best in terms of INT. The average ranks of PackNet, Piggyback, HAT, and PIGNN are $6.1$, $7.1$, $5.8$ and $7.0$, respectively.

When we average ranks across all graph problems and incremental settings (excluding parameter-isolation-based methods, which are not applicable), LwF (a regularization-based method) performs best with an average rank $2.3$, and 
CaT  (a replay-based method) performs worst with an average rank $8.2$.
However, overall, the performance difference between regularization-based and replay-based methods was not statistically significant within the 95\% confidence interval, 
as discussed in Appendix~\ref{sec:app:detailanalysis}.

Interestingly, in terms of INT, many CL methods are outperformed by the Bare model, which focuses only on learning the current task. 
In other words, the Bare model performs better on the current task than the outperformed CL methods, while it may suffer from catastrophic forgetting for past tasks.
This result implies that retaining knowledge from past tasks by the outperformed CL methods is less helpful (to the current task) than using all parameters to learn the current task.

We provide the full results on all scenarios and the detailed analysis in Appendix~\ref{sec:app:add_results} with a discussion on the FWT of the CL methods.
We also conduct hyperparameter sensitivity analyses and present the results in Appendix~\ref{sec:app:numberoftotaltask} and \ref{sec:app:pim}.

\subsection{Effects of the Number of Learned Tasks}
\label{sec:exp:numberoflearnedtask}
In Figure~\ref{fig:app_perfchange}, we report the performance curves, which show how the average performance (AP) (spec., AP at $k=n$) changes depending on the number of learned tasks $n$.
Note that most considered methods, including the Joint model, tend to perform worse, as the number of learned tasks increases, since they need to retain more previous knowledge.
However, opposite trends are observed on \arxiv and \molhiv, 
suggesting that they may have milder or more adaptable distribution shifts.
Interestingly, on \bitcoin under Task-IL, the performance of the Joint model and that of graph CL methods show different tendencies.
We provide the performance curves on all considered scenarios in Appendix~\ref{sec:app:add_results}.
\section{Conclusions and Future Directions}
\label{sec:conclusion}
In this work,
we define four incremental settings for evaluating continual learning methods for graph data (graph CL) based on four dimensions of changes, which are tasks, classes, domains, and time.
Then, we apply the settings to node-, link-, and graph-level learning problems. As a result, we provide \numscean benchmark scenarios from \numdata real-world datasets, which cover all $12$ combinations of the $4$ incremental settings and the $3$ levels of problems.
In addition, we propose \ourfw, a fool-proof and easy-to-use benchmark framework for the implementation and evaluation of graph CL methods.
Our benchmark results cover more extensive scenarios, CL methods, and evaluation metrics than the existing ones for graph CL, as summarized in Table~\ref{tab:coverage}. 
For reproducibility, we provide all source code required for reproducing the benchmark results and documents for users at \url{https://github.com/ShinhwanKang/BeGin}.

A limitation of our work is the relatively small number of tasks compared to CL benchmarks in other domains, primarily due to the lack of rich labels in graph data (with a maximum of $257$ in the datasets we considered), while image datasets like ILSVRC2012 often have $1000$ or more classes~\citep{lopez2017gradient,rebuffi2017icarl}.
Moreover, it is important to highlight that increasing the number of tasks for real-world graph datasets is not trivial without artificial manipulation of the datasets. 
Due to this difficulty, the number of tasks considered in previous studies of graph continual learning (ERGNN, CGNN, and CGLB) are only $41$, $9$, and $70$, respectively.
In the future, we plan to update our framework and benchmarks as datasets with richer labels become available, enabling a broader exploration of graph continual learning.

\begin{acks}
This work was supported by Institute of Information \& Communications Technology Planning \& Evaluation (IITP) grant funded by the Korea government (MSIT) (No. 2022-0-00157 / RS-2022-II220157, Robust, Fair, Extensible Data-Centric Continual Learning) 
(No. RS-2024-00438638, EntireDB2AI: Foundations and Software for Comprehensive Deep Representation Learning and Prediction on Entire Relational Databases)
(No. 2019-0-00075 / RS-2019-II190075, Artificial Intelligence Graduate School Program (KAIST)).
\end{acks}

\bibliographystyle{ACM-Reference-Format}
\bibliography{reference}

\appendix
\begin{appendices}
\setcounter{page}{1}

\section{Illustration of a Time-Evolving Graph}
\label{sec:app:intro_detail}

\begin{figure}[ht]
    \vspace{-2mm}
    \centering
    \includegraphics[width=1.\linewidth]{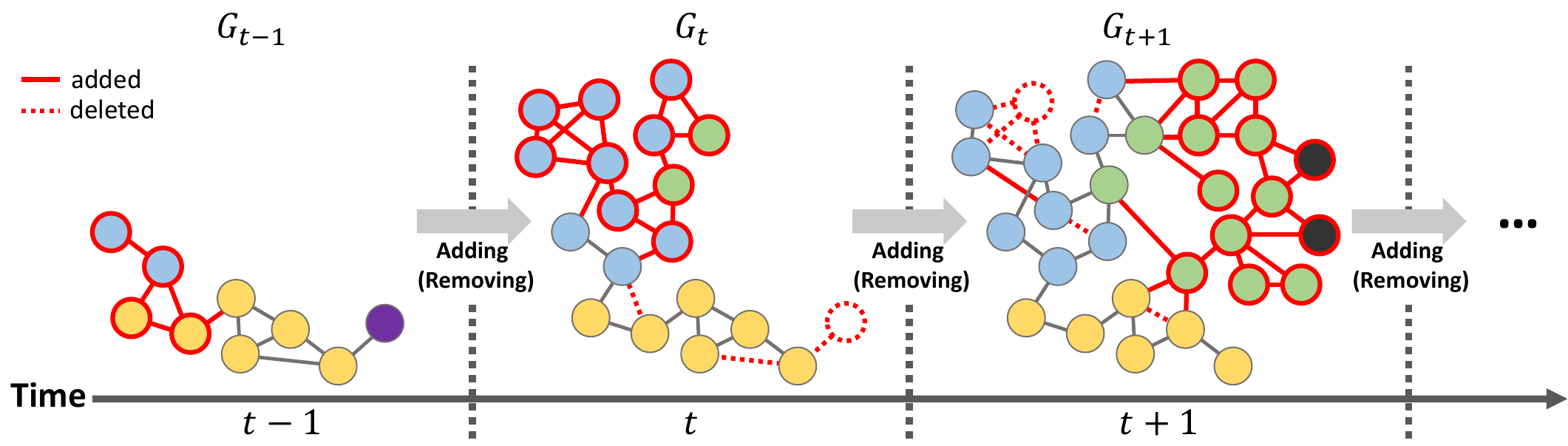}
    \caption{\textbf{Illustration of a time-evolving graph.} The colors of nodes represent their classes (or domains). New nodes and edges are depicted using solid red circles and lines, while removed nodes and edges are depicted using dotted red circles and lines.
    It is important to note that over time, there are changes in (a) the number of nodes, (b) the number of edges, (c) the number of classes (or domains), and (d) the distribution over classes.
    }
    \label{fig:app:graph}
\end{figure}

Figure~\ref{fig:app:graph} provides an example illustration of a time-evolving graph. It demonstrates that both nodes and edges can appear and disappear over time. Additionally, novel classes (or domains) may emerge and subsequently vanish, causing the distribution over classes to shift throughout the timeline.

\begin{itemize}[leftmargin=*]
\item At time $t-1$, the graph consists of $9$ nodes and $11$ edges. There are three classes (or domains) present in the graph: blue, orange, and purple. When observing the graph $G_{t-1}$, we notice that the most prominent class is orange, followed by blue and purple.
\item At time $t$, $10$ new nodes emerge, accompanied by $15$ incident edges. Notably, the node belonging to the purple class disappears, and two existing edges vanish. Additionally, a new class, represented by the color green, emerges in the graph. As a result of these changes, the blue class becomes the most dominant class.
\item At time $t+1$, $10$ new nodes belonging to the green class and $2$ new nodes belonging to a novel class, represented by the color black, emerge, accompanied by $22$ incident edges.
Moreover, an existing node and its incident edges are removed.
Additionally, a new edge is added among existing nodes, while three existing edges are removed.
As a result, the graph now consists of $29$ nodes, $40$ edges, and $4$ classes.
Furthermore, the dominance of the blue class diminishes, while the green class emerges as the new dominant class.
\end{itemize}

\section{Example: Implementing Elastic Weight Consolidation (EWC) using \ourfw}
\label{sec:app:ewc}
In this section, to demonstrate the usability of \ourfw, we provide a step-by-step example of implementing Elastic Weight Consolidation (EWC)~\citep{kirkpatrick2017overcoming} for node classification.

EWC uses the weighted L2 penalty term based on the learned weights and fisher information matrices from the previous tasks. The loss function $\mathcal{L}(\cdot)$ for EWC is as follows:
\begin{align}
\mathcal{L}(\Theta) = \mathcal{L}_i(\Theta) + \sum_{j=1}^{i-1} \frac{\lambda}{2} F_j(\Theta - \Theta^{*}_j)^2, \label{eq:EWC}
\end{align}
where $\Theta$ is the parameter of the model, $\mathcal{L}_i$ is the loss for the current task $i$, $\lambda$ is the hyperparameter for the L2 penalty term, $\Theta^{*}_j$ is the learned weights at task $j$, and $F_j$ is the fisher information matrix calculated from $\Theta^{*}_j$.

\smallsection{Step 1. Extending the base trainer.} 
For each problem considered in this paper, \method provides a \textit{base} trainer class that makes the training behavior exactly the same as the Bare model (i.e., basic incremental learning without any specific techniques for preventing catastrophic forgetting; refer to Section~\ref{sec:exp:setting}). 
Based on the base class, users can implement their CL algorithm by extending the class and substituting some default event functions with user-defined ones.

\begin{lstlisting}
for begin.trainers.nodes import NCTrainer

# Step 1. Extending the base trainer
class NCClassILEWCTrainer(NCTrainer):
    pass
\end{lstlisting}

\smallsection{Step 2. Initializing algorithm-specific states.} As stated in Eq.~\eqref{eq:EWC}, EWC requires storing the learned weights and Fisher information matrices from the previous tasks to compute the regularization term. However, they cannot be obtained on the current task. In order to resolve this issue, the trainer provides a dictionary called \texttt{training\_states}. The dictionary can be used to store intermediate results and can be shared by events in the form of an argument (i.e., input parameter) of the event functions. To set the initial states, the user can extend the base trainer with their modified \texttt{initTrainingStates()} event function, which initializes the states for running EWC.

\begin{lstlisting}
for begin.trainers.nodes import NCTrainer

class NCClassILEWCTrainer(NCTrainer):

    # Called only once, when the training procedure begins
    def initTrainingStates(self, model, optimizer):

        # Step 2. Initializing algorithm-specific states
        return {'fishers': [], 'params': []}
\end{lstlisting}

\smallsection{Step 3. Computing and storing the Fisher information matrix.} In order to compute the penalty term at task $i$, we need the learned weights $\Theta^{*}_j$ and the Fisher information matrix $F_j$ for every task $j < i$. Hence, we need to store them at the end of each task, and this can naturally be implemented in the event function \texttt{processAfterTraining()}, which is called at the end of each task. In the example below, \texttt{curr\_training\_states[`params'][j-1]} and \texttt{curr\_training\_states[`fishers'][j-1]} store the learned weights and the Fisher information matrix of task $j$, respectively. 

\begin{lstlisting}
class NCClassILEWCTrainer(NCTrainer):

    # (Continued from the previous code)

    # Called once for each task when the trainer completes training for the current task
    def processAfterTraining(self, task_id, curr_dataset, curr_model, curr_optimizer, curr_training_states):
        super().processAfterTraining(task_id, curr_dataset, curr_model, curr_optimizer, curr_training_states)

        # Step 3-1. Loading the parameters
        params = {name: torch.zeros_like(p) for name, p in curr_model.named_parameters()}
        fishers = {name: torch.zeros_like(p) for name, p in curr_model.named_parameters()}
        train_loader = self.prepareLoader(curr_dataset, curr_training_states)[0]

        # Step 3-2. Computing the Fisher information matrix
        total_num_items = 0
        for i, _curr_batch in enumerate(iter(train_loader)):
            curr_model.zero_grad()
            curr_results = self.inference(curr_model, _curr_batch, curr_training_states)
            curr_results['loss'].backward()
            curr_num_items =_curr_batch[1].shape[0]
            total_num_items += curr_num_items
            for name, p in curr_model.named_parameters():
                params[name] = p.data.clone().detach()
                fishers[name] += (p.grad.data.clone().detach() ** 2) * curr_num_items

        for name, p in curr_model.named_parameters():
            fishers[name] /= total_num_items

        # Step 3-3. Storing the paramters and the Fisher information matrix
        curr_training_states['fishers'].append(fishers)
        curr_training_states['params'].append(params)
\end{lstlisting}

\smallsection{Step 4. Computing $\mathcal{L}(\Theta)$ for regularization.} The penalty term in Eq.~\eqref{eq:EWC} is used for regularization during backpropagation. Thus, it should be computed during training, i.e., within the event function \texttt{processTrainIteration()}.
In our implementation below, we modify the event function \texttt{afterInference()}, which is called within the default implementation of \texttt{processTrainIteration()} after an inference step (i.e., a forward pass).
In the event function, the argument (i.e., input parameter) \texttt{curr\_training\_states} contains the Fisher information matrices and the previously learned parameters, which the penalty term \texttt{loss\_reg} is computed based on.
The event function also has the argument  \texttt{results}, which contains the prediction result and the loss of the current model computed in the \texttt{inference()} function. 
Thus, the overall loss including the penalty term can be obtained by summing up \texttt{results[`loss']} and \texttt{loss\_reg}.

\begin{lstlisting}
class NCClassILEWCTrainer(NCTrainer):

    # (Continued from the previous code)

    # Called right after an inference step (i.e., a forward pass) terminates within 
    # the default implementation of processTrainIteration()
    def afterInference(self, results, model, optimizer, _curr_batch, curr_training_states):
        
        # Step 4. Adding the penalty term to the loss function
        loss_reg = 0
        for _param, _fisher in zip(curr_training_states['params'], curr_training_states['fishers']):
            for name, p in model.named_parameters():
                l = self.lamb * _fisher[name]
                l = l * ((p - _param[name]) ** 2)
                loss_reg = loss_reg + l.sum()
        total_loss = results['loss'] + loss_reg
        
        total_loss.backward()
        optimizer.step()

        return {'loss': total_loss.item(), 'acc': self.eval_fn(results['preds'].argmax(-1), _curr_batch[0].ndata['label'][_curr_batch[1]])}
\end{lstlisting}

\smallsection{Summary.}
The above code presents the complete implementation of EWC for node classification under Class-IL. 
This example clearly demonstrates the usability of \ourfw in several ways:
\begin{itemize}[leftmargin=*]
    \item  
    CL methods can be implemented independently from evaluation code (e.g., data splitting and evaluation metric computation), which can be complex in graph CL. CL methods only need to respond to given queries.
    \item Due to the modularized structure of \method, CL methods can be implemented without being tied to dataset specifics or model architecture (e.g., graph neural networks) specifics.
    \item CL methods can be implemented in a small number of lines of code by adjusting or extending event functions (see Appendix~\ref{sec:app:eventfunctions} for the full list of event functions).
\end{itemize}
Similarly, various CL algorithms can be effortlessly and succinctly implemented with \ourfw, and their code is available at \cite{code}.

\section{Supported Event Functions in \ourfw}
\label{sec:app:eventfunctions}

If a user-defined event function is not provided, the trainer performs training and evaluation with the corresponding basic pre-implemented operations. Thus, users do not need to implement the whole training and evaluation procedure, but only the necessary parts. Currently, users can override the following built-in event functions: 
\begin{itemize}[leftmargin=*]
    \item \texttt{initTrainingStates()}: This function is called only once, when the training procedure begins.
    \item \texttt{preparePretraingLoader()}: This function is called once before the main training step, and if users opt for a pretraining phase, it should return the dataloaders needed for pretraining.
    \item \texttt{processPretraining()}: This function is called once right after the \texttt{preparePretrainingLoader()} event terminates, and if users opt for a pretraining phase, it should handle the pretraining process.
    \item \texttt{prepareLoader()}: This function is called once for each task when generating dataloaders for training, validation, and test. Given the dataset for each task, it should return dataloaders for training, validation, and test.
    \item \texttt{processBeforeTraining()}: This function is called once for each task, right after the \texttt{prepareLoader()} event function terminates.
    \item \texttt{processTrainIteration()}: This function is called for every training iteration. When the current batched inputs, model, and optimizer are given, it should perform a single training iteration and return the information or outcome during the iteration.
    \item \texttt{processEvalIteration()}: This function is called for every evaluation iteration. When the current batched inputs and trained model are given, it should perform a single evaluation iteration and return the information or outcome during the iteration.
    \item \texttt{inference()}: This function is called for every inference step in the training procedure.
    \item \texttt{beforeInference()}: This function is called right before  \texttt{inference()} begins.
    \item \texttt{afterInference()}: This function is called right after  \texttt{inference()} terminates.
    \item \texttt{\_reduceTrainingStats()}: This function is called at the end of every training step. Given the returned values of the \texttt{processTrainIteration()} event function, it should return overall and reduced statistics of the current training step.
    \item \texttt{\_reduceEvalStats()}: This function is called at the end of every evaluation step. Given the returned values of the \texttt{processEvalIteration()} event function, it should return overall and reduced statistics of the current evaluation step.
    \item \texttt{processTrainingLogs()}: This function is called right after the \texttt{reduceTrainingStats()} event function terminates.
    \item \texttt{processAfterEachIteration()}: This function is called at the end of the training iteration. When the outcome from \texttt{reduceTrainingStats()} and \texttt{reduceEvalStats()} are given, it should determine whether the trainer should stop training for the current task or not.
    \item \texttt{processAfterTraining()}: This function is called once for each task when the trainer completes training for the current task.
\end{itemize}
The arguments and detailed roles of these event functions are provided in the official document of \method, which is available at \citep{code}.

\section{Full Experiment Results}
\label{sec:app:add_results}

\begin{figure}[ht]
    \centering
    \includegraphics[width=0.83\textwidth]{figure/plots/legend.pdf}
    \begin{subfigure}[b]{0.242\textwidth}
         \centering
         \includegraphics[width=\textwidth]{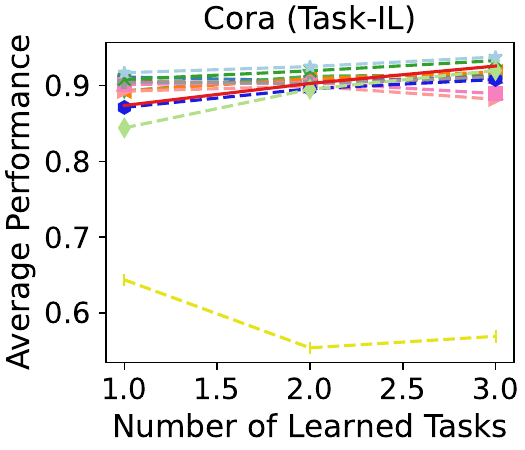}
         \caption{\makecell{\cora \\ (NC, Task-IL)}}
    \end{subfigure}
    \begin{subfigure}[b]{0.242\textwidth}
         \centering
         \includegraphics[width=\textwidth]{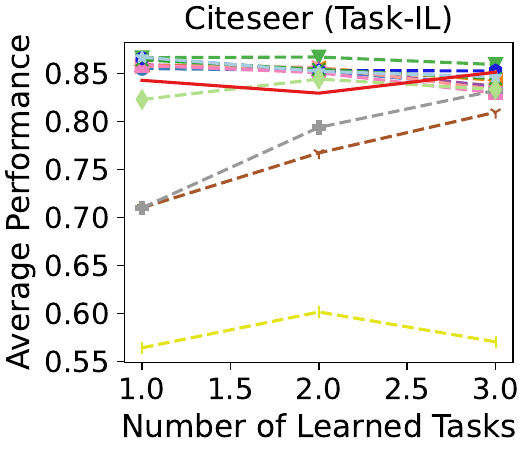}
         \caption{\makecell{\citeseer \\ (NC, Task-IL)}}
    \end{subfigure}
    \begin{subfigure}[b]{0.242\textwidth}
         \centering
         \includegraphics[width=\textwidth]{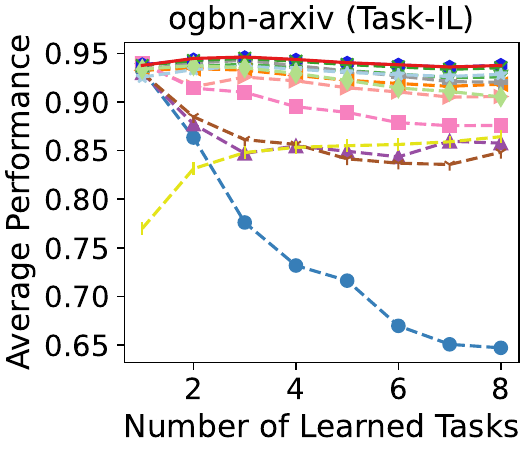}
         \caption{\makecell{\arxiv \\ (NC, Task-IL)}}
    \end{subfigure}
    \begin{subfigure}[b]{0.242\textwidth}
         \centering
         \includegraphics[width=\textwidth]{figure/plots/node_task_corafull.pdf}
         \caption{\makecell{\corafull \\ (NC, Task-IL)}}
    \end{subfigure}
    \begin{subfigure}[b]{0.242\textwidth}
         \centering
         \includegraphics[width=\textwidth]{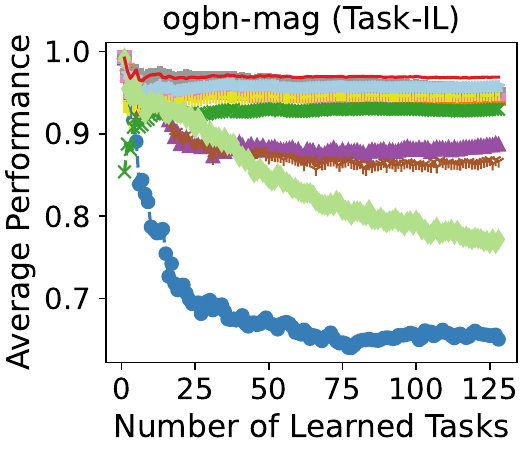}
         \caption{\makecell{\magdata \\ (NC, Task-IL)}}
    \end{subfigure}
    \begin{subfigure}[b]{0.242\textwidth}
         \centering
         \includegraphics[width=\textwidth]{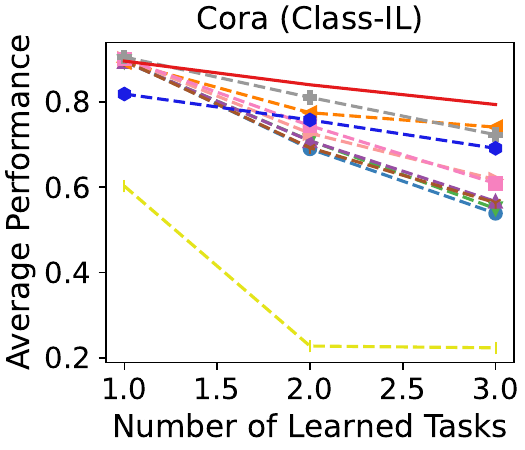}
         \caption{\makecell{\cora \\ (NC, Class-IL)}}
    \end{subfigure}
    \begin{subfigure}[b]{0.242\textwidth}
         \centering
         \includegraphics[width=\textwidth]{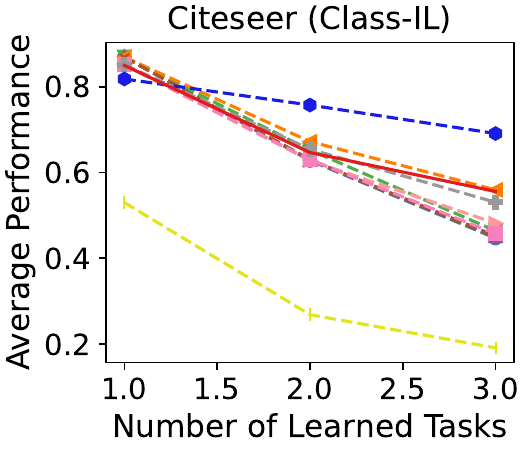}
         \caption{\makecell{\citeseer \\ (NC, Class-IL)}}
    \end{subfigure}
    \begin{subfigure}[b]{0.242\textwidth}
         \centering
         \includegraphics[width=\textwidth]{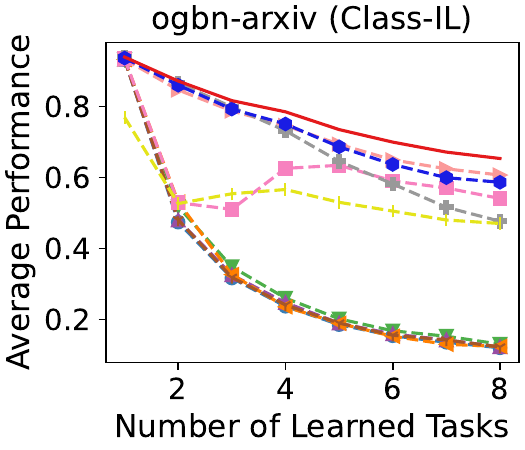}
         \caption{\makecell{\arxiv \\ (NC, Class-IL)}}
    \end{subfigure}
    \begin{subfigure}[b]{0.242\textwidth}
         \centering
         \includegraphics[width=\textwidth]{figure/plots/node_class_ogbn-products.pdf}
         \caption{\makecell{\products \\ (NC, Class-IL)}}
    \end{subfigure}
    \begin{subfigure}[b]{0.242\textwidth}
         \centering
         \includegraphics[width=\textwidth]{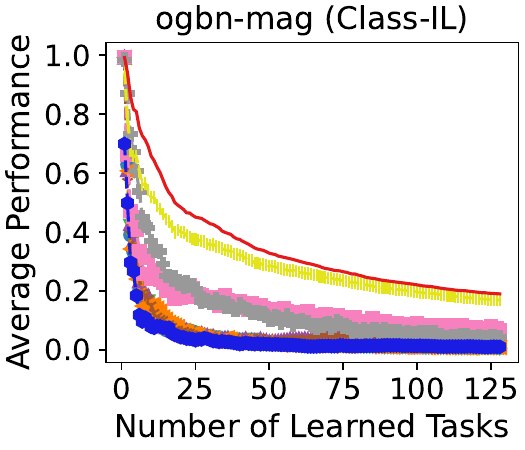}
         \caption{\makecell{\magdata \\ (NC, Class-IL)}}
    \end{subfigure}
    \begin{subfigure}[b]{0.242\textwidth}
         \centering
         \includegraphics[width=\textwidth]{figure/plots/node_domain_ogbn-proteins.pdf}
         \caption{\makecell{\proteins
         \\ (NC, Domain-IL)}}
    \end{subfigure}
    \begin{subfigure}[b]{0.242\textwidth}
         \centering
         \includegraphics[width=\textwidth]{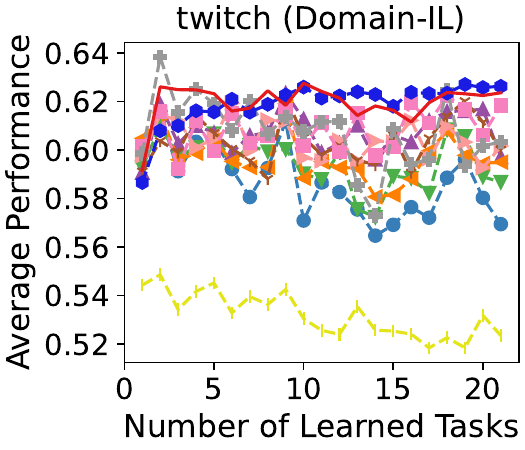}
         \caption{\makecell{\twitch \\ (NC, Domain-IL)}}
    \end{subfigure}
    \begin{subfigure}[b]{0.242\textwidth}
         \centering
         \includegraphics[width=\textwidth]{figure/plots/node_time_ogbn-arxiv.pdf}
         \caption{\makecell{\arxiv \\ (NC, Time-IL)}}
    \end{subfigure}
    \begin{subfigure}[b]{0.242\textwidth}
         \centering
         \includegraphics[width=\textwidth]{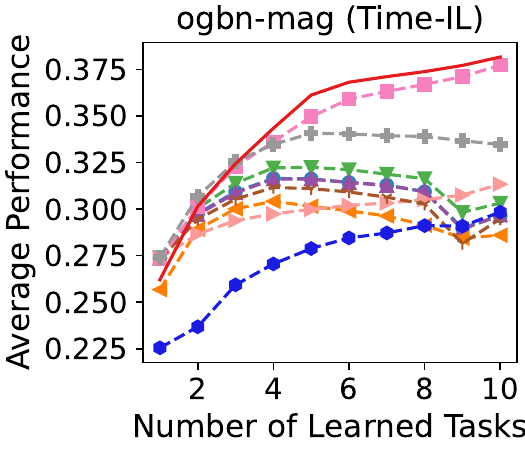}
         \caption{\makecell{\magdata \\ (NC, Time-IL)}}
    \end{subfigure}
    \caption{\textbf{Change of Average Performance (AP) during continual learning (for NC).}}
    \label{fig:app_perfchange_nc}
\end{figure}

\begin{figure}[ht]
    \centering
    \includegraphics[width=0.83\textwidth]{figure/plots/legend.pdf}
    \begin{subfigure}[b]{0.242\textwidth}
         \centering
         \includegraphics[width=\textwidth]{figure/plots/lc_task_bitcoin-otc.pdf}
         \caption{\makecell{\bitcoin \\ (LC, Task-IL)}}
     \end{subfigure}
     \begin{subfigure}[b]{0.242\textwidth}
         \centering
         \includegraphics[width=\textwidth]{figure/plots/lc_class_bitcoin-otc.pdf}
         \caption{\makecell{\bitcoin \\ (LC, Class-IL)}}
     \end{subfigure}
     \begin{subfigure}[b]{0.242\textwidth}
         \centering
         \includegraphics[width=\textwidth]{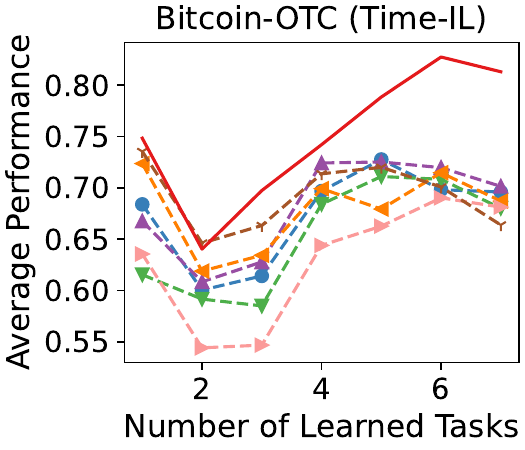}
         \caption{\makecell{\bitcoin \\ (LC, Time-IL)}}
     \end{subfigure}
     \\
     \begin{subfigure}[b]{0.242\textwidth}
         \centering
         \includegraphics[width=\textwidth]{figure/plots/lp_domain_wiki-cs.pdf}
         \caption{\makecell{\wikics \\ (LP, Domain-IL)}}
     \end{subfigure}
     \begin{subfigure}[b]{0.242\textwidth}
         \centering
         \includegraphics[width=\textwidth]{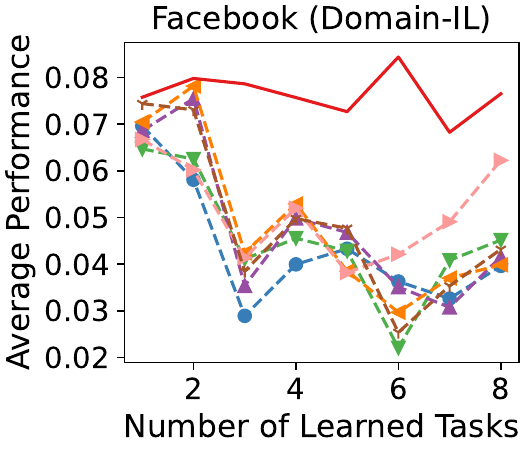}
         \caption{\makecell{\facebook \\ (LP, Domain-IL)}}
     \end{subfigure}
     \begin{subfigure}[b]{0.242\textwidth}
         \centering
         \includegraphics[width=\textwidth]{figure/plots/lp_time_ogbl-collab.pdf}
         \caption{\makecell{\collab \\ (LP, Time-IL)}}
     \end{subfigure}
     \\
     \begin{subfigure}[b]{0.242\textwidth}
         \centering
         \includegraphics[width=\textwidth]{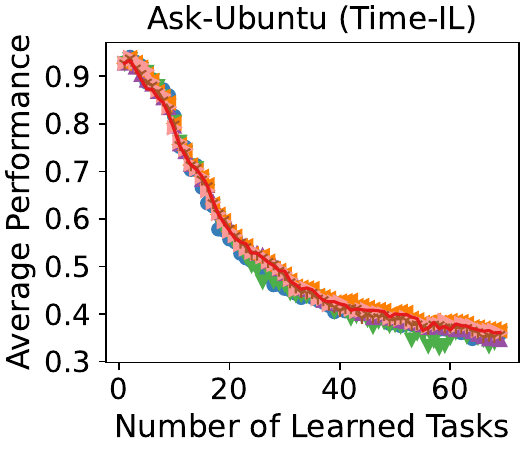}
         \caption{\makecell{\askubuntu \\ (LP, Time-IL)}}
     \end{subfigure}
     \begin{subfigure}[b]{0.242\textwidth}
         \centering
         \includegraphics[width=\textwidth]{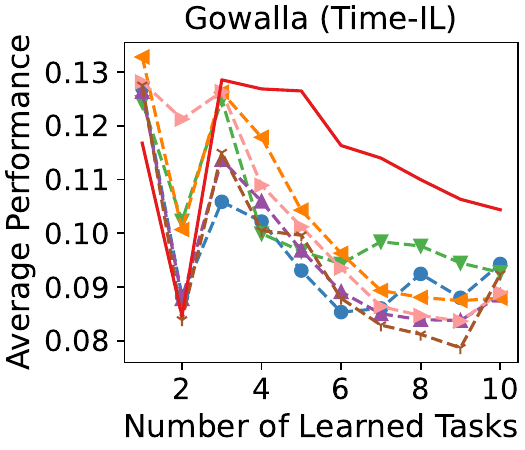}
         \caption{\makecell{\gowalla} \\ (LP, Time-IL)}
     \end{subfigure}
     \begin{subfigure}[b]{0.242\textwidth}
         \centering
         \includegraphics[width=\textwidth]{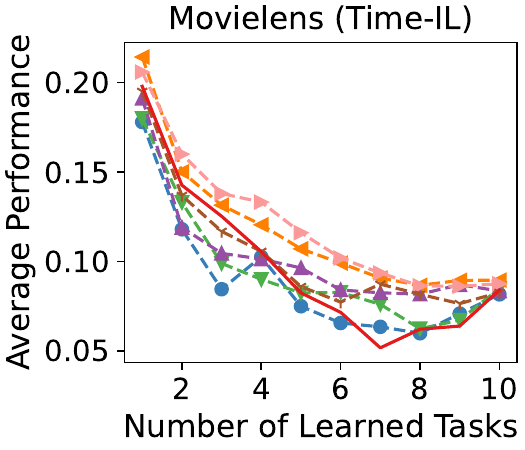}
         \caption{\makecell{\movie \\ (LP, Time-IL)}}
     \end{subfigure}
    \caption{\textbf{Change of Average Performance (AP) during continual learning (for LC and LP).}}
    \label{fig:app_perfchange_lclp}
\end{figure}

\begin{figure}[h]
    \centering
    \includegraphics[width=0.83\textwidth]{figure/plots/legend.pdf}
    \begin{subfigure}[b]{0.242\textwidth}
         \centering
         \includegraphics[width=\textwidth]{figure/plots/graph_task_mnist.pdf}
         \caption{\makecell{\mnist \\ (GC, Task-IL)}}
    \end{subfigure}
    \begin{subfigure}[b]{0.242\textwidth}
         \centering
         \includegraphics[width=\textwidth]{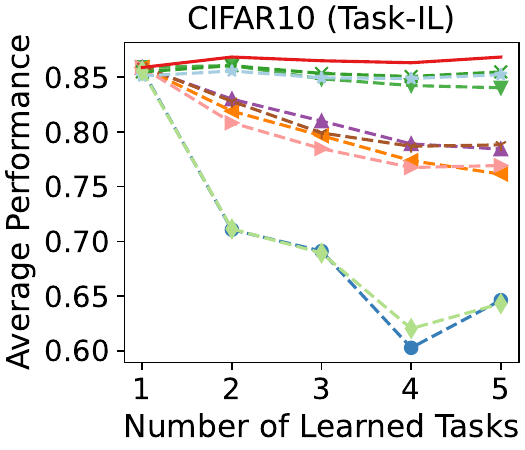}
         \caption{\makecell{\cifar \\ (GC, Task-IL)}}
    \end{subfigure}
    \begin{subfigure}[b]{0.242\textwidth}
         \centering
         \includegraphics[width=\textwidth]{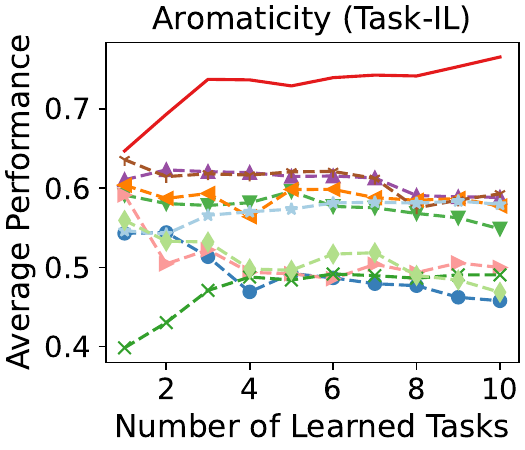}
         \caption{\makecell{\aroma \\ (GC, Task-IL)}}
    \end{subfigure}
    \begin{subfigure}[b]{0.242\textwidth}
         \centering
         \includegraphics[width=\textwidth]{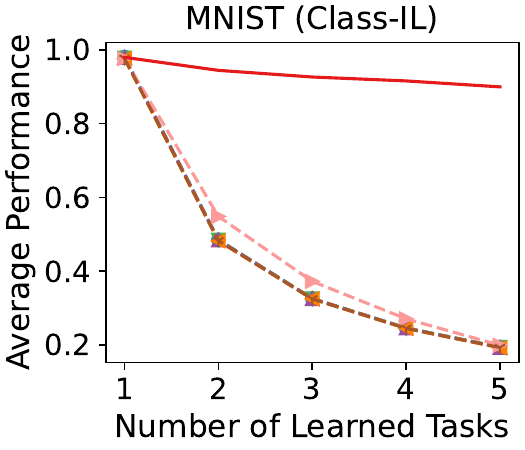}
         \caption{\makecell{\mnist \\ (GC, Class-IL)}}
    \end{subfigure}
    \\
    \begin{subfigure}[b]{0.242\textwidth}
         \centering
         \includegraphics[width=\textwidth]{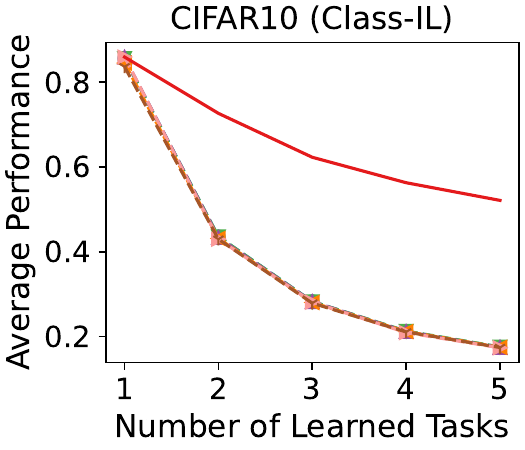}
         \caption{\makecell{\cifar \\ (GC, Class-IL)}}
    \end{subfigure}
    \begin{subfigure}[b]{0.242\textwidth}
         \centering
         \includegraphics[width=\textwidth]{figure/plots/graph_class_aromaticity.pdf}
         \caption{\makecell{\aroma \\ (GC, Class-IL)}}
    \end{subfigure}
    \begin{subfigure}[b]{0.242\textwidth}
         \centering
         \includegraphics[width=\textwidth]{figure/plots/graph_domain_ogbg-molhiv.pdf}
         \caption{\makecell{\molhiv \\ (GC, Domain-IL)}}
    \end{subfigure}
    \begin{subfigure}[b]{0.242\textwidth}
         \centering
         \includegraphics[width=\textwidth]{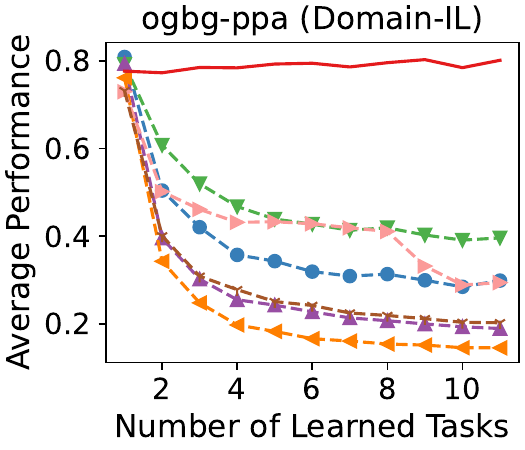}
         \caption{\makecell{\ppa \\ (GC, Domain-IL)}}
    \end{subfigure}
    \\
    \begin{subfigure}[b]{0.242\textwidth}
         \centering
         \includegraphics[width=\textwidth]{figure/plots/graph_time_nyc-taxi.pdf}
         \caption{\makecell{\nyctaxi \\ (GC, Time-IL)}}
    \end{subfigure}
    \begin{subfigure}[b]{0.242\textwidth}
         \centering
         \includegraphics[width=\textwidth]{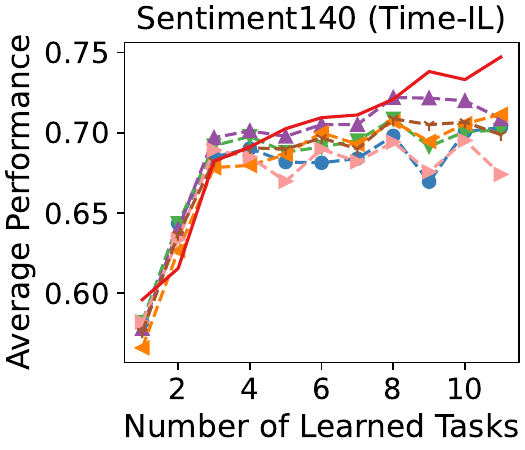}
         \caption{\makecell{\sentiment \\ (GC, Time-IL)}}
    \end{subfigure}
    \begin{subfigure}[b]{0.242\textwidth}
         \centering
         \includegraphics[width=\textwidth]{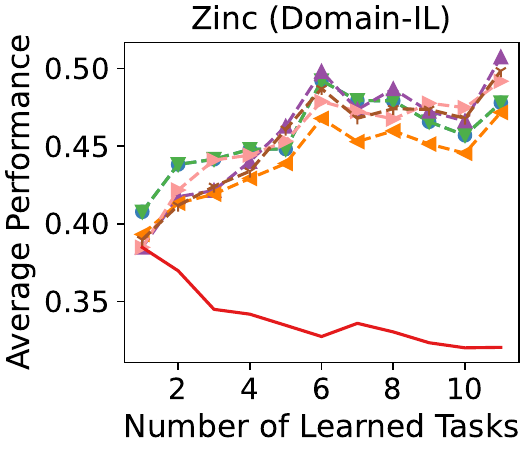}
         \caption{\makecell{\zinc \\ (GR, Domain-IL)}}
    \end{subfigure}
    \begin{subfigure}[b]{0.242\textwidth}
         \centering
         \includegraphics[width=\textwidth]{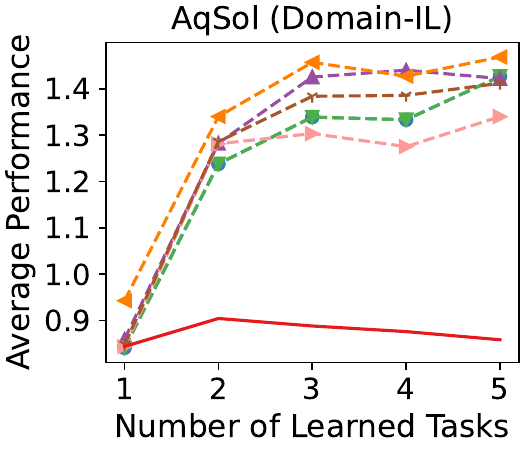}
         \caption{\makecell{\aqsol \\ (GR, Domain-IL)}}
    \end{subfigure}
    \caption{\textbf{Change of Average Performance (AP) during continual learning (for GC and GR).}}
    
    \label{fig:app_perfchange_gc}
\end{figure}

We report the experimental results for all \numscean benchmark scenarios.
Tables~\ref{tab:results_ap_full}-\ref{tab:results_fwt} show average performance (AP), average forgetting (AF), Intransigence (INT), and forward transfer (FWT), respectively.
Figures~\ref{fig:app_perfchange_nc}-\ref{fig:app_perfchange_gc} show the changes in AP with respect to the number of learned tasks for node-, link-, and graph-level problems, respectively.

\subsection{Detailed Analysis of Node Classification Performance}
\label{sec:app:detailanalysis}
We conducted an analysis to assess the category-based performance differences across various metrics and incremental settings, specifically focusing on Node Classification (NC) problems. To compare these differences, we calculated the rank of each method across all benchmark scenarios, with the exception of the \magdata dataset\footnote{Note that GEM ran out of time in \magdata on both Task-IL and Class-IL settings.} and CaT.\footnote{CaT was excluded from the analysis as it was outperformed even by the Bare algorithm, particularly when the condensed graph was small. We suspect that this is due to the condensation process rather than the replay-based mechanism.} Subsequently, we visualized the distributions of the ranks for each category using box plots.

\begin{figure}[t!]
    \centering
    \begin{subfigure}[b]{\textwidth}
         \centering
         \includegraphics[width=0.71\linewidth]{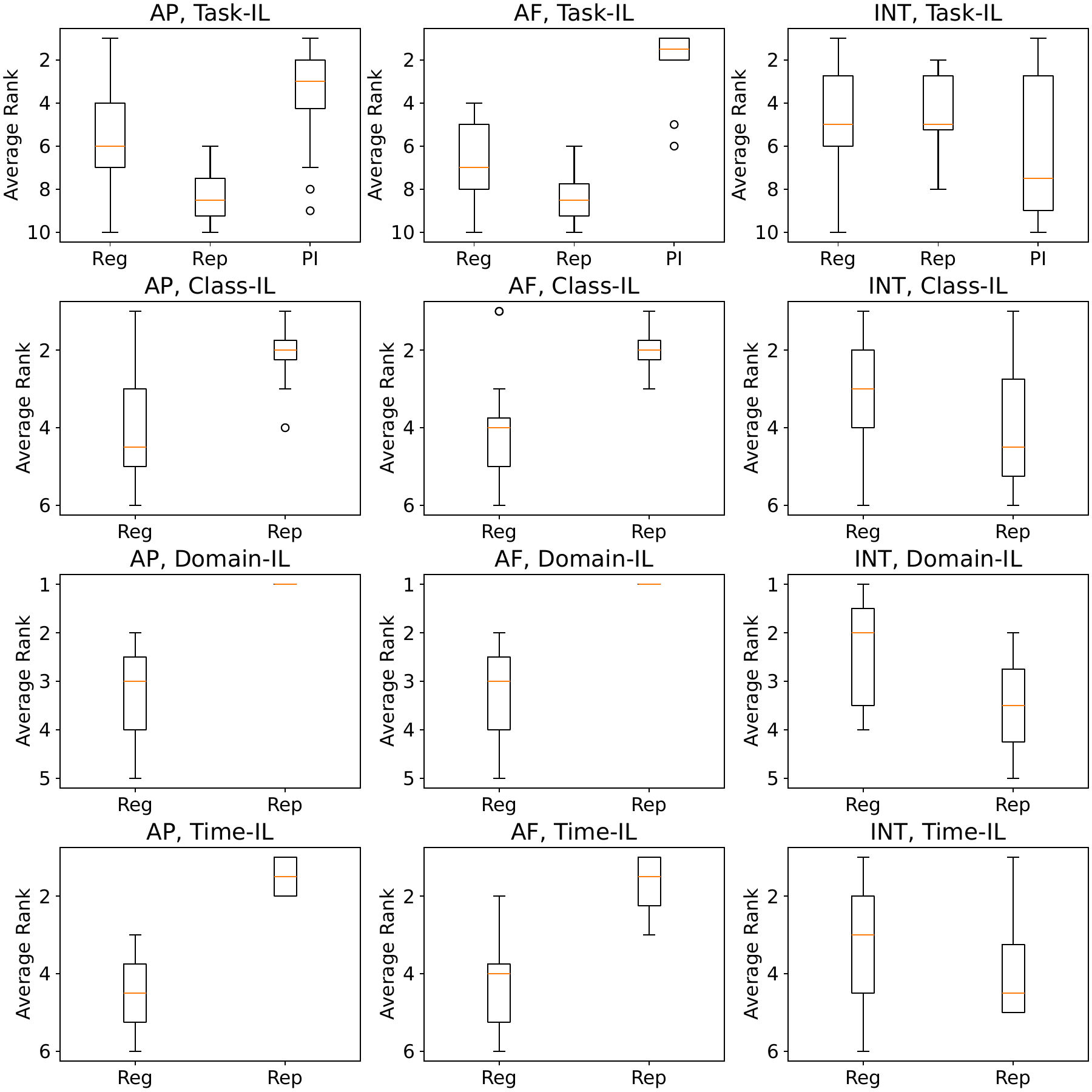}
         \caption{Box plot}
    \end{subfigure}
    \begin{subfigure}[b]{\textwidth}
        \centering
        \scalebox{0.7}{
        \begin{tabular}{c|c|cc|cc}
        \toprule
        Final Metric & Incremental Setting & First Distribution & Second Distribution & t-statistic & p-value \\
        \midrule
        \multirow{5}{*}{\makecell{ \\ AP}} & \multirow{3}{*}{Task-IL} & \textbf{Regularization-based} & Replay-based & -2.9043 & \bbf{0.0087} \\
        & & Regularization-based & \textbf{Parameter-isolation-based} & 2.7625 & \bbf{0.0095} \\
        & & Replay-based & \textbf{Parameter-isolation-based} & 5.6605 & \textbf{1.4002e-05} \\
        \cmidrule{2-6}
        & Class-IL & Regularization-based & \textbf{Replay-based} & 3.9505 & \bbf{0.0008} \\
        \cmidrule{2-6}
        & Time-IL & Regularization-based & \textbf{Replay-based} & 5.8621 & \bbf{0.0002} \\
        \midrule \midrule
        \multirow{5}{*}{\makecell{ \\ AF}} & \multirow{3}{*}{Task-IL} & \textbf{Regularization-based} & Replay-based & -2.2484 & \bbf{0.0368} \\
        & & Regularization-based & \textbf{Parameter-isolation-based} & 7.9886 & \textbf{1.0467e-08} \\
        & & Replay-based & \textbf{Parameter-isolation-based} & 10.3749 & \textbf{3.5271e-08} \\
        \cmidrule{2-6}
        & Class-IL & Regularization-based & \textbf{Replay-based} & 4.5265 & \bbf{0.0002} \\
        \cmidrule{2-6}
        & Time-IL & Regularization-based & \textbf{Replay-based} & 3.6457 & \bbf{0.0058} \\
        \midrule \midrule
        \multirow{5}{*}{\makecell{ \\ INT}} & \multirow{3}{*}{Task-IL} & Regularization-based & Replay-based & 0.2567 & 0.8004 \\
        & & Regularization-based & Parameter-isolation-based & -1.3228 & 0.1968 \\
        & & Replay-based & Parameter-isolation-based & -1.4776 & 0.1542 \\
        \cmidrule{2-6}
        & Class-IL & Regularization-based & Replay-based & -1.0515 & 0.3127 \\
        \cmidrule{2-6}
        & Time-IL & Regularization-based & Replay-based & -0.3261 & 0.7555 \\
        \bottomrule
        \end{tabular}
        }
        \caption{Welch's t-test result}
    \end{subfigure}
    \caption{\textbf{Detailed analysis of node classification (NC) performance in terms of AP, AF, and INT.} In the figure, Reg. represents the regularization-based methods (LwF, EWC, MAS, and TWP), Rep. represents the replay-based methods (GEM and ERGNN), and PI represents the parameter-isolation-based methods (PackNet, Piggyback, HAT, and PI-GNN). By performing Welch's t-test, we demonstrate that the category-based differences in terms of AP and AF are statistically significant within the 95\% confidence interval. We highlighted the $p$-values and outperforming distributions in bold if the difference between the two distributions of ranks is statistically significant.}
    \label{fig:app:detailed_analysis}
\end{figure}

As shown in Figure \ref{fig:app:detailed_analysis}, parameter-isolation-based methods outperformed the regularization- and replay-based methods in Task-IL. Furthermore, we observed that replay-based methods generally achieved better overall performance compared to regularization-based methods in all incremental settings, except for Task-IL, in terms of AP and AF. Additionally, by conducting Welch's t-test, we discovered that the category-based performance differences we reported were statistically significant within the 95\% confidence interval. The following intuition can help explain these experimental results:
\begin{itemize}
    \item \textbf{Class distribution shifts are more challenging in Class-IL and Time-IL settings compared to Task-IL settings:}
    In the Task-IL setting, a model can assign (partially) separate classifiers for each task since the given data explicitly indicates which task each query belongs to. This approach helps mitigate class distribution shifts across tasks effectively.
    In contrast, in the Class-IL setting, the task associated with each query is not provided, requiring a single classifier to handle all tasks. 
    This makes dealing with class distribution shifts more challenging, requiring the classifier to allocate probabilities not only to seen classes in previous tasks but also to unseen classes.
    A similar challenge arises in Time-IL settings, where class distributions shift over time, with new classes emerging, as illustrated in Figure~\ref{fig:int:increasing}.
    \item \textbf{Why replay-based methods are superior in Class-IL and Time-IL settings}:
    Replay-based methods are trained explicitly using data sampled from previous tasks, allowing the learned models to undergo a smaller shift in class distribution, regardless of the incremental learning setting.
    In contrast, regularization-based methods rely solely on implicit constraints without explicitly revisiting prior data. As a result, replay-based methods outperform regularization-based methods in Class-IL and Time-IL settings, where dealing with class distribution shifts is more challenging.
    \item \textbf{Why parameter-isolation-based methods are superior in Task-IL settings}: 
    By their mechanisms, parameter-isolation-based methods (except for HAT) ensure no forgetting in Task-IL settings, allowing them to outperform replay- and regularization-based methods in the settings.
    \item \textbf{Why regularization-based methods are better than replay-based methods in Task-IL settings}: 
    In contrast to Class-IL settings, class distribution shifts are easier to mitigate in Task-IL settings, as discussed above. Consequently, the advantage of replay-based methods, which address these shifts by revisiting prior data, is diminished. Instead, regularization-based methods, which minimize parameter changes to preserve prior knowledge, are better suited for this scenario. 
\end{itemize}

\begin{table*}[t!]
\caption{\textbf{Performance in terms of Average Performance (AP, the higher, the better)}. In each setting, the best score is in bold, and the second-best score is underlined. Each number is rescaled to a range of 100. O.O.M: out of memory. O.O.T: out of time ($>$ 24 hours). N/A: not applicable to the problems or scenarios.}
\label{tab:results_ap_full}
\setlength{\tabcolsep}{1pt}
\renewcommand{\arraystretch}{1.1}
\centering
\begin{subtable}[]{\textwidth}
\centering
\scalebox{0.65}{
\begin{tabular}{c|ccccccccccc}
\toprule
Problem & \multicolumn{11}{c}{Node Classification (NC)} \\
\midrule
Methods
& \makecell{\cora \\ (Task-IL)} & \makecell{\citeseer \\ (Task-IL)} & \makecell{\arxiv \\ (Task-IL)} & \makecell{\corafull \\ (Task-IL)} & \makecell{\magdata \\ (Task-IL)} & \makecell{\cora \\ (Class-IL)} & \makecell{\citeseer \\ (Class-IL)} & \makecell{\arxiv \\ (Class-IL)} & \makecell{\products \\ (Class-IL)} & \makecell{\magdata \\ (Class-IL)} & \makecell{\proteins \\ (Domain-IL)} \\
\midrule
Bare & 90.3±1.8 & 83.6±2.9 & 65.0±7.6 & 75.9±1.9 & 65.1±1.6 & 54.1±3.6 & 44.7±4.0 & 12.0±0.4 & 10.5±0.9 & 0.8±0.1 & 69.0±2.2 \\
LwF & 91.5±1.2 & \bbf{85.9±1.8} & 92.6±0.8 & 81.9±1.7 & O.O.T & 55.0±2.1 & 46.4±3.9 & 13.1±0.9 & 10.6±1.0 & 0.8±0.1 & 71.4±2.0 \\
EWC & 91.2±1.3 & 83.7±2.9 & 85.8±2.2 & 86.9±2.2 & 88.7±1.4 & 56.7±5.2 & 45.2±3.7 & 12.3±0.4 & 11.6±2.1 & 2.7±0.4 & 76.1±1.1 \\
MAS & 91.8±1.7 & 84.3±2.3 & 91.8±0.8 & 97.2±0.6 & 94.3±0.9 & \bbf{74.1±1.4} & \bbf{56.0±2.4} & 12.5±1.7 & 9.7±1.8 & 0.8±0.4 & 69.4±1.6 \\
GEM & 88.2±3.1 & 83.4±2.9 & 90.6±0.7 & 83.6±2.0 & O.O.T & 61.8±2.9 & 48.2±3.3 & \bbf{60.7±1.5} & \uls{29.0±4.2} & O.O.T & \uls{81.0±0.3} \\
TWP & 91.0±1.5 & 83.6±2.6 & 84.8±1.5 & 90.0±1.8 & 86.7±2.1 & 56.4±3.1 & 45.0±3.7 & 12.3±0.5 & 10.8±1.4 & 2.7±0.8 & O.O.M \\
ERGNN & 89.0±3.1 & 82.9±2.1 & 87.6±1.5 & 90.0±2.4 & 94.8±0.5 & 60.9±2.6 & 45.7±4.3 & 54.1±1.4 & \bbf{42.5±7.8} & \bbf{17.0±2.1} & N/A \\
CGNN & 91.1±1.5 & 82.9±3.2 & 92.0±0.7 & 91.0±1.6 & \uls{95.3±0.6} & \uls{72.3±1.7} & \uls{53.1±3.5} & 47.7±5.0 & O.O.M & 16.0±1.2 & N/A \\
CaT & 56.9±10.7 & 57.0±6.5 & 86.4±1.7 & 70.4±2.7 & 94.8±0.6 & 22.4±5.0 & 19.1±3.1 & 47.1±3.6 & 9.7±0.8 & \uls{16.9±0.7} & N/A \\
PI-GNN & 90.8±2.8 & \uls{85.3±3.0} & \bbf{93.8±0.6} & \uls{97.5±0.6} & O.O.T & 69.1±5.4 & 50.2±6.8 & \uls{58.7±2.4} & 7.3±1.0 & 1.0±0.5 & \bbf{82.0±0.3} \\
PackNet & \uls{93.3±1.8} & 84.5±1.7 & \uls{93.5±0.7} & 97.3±0.5 & 93.0±1.2 & N/A & N/A & N/A & N/A & N/A & N/A \\
Piggyback & \bbf{93.8±1.6} & 84.6±2.4 & 92.8±0.7 & \bbf{97.7±0.4} & \bbf{95.7±0.8} & N/A & N/A & N/A & N/A & N/A & N/A \\
HAT & 92.0±2.0 & 83.3±4.3 & 90.5±1.3 & 91.5±1.9 & 77.1±2.2 & N/A & N/A & N/A & N/A & N/A & N/A \\
\midrule
Joint & 92.4±1.5 & 85.1±2.7 & 93.8±0.6 & 97.7±0.5 & 96.9±0.3 & 80.0±2.0 & 55.6±4.0 & 65.4±1.6 & 67.4±4.8 & 19.0±0.6 & 73.2±0.2 \\
\bottomrule
\end{tabular}
}
\end{subtable}

\vspace{2mm}

\begin{subtable}[]{\textwidth}
\centering
\scalebox{0.65}{
\begin{tabular}{c|ccc|ccc|cccccc}
\toprule
Problem & \multicolumn{3}{c|}{Node Classification (NC)} & \multicolumn{3}{c|}{Link Classification (LC)} & \multicolumn{6}{c}{Link Prediction (LP)} \\
\midrule
Methods  & \makecell{\twitch \\ (Domain-IL)} & \makecell{\arxiv \\ (Time-IL)} & \makecell{\magdata \\ (Time-IL)} & \makecell{\bitcoin \\ (Task-IL)} & \makecell{\bitcoin \\ (Class-IL)} & \makecell{\bitcoin \\ (Time-IL)} & \makecell{\wikics \\ (Domain-IL)} & \makecell{\facebook \\ (Domain-IL)} & \makecell{\collab \\ (Time-IL)} & \makecell{\askubuntu \\ (Time-IL)} & \makecell{\gowalla \\ (Time-IL)} & \makecell{\movie \\ (Time-IL)} \\
\midrule
Bare & 56.9±3.0 & 62.8±0.4 & 29.6±0.1 & 64.8±7.1 & 24.3±3.0 & \uls{69.6±1.9} & 6.8±3.5 & 4.0±0.9 & \uls{34.7±2.5} & 35.8±0.9 & \bbf{9.4±0.5} & 8.2±0.9 \\
LwF & 58.7±3.5 & 63.9±0.4 & 30.3±0.1 & 70.4±3.2 & 24.2±2.9 & 68.0±2.1 & 6.5±2.8 & \uls{4.5±1.9} & 34.3±2.8 & 35.8±1.9 & \uls{9.3±0.3} & 8.5±1.1 \\
EWC & 59.7±1.6 & 61.3±0.4 & 29.7±0.1 & 68.2±4.6 & 24.2±3.0 & \bbf{70.2±2.1} & 11.3±2.6 & 4.2±1.2 & 23.2±1.4 & 34.6±1.5 & 8.8±0.4 & 8.3±0.6 \\
MAS & 59.5±3.0 & 58.8±0.7 & 28.6±0.6 & \uls{70.6±3.3} & \uls{24.4±2.3} & 68.7±1.7 & 10.0±3.5 & 4.0±1.4 & 22.7±7.5 & \bbf{36.5±0.5} & 8.8±0.5 & \bbf{9.0±0.4} \\
GEM & 60.2±1.8 & 65.4±0.3 & 31.3±0.2 & 70.0±3.5 & \bbf{28.7±4.2} & 68.1±1.9 & \bbf{18.0±2.9} & \bbf{6.2±1.6} & \bbf{38.8±6.3} & \uls{36.0±0.7} & 8.9±0.6 & \uls{8.7±0.5} \\
TWP & 60.0±2.4 & 59.3±1.0 & 29.5±0.1 & 67.3±4.7 & 24.3±2.8 & 66.4±1.4 & \uls{13.3±4.3} & 4.3±1.2 & 30.8±3.0 & 35.7±1.3 & 9.2±0.5 & 8.3±0.6 \\
ERGNN & \uls{61.8±3.0} & \bbf{69.6±0.4} & \bbf{37.9±0.1} & N/A & N/A & N/A & N/A & N/A & N/A & N/A & N/A & N/A \\
CGNN & 60.3±2.7 & \uls{67.6±0.5} & \uls{33.5±0.1} & N/A & N/A & N/A & N/A & N/A & N/A & N/A & N/A & N/A \\
CaT & 52.3±2.5 & 58.8±0.4 & O.O.T & N/A & N/A & N/A & N/A & N/A & N/A & N/A & N/A & N/A \\
PI-GNN & \bbf{62.6±0.9} & 64.6±0.5 & 29.8±0.2 & N/A & N/A & N/A & N/A & N/A & N/A & N/A & N/A & N/A \\
PackNet & N/A & N/A & N/A & \bbf{71.8±3.0} & N/A & N/A & N/A & N/A & N/A & N/A & N/A & N/A \\
Piggyback & N/A & N/A & N/A & 68.1±2.3 & N/A & N/A & N/A & N/A & N/A & N/A & N/A & N/A \\
HAT & N/A & N/A & N/A & 62.7±6.1 & N/A & N/A & N/A & N/A & N/A & N/A & N/A & N/A \\
\midrule
Joint & 62.4±0.8 & 70.4±0.5 & 38.2±0.2 & 73.5±3.5 & 37.7±3.1 & 80.0±1.2 & 22.9±1.8 & 7.7±1.1 & 58.8±2.5 & 36.1±1.3 & 10.4±0.9 & 8.4±0.9 \\
\bottomrule
\end{tabular}
}
\end{subtable}

\vspace{2mm}

\begin{subtable}[]{\textwidth}
\centering
\scalebox{0.65}{
\begin{tabular}{c|cccccccccc|cc}
\toprule
Problem & \multicolumn{10}{c|}{Graph Classification (GC)} & \multicolumn{2}{c}{Graph Regression (GR)} \\
\midrule
Methods
& \makecell{\mnist \\ (Task-IL)} & \makecell{\cifar \\ (Task-IL)} & \makecell{\aroma \\ (Task-IL)} & \makecell{\mnist \\ (Class-IL)} & \makecell{\cifar \\ (Class-IL)} & \makecell{\aroma \\ (Class-IL)} & \makecell{\molhiv \\ (Domain-IL)} & \makecell{\ppa \\ (Domain-IL)} & \makecell{\nyctaxi \\ (Time-IL)} & \makecell{\sentiment \\ (Time-IL)} & \makecell{\zinc \\ (Domain-IL)} & \makecell{\aqsol \\ (Domain-IL)} \\
\midrule
Bare & 69.1±8.1 & 64.6±7.4 & 45.8±4.4 & \uls{19.4±0.5} & \uls{17.5±0.8} & 6.2±1.4 & 68.6±3.9 & \uls{29.9±1.8} & 70.9±0.6 & 70.3±1.0 & \uls{-0.478±0.049} & -1.427±0.155 \\
LwF & \uls{96.5±0.8} & 84.0±3.0 & 54.9±4.4 & \uls{19.4±0.5} & \bbf{17.6±0.9} & \uls{6.4±1.3} & 70.2±4.8 & \bbf{39.6±1.2} & 71.1±0.2 & 70.8±1.3 & \uls{-0.478±0.049} & -1.427±0.155  \\
EWC & 88.8±4.7 & 78.4±4.2 & \uls{58.8±3.0} & 19.3±0.6 & 17.4±0.9 & \uls{6.4±1.2} & \bbf{74.1±3.2} & 18.9±2.6 & \bbf{72.0±1.4} & \uls{70.9±1.0} & -0.508±0.057 & -1.422±0.077 \\
MAS & 84.3±4.4 & 76.2±4.6 & 57.7±4.8 & 19.2±0.6 & \uls{17.5±0.9} & 5.8±1.1 & 68.0±3.0 & 14.5±2.1 & \uls{71.4±0.5} & \bbf{71.2±0.8} & \bbf{-0.472±0.051} & -1.469±0.172 \\
GEM & 89.4±2.8 & 76.9±3.0 & 50.0±4.5 & \bbf{19.9±2.0} & 17.4±1.0 & \bbf{6.5±1.3} & 70.7±3.3 & 29.4±1.6 & 70.6±6.6 & 67.4±1.9 & -0.492±0.029 & \bbf{-1.340±0.088} \\
TWP & 88.3±6.0 & 78.8±4.4 & \bbf{59.3±4.3} & 19.3±0.5 & 17.4±0.9 & 6.2±0.9 & \uls{73.9±3.4} & 20.2±2.1 & 71.2±0.3 & 69.9±0.8 & -0.499±0.057 & \uls{-1.412±0.143} \\
PackNet & \bbf{97.2±0.7} & \bbf{85.5±2.3} & 49.1±8.9 & N/A & N/A & N/A & N/A & N/A & N/A & N/A & N/A & N/A \\
Piggyback & 96.2±1.1 & \uls{85.2±2.1} & 57.9±4.2 & N/A & N/A & N/A & N/A & N/A & N/A & N/A & N/A & N/A \\
HAT & 69.8±8.1 & 64.3±7.8 & 45.5±3.6 & N/A & N/A & N/A & N/A & N/A & N/A & N/A & N/A & N/A \\
\midrule
Joint & 97.7±0.7 & 86.8±2.1 & 76.5±2.9 & 90.0±0.4 & 52.1±0.3 & 28.6±1.8 & 77.9±1.2 & 80.1±1.3 & 86.3±0.4 & 74.7±1.0 & -0.320±0.007 & -0.858±0.012 \\
\bottomrule
\end{tabular}
}
\end{subtable}
\end{table*}

\begin{table*}[t!]
\caption{\textbf{Performance in terms of Average Forgetting (AF, the lower, the better)}. In each setting, the best score is in bold, and the second best score is underlined. Each number is rescaled to a range of 100. O.O.M: out of memory. O.O.T: out of time ($>$ 24 hours). N/A: not applicable to the problems or scenarios.}
\label{tab:results_af_full}
\setlength{\tabcolsep}{1pt}
\renewcommand{\arraystretch}{1.2}
\centering
\begin{subtable}[]{\textwidth}
\centering
\scalebox{0.65}{
\begin{tabular}{c|ccccccccccc}
\toprule
Problem & \multicolumn{11}{c}{Node Classification (NC)} \\
\midrule
Methods
& \makecell{\cora \\ (Task-IL)} & \makecell{\citeseer \\ (Task-IL)} & \makecell{\arxiv \\ (Task-IL)} & \makecell{\corafull \\ (Task-IL)} & \makecell{\magdata \\ (Task-IL)} & \makecell{\cora \\ (Class-IL)} & \makecell{\citeseer \\ (Class-IL)} & \makecell{\arxiv \\ (Class-IL)} & \makecell{\products \\ (Class-IL)} & \makecell{\magdata \\ (Class-IL)} & \makecell{\proteins \\ (Domain-IL)} \\
\midrule
Bare & 2.6±2.3 & 4.2±3.0 & 32.5±8.5 & 22.6±2.0 & 32.3±1.7 & 56.5±3.9 & 55.0±6.6 & 93.4±0.8 & 85.0±4.9 & 96.9±0.4 & 13.1±3.4 \\
LwF & 1.2±1.9 & 1.1±1.7 & 1.0±0.4 & 16.5±1.5 & O.O.T & 53.4±2.5 & 53.9±6.0 & 91.7±1.0 & 84.5±5.3 & 96.8±0.4 & 5.4±2.6 \\
EWC & 2.0±1.4 & 4.1±3.3 & 8.3±2.4 & 11.4±2.4 & 7.8±1.8 & 50.8±8.7 & 54.2±6.5 & 92.4±0.6 & 82.9±7.4 & 50.3±6.8 & 7.4±2.5 \\
MAS & 0.5±0.7 & 2.1±1.3 & 0.6±0.4 & \uls{0.1±0.3} & 0.6±0.5 & \bbf{22.7±3.9} & \bbf{28.3±5.0} & 89.4±4.4 & 73.6±7.1 & 57.4±4.4 & 1.2±2.7 \\
GEM & 6.0±5.7 & 4.1±2.6 & 3.6±0.5 & 14.7±2.1 & O.O.T & 44.4±4.6 & 50.7±5.3 & 22.8±3.0 & 50.3±8.4 & O.O.T & \uls{0.3±2.8} \\
TWP & 2.5±1.8 & 3.8±2.4 & 9.4±1.3 & 8.2±1.9 & 10.0±2.3 & 52.4±4.1 & 54.5±6.3 & 92.4±0.9 & 82.9±6.1 & 65.5±8.2 & O.O.M \\
ERGNN & 5.2±5.9 & 5.0±2.4 & 6.3±1.2 & 7.8±2.2 & 2.4±0.4 & 44.7±4.5 & 51.8±5.8 & \bbf{-2.3±7.1} & 37.5±12.1 & 71.7±2.4 & N/A \\
CGNN & 2.3±1.6 & 5.0±3.4 & 1.6±0.3 & 6.5±1.4 & 1.9±0.5 & 25.1±3.4 & \uls{38.4±5.5} & 43.7±6.4 & O.O.M & 59.4±1.1 & N/A \\
CaT & 6.3±15.3 & 3.3±6.1 & \uls{0.2±1.7} & 3.1±2.7 & \bbf{-0.8±0.4} & 52.4±13.2 & 53.4±8.1 & \uls{22.5±6.4} & \bbf{5.6±1.0} & \uls{11.5±5.1} & N/A \\
PI-GNN & \uls{0.0±0.0} & \uls{0.0±0.0} & \bbf{0.0±0.0} & \bbf{0.0±0.0} & O.O.T & \uls{24.2±3.9} & 39.7±13.5 & 29.8±3.4 & \uls{24.0±3.5} & \bbf{10.2±1.1} & \bbf{-0.8±0.4} \\
PackNet & \uls{0.0±0.0} & \uls{0.0±0.0} & \bbf{0.0±0.0} & \bbf{0.0±0.0} & \uls{0.0±0.0} & N/A & N/A & N/A & N/A & N/A & N/A \\
Piggyback & \uls{0.0±0.0} & \uls{0.0±0.0} & \bbf{0.0±0.0} & \bbf{0.0±0.0} & \uls{0.0±0.0} & N/A & N/A & N/A & N/A & N/A & N/A \\
HAT & \bbf{-2.1±1.9} & \bbf{-1.3±1.8} & 3.2±1.2 & 6.6±2.0 & 20.2±2.2 & N/A & N/A & N/A & N/A & N/A & N/A \\
\bottomrule
\end{tabular}
}
\end{subtable}

\vspace{2mm}

\begin{subtable}[]{\textwidth}
\centering
\scalebox{0.65}{
\begin{tabular}{c|ccc|ccc|cccccc}
\toprule
Problem & \multicolumn{3}{c|}{Node Classification (NC)} & \multicolumn{3}{c|}{Link Classification (LC)} & \multicolumn{6}{c}{Link Prediction (LP)} \\
\midrule
Methods & \makecell{\twitch \\ (Domain-IL)} & \makecell{\arxiv \\ (Time-IL)} & \makecell{\magdata \\ (Time-IL)} & \makecell{\bitcoin \\ (Task-IL)} & \makecell{\bitcoin \\ (Class-IL)} & \makecell{\bitcoin \\ (Time-IL)} & \makecell{\wikics \\ (Domain-IL)} & \makecell{\facebook \\ (Domain-IL)} & \makecell{\collab \\ (Time-IL)} & \makecell{\askubuntu \\ (Time-IL)} & \makecell{\gowalla \\ (Time-IL)} & \makecell{\movie \\ (Time-IL)} \\
\midrule
Bare & 5.5±3.4 & -4.9±0.6 & 5.1±0.1 & 11.4±8.7 & 72.2±4.7 & 11.8±2.7 & 29.9±4.3 & 5.9±1.6 & 24.5±2.2 & \bbf{4.5±0.8} & 0.5±0.8 & \bbf{0.8±1.3}\\
LwF & 4.3±3.6 & -4.4±0.6 & 4.8±0.2 & 3.5±2.1 & 72.6±4.7 & \uls{9.2±4.2} & 31.1±3.1 & 5.9±2.2 & 35.1±3.3 & \uls{5.3±1.5} & 1.1±0.6 & \uls{1.0±1.6}\\
EWC & 3.6±1.7 & -8.4±1.2 & 4.4±0.1 & 6.3±4.6 & 72.4±6.0 & 9.9±2.1 & 14.6±4.6 & 6.4±1.4 & \bbf{5.5±3.7} & 7.5±1.1 & \uls{0.4±0.6} & 2.1±1.5\\
MAS & 3.9±3.0 & -11.6±1.6 & 1.8±0.8 & 2.9±2.5 & 72.6±4.9 & 11.5±2.5 & \uls{13.4±4.7} & 5.9±1.6 & \uls{7.3±6.6} & 6.3±0.5 & \uls{0.4±0.6} & 2.8±0.5\\
GEM & 1.9±1.8 & \uls{-13.6±0.6} & -1.1±0.3 & \uls{2.6±2.1} & \bbf{57.9±10.0} & \bbf{4.2±3.3} & \textbf{4.4±5.7} & \bbf{3.2±2.3} & 29.9±6.5 & 6.3±1.0 & 1.0±0.9 & 3.4±0.7\\
TWP & 3.4±2.3 & -3.9±1.1 & 4.4±0.1 & 8.2±6.0 & \uls{72.1±5.5} & 19.3±1.8 & 18.3±2.8 & \uls{5.7±1.2} & 17.6±1.7 & 6.7±1.1 & \bbf{0.0±0.4} & 2.4±1.0\\
ERGNN & \bbf{0.2±3.1} & -9.5±0.5 & \bbf{-5.7±0.1} & N/A & N/A & N/A & N/A & N/A & N/A & N/A & N/A & N/A\\
CGNN & 2.1±2.9 & -7.6±0.7 & 1.3±0.1 & N/A & N/A & N/A & N/A & N/A & N/A & N/A & N/A & N/A \\
CaT & 2.8±1.9 & -4.0±0.6 & O.O.T & N/A & N/A & N/A & N/A & N/A & N/A & N/A & N/A & N/A \\
PI-GNN & \uls{0.5±0.9} & \bbf{-15.7±0.6} & \uls{-1.3±0.2} & N/A & N/A & N/A & N/A & N/A & N/A & N/A & N/A & N/A \\
PackNet & N/A & N/A & N/A & \bbf{0.0±0.0} & N/A & N/A & N/A & N/A & N/A & N/A & N/A & N/A \\
Piggyback & N/A & N/A & N/A & \bbf{0.0±0.0} & N/A & N/A & N/A & N/A & N/A & N/A & N/A & N/A \\
HAT & N/A & N/A & N/A & 13.4±8.2 & N/A & N/A & N/A & N/A & N/A & N/A & N/A & N/A \\
\bottomrule
\end{tabular}
}
\end{subtable}

\vspace{2mm}

\begin{subtable}[]{\textwidth}
\centering
\scalebox{0.65}{
\begin{tabular}{c|cccccccccc|cc}
\toprule
Problem & \multicolumn{10}{c|}{Graph Classification (GC)} & \multicolumn{2}{c}{Graph Regression (GR)} \\
\midrule
Methods
& \makecell{\mnist \\ (Task-IL)} & \makecell{\cifar \\ (Task-IL)} & \makecell{\aroma \\ (Task-IL)} & \makecell{\mnist \\ (Class-IL)} & \makecell{\cifar \\ (Class-IL)} & \makecell{\aroma \\ (Class-IL)} & \makecell{\molhiv \\ (Domain-IL)} & \makecell{\ppa \\ (Domain-IL)} & \makecell{\nyctaxi \\ (Time-IL)} & \makecell{\sentiment \\ (Time-IL)} & \makecell{\zinc \\ (Domain-IL)} & \makecell{\aqsol \\ (Domain-IL)} \\
\midrule
Bare & 35.6±9.9 & 27.0±8.2 & 18.2±5.1 & 97.8±0.8 & 85.6±2.1 & 69.4±2.6 & 10.7±5.3 & 60.1±1.9 & \uls{2.8±1.3} & \uls{-2.6±1.6} & 10.6±5.3 & 70.7±19.4 \\
LwF & \uls{1.2±0.4} & \uls{3.0±1.5} & 16.4±4.6 & 97.6±0.9 & 85.7±2.3 & 69.5±2.6 & 9.1±5.7 & 49.3±1.5 & \bbf{1.4±0.7} & -1.5±1.1 & 10.6±5.3 & 70.7±19.4 \\
EWC & 8.4±4.7 & 5.5±2.7 & 8.7±3.0 & 97.7±0.8 & 85.5±2.2 & 67.4±3.4 & \uls{3.8±2.8} & \bbf{6.2±1.0} & 7.1±2.4 & -1.1±1.1 & 10.3±6.0 & \uls{49.5±8.6} \\
MAS & 10.3±4.8 & 7.7±3.9 & 8.2±3.1 & \uls{97.3±0.8} & 84.9±2.6 & \uls{64.7±4.0} & \bbf{0.7±3.1} & \uls{7.4±1.1} & 5.6±0.6 & \bbf{-3.0±0.7}  & \uls{9.8±5.1} & 62.5±16.3\\
GEM & 9.0±2.9 & 10.6±2.3 & 16.9±3.2 & \bbf{86.6±8.0} & \bbf{84.3±2.2} & \bbf{33.7±9.6} & 6.5±4.0 & 51.5±3.1 & 6.5±6.5 & -0.0±2.8 & \bbf{7.2±3.0} & \bbf{48.0±10.2}\\
TWP & 9.6±6.2 & 5.6±3.0 & \uls{8.1±2.3} & 97.4±0.8 & \uls{84.4±2.6} & 67.8±3.4 & 5.0±3.4 & \uls{7.4±1.2} & 5.5±0.9 & -1.6±1.0 & 10.9±6.0 & 51.1±16.8\\
PackNet & \bbf{0.0±0.0} & \bbf{0.0±0.0} & \bbf{0.0±0.0} & N/A & N/A & N/A & N/A & N/A & N/A & N/A & N/A & N/A \\
Piggyback & \bbf{0.0±0.0} & \bbf{0.0±0.0} & \bbf{0.0±0.0} & N/A & N/A & N/A & N/A & N/A & N/A & N/A & N/A & N/A \\
HAT & 34.7±9.9 & 27.2±8.9 & 19.0±5.5 & N/A & N/A & N/A & N/A & N/A & N/A & N/A & N/A & N/A \\
\bottomrule
\end{tabular}
}
\end{subtable}
\end{table*}

\begin{table*}[t!]
\caption{\textbf{Performance in terms of Intransigence (INT, the lower, the better)}. In each setting, the best score is in bold, and the second best score is underlined. Each number is rescaled to a range of 100. O.O.M: out of memory. O.O.T: out of time ($>$ 24 hours). N/A: not applicable to the problems or scenarios.}
\label{tab:results_int_full}
\setlength{\tabcolsep}{1pt}
\renewcommand{\arraystretch}{1.2}
\centering
\begin{subtable}[]{\textwidth}
\centering
\scalebox{0.65}{
\begin{tabular}{c|ccccccccccc}
\toprule
Problem & \multicolumn{11}{c}{Node Classification (NC)} \\
\midrule
Methods
& \makecell{\cora \\ (Task-IL)} & \makecell{\citeseer \\ (Task-IL)} & \makecell{\arxiv \\ (Task-IL)} & \makecell{\corafull \\ (Task-IL)} & \makecell{\magdata \\ (Task-IL)} & \makecell{\cora \\ (Class-IL)} & \makecell{\citeseer \\ (Class-IL)} & \makecell{\arxiv \\ (Class-IL)} & \makecell{\products \\ (Class-IL)} & \makecell{\magdata \\ (Class-IL)} & \makecell{\proteins \\ (Domain-IL)} \\
\midrule
Bare & 0.1±1.3 & -2.3±5.2 & 0.4±0.1 & \uls{-0.2±0.3} & -0.6±0.1 & \bbf{-6.4±1.8} & -12.9±5.3 & \bbf{-16.8±3.1} & \bbf{-13.2±3.7} & \bbf{-62.5±1.6} & -6.7±0.6 \\
LwF & -0.1±0.5 & \bbf{-2.6±5.2} & 0.4±0.2 & \uls{-0.2±0.3} & O.O.T & -5.2±2.0 & \bbf{-13.9±5.2} & \uls{-16.4±3.0} & \uls{-12.9±4.0} & \uls{-62.4±1.6} & -2.3±2.1 \\
EWC & -0.5±1.1 & \uls{-2.4±5.4} & 0.8±0.2 & \uls{-0.2±0.2} & 0.0±0.3 & -5.2±2.3 & -12.9±5.2 & -16.3±3.1 & -12.4±4.0 & -18.2±6.6 & \bbf{-8.9±0.9} \\
MAS & -0.1±1.0 & -1.6±5.0 & 1.4±0.2 & 0.4±0.4 & 1.6±0.5 & -3.8±2.8 & -6.4±5.0 & -14.2±3.1 & -2.3±5.4 & -23.4±4.9 & 3.4±2.6 \\
GEM & -0.1±0.7 & -2.1±5.1 & \uls{0.1±0.1} & \uls{-0.2±0.3} & O.O.T & \uls{-6.0±1.8} & \uls{-13.5±5.3} & -3.7±2.0 & -0.9±4.4 & O.O.T & \uls{-8.3±2.5} \\
TWP & -0.6±0.9 & -2.2±5.3 & 0.8±0.1 & 0.1±0.2 & 0.0±0.1 & -5.9±2.1 & -12.8±5.3 & -16.2±3.0 & -11.6±4.1 & -21.0±5.8 & O.O.M \\
ERGNN & -0.3±1.1 & -2.2±5.3 & 0.7±0.2 & 0.2±0.3 & \bbf{-0.7±0.2} & -5.3±1.8 & -11.8±5.6 & 24.9±5.9 & -3.9±4.5 & -53.8±1.8 & N/A \\
CGNN & -0.5±0.9 & -2.2±5.2 & 0.4±0.1 & 0.3±0.3 & \bbf{-0.7±0.3} & -3.7±1.1 & -10.2±5.1 & -9.0±2.4 & O.O.M & -40.5±1.1 & N/A \\
CaT & 31.0±10.0 & 24.8±5.9 & 7.2±1.9 & 24.3±2.6 & 3.0±0.8 & 28.1±8.5 & 13.8±10.2 & 10.2±4.7 & 52.9±5.9 & 12.0±2.2 & N/A \\
PI-GNN & 1.3±1.5 & -1.2±4.0 & \bbf{0.0±0.1} & 0.1±0.8 & O.O.T & 0.1±6.3 & -8.2±6.2 & -7.8±3.4 & 44.2±4.6 & 23.3±1.9 & -7.6±3.2 \\
PackNet & \uls{-1.2±0.9} & 2.2±2.3 & 0.3±0.1 & 0.3±0.4 & 3.5±1.0 & N/A & N/A & N/A & N/A & N/A & N/A \\
Piggyback & \bbf{-1.7±1.2} & 2.0±2.0 & 1.0±0.1 & 0.0±0.3 & 1.8±0.8 & N/A & N/A & N/A & N/A & N/A & N/A \\
HAT & 1.5±1.6 & 4.2±3.9 & 3.2±1.2 & \bbf{-0.3±0.3} & \bbf{-0.7±0.2} & N/A & N/A & N/A & N/A & N/A & N/A \\
\bottomrule
\end{tabular}
}
\end{subtable}

\vspace{2mm}

\begin{subtable}[]{\textwidth}
\centering
\scalebox{0.65}{
\begin{tabular}{c|ccc|ccc|cccccc}
\toprule
Problem & \multicolumn{3}{c|}{Node Classification (NC)} & \multicolumn{3}{c|}{Link Classification (LC)} & \multicolumn{6}{c}{Link Prediction (LP)} \\
\midrule
Methods & \makecell{\twitch \\ (Domain-IL)} & \makecell{\arxiv \\ (Time-IL)} & \makecell{\magdata \\ (Time-IL)} & \makecell{\bitcoin \\ (Task-IL)} & \makecell{\bitcoin \\ (Class-IL)} & \makecell{\bitcoin \\ (Time-IL)} & \makecell{\wikics \\ (Domain-IL)} & \makecell{\facebook \\ (Domain-IL)} & \makecell{\collab \\ (Time-IL)} & \makecell{\askubuntu \\ (Time-IL)} & \makecell{\gowalla \\ (Time-IL)} & \makecell{\movie \\ (Time-IL)} \\
\midrule
Bare & -0.3±1.0 & -5.3±1.3 & -0.3±0.1 & 0.5±1.4 & -21.2±6.7 & \uls{0.3±1.7} & \uls{-14.7±2.2} & -0.9±1.5 & 3.1±3.7 & 1.7±0.8 & \uls{-9.9±0.5} & 0.8±0.7 \\
LwF & -0.9±0.8 & -6.9±1.7 & \bbf{-0.8±0.1} & \uls{0.1±1.6} & \uls{-21.3±6.4} & 4.1±2.9 & \bbf{-15.5±3.4} & \bbf{-1.5±1.8} & \bbf{-6.8±4.6} & 0.8±0.5 & \bbf{-10.3±0.5} & 0.3±1.4 \\
EWC & \uls{-1.3±0.7} & -0.5±2.1 & 0.2±0.1 & 0.4±1.0 & -21.2±6.6 & 1.4±2.1 & -4.2±6.1 & \bbf{-1.5±2.7} & 33.3±5.1 & -0.1±0.8 & -9.2±0.2 & -0.5±1.3 \\
MAS & \bbf{-1.4±1.0} & 5.1±2.0 & 3.6±0.2 & 0.3±1.8 & \bbf{-21.5±7.0} & 1.4±1.6 & -1.7±3.1 & -1.0±2.1 & 32.0±5.1 & \bbf{-0.8±0.5} & -9.1±0.2 & \uls{-1.8±0.6} \\
GEM & -0.1±0.9 & 0.4±1.8 & 3.5±0.4 & 1.1±1.3 & -16.0±8.0 & 8.3±2.4 & -0.9±4.1 & -0.8±2.3 & \uls{-6.2±3.8} & \uls{-0.3±0.4} & -9.8±0.3 & \bbf{-2.2±0.5}\\
TWP & \uls{-1.3±0.9} & -2.8±1.5 & 0.4±0.1 & \bbf{0.0±1.2} & -21.0±6.0 & \bbf{-3.0±1.1} & -9.8±6.0 & \uls{-1.1±2.3} & 13.9±5.0 & \uls{-0.3±0.4} & -9.2±0.3 & -0.7±0.5 \\
ERGNN & -0.1±1.1 & \bbf{-7.7±1.6} & 1.1±0.1 & N/A & N/A & N/A & N/A & N/A & N/A & N/A& N/A & N/A \\
CGNN & -0.4±1.3 & \uls{-7.6±1.4} & \bbf{-0.8±0.1} & N/A & N/A & N/A & N/A & N/A & N/A & N/A& N/A & N/A  \\
CaT & 6.9±1.1 & -2.0±1.6 & O.O.T & N/A & N/A & N/A & N/A & N/A & N/A & N/A& N/A & N/A  \\
PI-GNN & -1.2±0.7 & 3.5±1.5 & 4.5±0.1 & N/A & N/A & N/A & N/A & N/A & N/A & N/A& N/A & N/A \\
PackNet & N/A & N/A & N/A & \uls{0.1±1.9} & N/A & N/A & N/A & N/A & N/A & N/A& N/A & N/A \\
Piggyback & N/A & N/A & N/A & 3.1±3.3 & N/A & N/A & N/A & N/A & N/A & N/A& N/A & N/A \\
HAT & N/A & N/A & N/A & 0.3±1.3 & N/A & N/A & N/A & N/A & N/A & N/A& N/A & N/A \\
\bottomrule
\end{tabular}
}
\end{subtable}

\vspace{2mm}

\begin{subtable}[]{\textwidth}
\centering
\scalebox{0.65}{
\begin{tabular}{c|cccccccccc|cc}
\toprule
Problem & \multicolumn{10}{c|}{Graph Classification (GC)} & \multicolumn{2}{c}{Graph Regression (GR)} \\
\midrule
Methods
& \makecell{\mnist \\ (Task-IL)} & \makecell{\cifar \\ (Task-IL)} & \makecell{\aroma \\ (Task-IL)} & \makecell{\mnist \\ (Class-IL)} & \makecell{\cifar \\ (Class-IL)} & \makecell{\aroma \\ (Class-IL)} & \makecell{\molhiv \\ (Domain-IL)} & \makecell{\ppa \\ (Domain-IL)} & \makecell{\nyctaxi \\ (Time-IL)} & \makecell{\sentiment \\ (Time-IL)} & \makecell{\zinc \\ (Domain-IL)} & \makecell{\aqsol \\ (Domain-IL)} \\
\midrule
Bare & \bbf{0.0±0.3} & \uls{0.4±0.3} & 10.6±3.4 & \bbf{-4.5±0.5} & \uls{-20.3±1.6} & -29.0±4.0 & \uls{-0.4±2.5} & \bbf{-5.4±1.3} & 9.8±1.0 & 2.7±1.2 & \bbf{3.7±0.7} & \bbf{-0.6±2.4} \\
LwF & \uls{0.1±0.1} & \bbf{0.2±0.4} & \bbf{3.1±2.6} & \uls{-4.3±0.6} & \bbf{-20.4±1.7} & \bbf{-29.3±4.9} & \bbf{-0.5±1.9} & \uls{-5.3±1.2} & 10.9±0.8 & \uls{1.8±1.2} & \bbf{3.7±0.7} & \bbf{-0.6±2.4} \\
EWC & 2.1±0.6 & 3.8±1.2 & \uls{6.1±2.1} & -4.2±0.6 & -20.1±1.7 & -27.4±5.0 & 0.7±1.4 & 54.6±2.4 & \bbf{4.7±1.8} & \bbf{0.8±0.9} & \uls{3.8±1.3} & 15.8±4.7 \\
MAS & 5.1±1.1 & 4.3±1.3 & 7.6±2.9 & -3.9±0.6 & -19.7±1.8 & -24.3±5.2 & 9.8±2.7 & 57.9±1.6 & 6.7±0.8 & 2.2±0.8 & 7.0±1.3 & 10.0±8.8 \\
GEM & 1.1±0.5 & 1.3±0.5 & 7.5±2.8 & 4.0±5.2 & -19.1±2.0 & 2.8±9.0 & 1.5±1.9 & 2.9±2.0 & \uls{6.6±2.3} & 3.2±1.6 & 8.2±2.0 & \uls{8.8±4.4} \\
TWP & 1.6±0.4 & 3.3±0.9 & 6.2±1.9 & -4.0±0.6 & -19.3±1.8 & -27.5±5.2 & -0.3±0.9 & 52.1±2.1 & 7.0±1.0 & 2.2±1.0 & 5.5±1.3 & 13.5±4.6 \\
PackNet & 0.4±0.3 & 1.2±0.5 & 23.7±8.9 & N/A & N/A & N/A & N/A & N/A & N/A & N/A & N/A & N/A\\
Piggyback & 1.4±0.7 & 1.9±0.7 & 14.9±3.7 & N/A & N/A & N/A & N/A & N/A & N/A & N/A & N/A & N/A\\
HAT & 0.0±0.3 & 0.6±0.4 & 9.4±2.8 & N/A & N/A & N/A & N/A & N/A & N/A & N/A & N/A & N/A\\
\bottomrule
\end{tabular}
}
\end{subtable}
\end{table*}

\begin{table*}[t]
\caption{\textbf{Performance in terms of Forward Transfer (FWT, the higher, the better)}. In each setting, the best score is in bold, and the second best score is underlined. Each number is rescaled to a range of 100. O.O.M: out of memory.}
\label{tab:results_fwt}
\centering
\scalebox{0.67}{
\begin{tabular}{c|cccccccc}
\toprule
\multirow{2}{*}{Methods} & \multicolumn{8}{c}{Domain-IL} \\
& \proteins (NC) & \twitch (NC) & \wikics (LP) & \facebook (LP) & \molhiv (GC) & \ppa (GC) & \zinc (GR) & \aqsol (GR) \\
\midrule
Bare & 15.9±2.4 & 6.1±4.2 & 3.3±1.8 & 1.2±2.9 & 17.9±5.5 & 15.7±1.3 & 1.017±0.037 & \uls{1.605±0.244} \\
LwF & 16.4±2.6 & 6.6±4.2 & 3.2±1.6 & \bbf{1.9±2.0} & 19.5±5.3 & \bbf{19.7±1.5} & 1.017±0.037 & \uls{1.605±0.244} \\
EWC & \uls{17.1±2.9} & \bbf{8.3±4.4} & 6.8±2.2 & 0.9±2.4 & \uls{20.4±4.3} & 11.5±1.7 & 1.020±0.026 & 1.563±0.292\\
MAS & 16.5±3.3 & 7.2±4.3 & 5.6±2.7 & 0.6±2.1 & 15.5±5.7 & 8.5±1.4 & \bbf{1.040±0.029} & 1.562±0.305 \\
GEM & 16.6±3.3 & 7.4±4.3 & \uls{6.9±2.8} & 0.8±2.0 & 17.1±4.5 & \uls{18.1±2.0} & 1.025±0.026 & \bbf{1.630±0.200}\\
TWP & O.O.M & 7.5±4.1 & \bbf{7.5±2.1} & \uls{1.3±2.3} & \bbf{21.4±4.7} & 12.0±1.3 & \uls{1.026±0.030} & 1.589±0.276 \\
ERGNN & N/A & 7.9±4.3 & N/A & N/A & N/A & N/A & N/A & N/A \\
CGNN & N/A & \uls{8.1±4.4} & N/A & N/A & N/A & N/A & N/A & N/A \\
CaT & N/A & 0.5±4.2 & N/A & N/A & N/A & N/A & N/A & N/A \\
PI-GNN & \bbf{18.7±1.6} & \bbf{11.5±3.6} & N/A & N/A & N/A & N/A & N/A & N/A \\
\bottomrule
\end{tabular}
}
\end{table*}

\subsection{Forward Transfer}
\label{sec:app:fwt}

We report the forward transfer (FWT) of the considered methods in Table~\ref{tab:results_fwt}.
Note that we are able to compute FWT only in Domain-IL. Specifically, in the other settings, the set of classes may vary with tasks (i.e., $\SCi\neq \SC_{i-1}$), and thus it may not be possible to infer classes on the next task (i.e., $\SCi$) right after being trained for the current task (i.e., $\mathcal{T}_{i-1}$) without being trained for the next task (i.e., $\mathcal{T}_{i}$). 
For the NC problem, PI-GNN performed best.
For the LP, GC, and GR problems, except for the the \ppa and \aqsol datasets, TWP performed best or second best.

\section{Effects of the Number of Total Tasks}
\label{sec:app:numberoftotaltask}
We conducted additional experiments to investigate the effect of the number of total tasks $N$ on the performance of graph CL methods.
Accordingly, we changed the number of classes considered (additionally) in each task proportionally to the reciprocal of the number of tasks.
Under Task- and Class-IL settings on \arxiv, we measured how the performance changes depending on the number of total tasks from $5$ to $20$.
In Table~\ref{tab:results_numtask_AP}, we report the benchmark final (i.e., when $k$ is equal to $N$) performance in terms of AP, AF, and INT, respectively.

\begin{table*}[t]
\caption{\textbf{Effects of the number of total tasks
on Average Performance, Average Forgetting, and Intransigence.} In each setting, the best score is in bold, and the second best score is underlined. Each number is rescaled to a range of 100. O.O.M: out of memory. N/A: methods are not applicable to the problems or scenarios.}
\label{tab:results_numtask_AP}
\begin{subtable}[]{\textwidth}
\centering
\scalebox{0.75}{
\begin{tabular}{c|ccc|ccc}
\toprule
\multirow{2}{*}{Methods} & \multicolumn{3}{c|}{\arxiv (Task-IL)} & \multicolumn{3}{c}{\arxiv (Class-IL)} \\
& $N = 5$ & $N = 8$ & $N = 20$ & $N = 5$ & $N = 8$ & $N = 20$ \\
\midrule
Bare & 64.9±6.5 & 65.0±7.6 & 81.3±3.5 & 18.5±1.0 & 12.0±0.4 & 5.3±0.7 \\
LwF & \uls{89.5±1.6} & 92.6±0.8 & 96.3±1.0 & 19.9±1.3 & 13.1±0.9 & 5.7±0.9 \\
EWC & 79.4±3.6 & 85.8±2.2 & 93.4±0.9 & 18.8±0.8 & 12.3±0.4 & 5.9±1.2 \\
MAS & 88.2±1.7 & 91.8±0.8 & 96.4±0.9 & 18.7±1.4 & 12.5±1.7 & 4.8±0.7 \\
GEM & 87.3±1.8 & 90.6±0.7 & 95.9±0.9 & \bbf{62.6±2.1} & \bbf{60.7±1.5} & \bbf{55.1±1.1} \\
TWP & 79.2±4.0 & 84.8±1.5 & 93.2±1.7 & 18.9±1.0 & 12.3±0.5 & 5.9±1.0 \\
ERGNN & 81.3±2.2 & 87.6±1.5 & 94.9±1.0 & \uls{53.8±2.3} & \uls{54.1±1.4} & \uls{44.1±8.8} \\
CGNN & 89.4±1.5 & 92.0±0.7 & 96.4±0.9 & 53.3±4.0 & 47.7±5.0 & 39.5±4.6 \\
PackNet & \bbf{90.4±1.6} & \bbf{93.5±0.7} & \bbf{97.1±0.7} & N/A & N/A & N/A \\
Piggyback & 89.3±1.7 & \uls{92.8±0.7} & \bbf{97.1±0.7} & N/A & N/A & N/A \\
HAT & 87.4±1.8 & 90.5±1.3 & 95.1±1.4 & N/A & N/A & N/A \\
\midrule
Joint & 90.6±1.5 & 93.8±0.6 & 97.4±0.7 & 68.1±2.3 & 65.4±1.6 & 57.8±1.6 \\
\bottomrule
\end{tabular}
}
\caption{Average Performance (AP, the higher, the better)}
\end{subtable}

\begin{subtable}[]{\textwidth}
\centering
\scalebox{0.75}{
\begin{tabular}{c|ccc|ccc}
\toprule
\multirow{2}{*}{Methods} & \multicolumn{3}{c|}{\arxiv (Task-IL)} & \multicolumn{3}{c}{\arxiv (Class-IL)} \\
& $N = 5$ & $N = 8$ & $N = 20$ & $N = 5$ & $N = 8$ & $N = 20$ \\
\midrule
Bare & 31.7±7.4 & 32.5±8.5 & 16.9±3.4 & 89.8±1.7 & 93.4±0.8 & 96.9±1.3 \\
LwF & 1.0±0.4 & 1.0±0.4 & 1.0±0.4 & 87.9±1.6 & 91.7±1.0 & 96.3±1.2 \\
EWC & 13.3±3.7 & 8.3±2.4 & 4.1±1.1 & 89.2±1.8 & 92.4±0.6 & 96.1±1.4 \\
MAS & \bbf{0.0±0.1} & \uls{0.6±0.4} & \bbf{0.0±0.1} & 86.6±1.6 & 89.4±4.4 & 58.4±9.1 \\
GEM & 4.1±0.8 & 3.6±0.5 & 1.6±0.5 & \uls{22.0±3.3} & \uls{22.8±3.0} & \bbf{25.8±1.8} \\
TWP & 13.1±4.0 & 9.4±1.3 & 4.0±1.6 & 89.0±2.1 & 92.4±0.9 & 95.9±1.3 \\
ERGNN & 8.8±2.4 & 6.3±1.2 & 2.1±0.5 & \bbf{-13.8±8.6} & \bbf{-2.3±7.1} & \uls{34.9±9.5} \\
CGNN & 1.6±0.3 & 1.6±0.3 & 1.1±0.4 & 36.7±6.1 & 43.7±6.4 & 51.1±5.2 \\
PackNet & \bbf{0.0±0.0} & \bbf{0.0±0.0} & \bbf{0.0±0.0} & N/A & N/A & N/A \\
Piggyback & \bbf{0.0±0.0} & \bbf{0.0±0.0} & \bbf{0.0±0.0} & N/A & N/A & N/A \\
HAT & 3.2±1.3 & 3.2±1.2 & 2.3±1.1 & N/A & N/A & N/A \\
\bottomrule
\end{tabular}
}
\caption{Average Forgetting (AF, the lower, the better)}
\end{subtable}

\begin{subtable}[]{\textwidth}
\label{tab:results_numtask_INT}
\centering
\scalebox{0.75}{
\begin{tabular}{c|ccc|ccc}
\toprule
\multirow{2}{*}{Methods} & \multicolumn{3}{c|}{\arxiv (Task-IL)} & \multicolumn{3}{c}{\arxiv (Class-IL)} \\
& $N = 5$ & $N = 8$ & $N = 20$ & $N = 5$ & $N = 8$ & $N = 20$ \\
\midrule
Bare & 0.5±0.1 & 0.4±0.1 & \uls{0.0±0.1} & \bbf{-12.7±3.4} & \bbf{-16.8±3.1} & \bbf{-25.1±3.4} \\
LwF & 0.5±0.1 & 0.4±0.2 & 0.1±0.1 & \bbf{-12.7±3.4} & \uls{-16.4±3.0} & -24.9±3.4 \\
EWC & 0.5±0.1 & 0.8±0.2 & \uls{0.0±0.2} & -12.6±3.4 & -16.3±3.1 & \uls{-24.9±3.4} \\
MAS & 2.5±0.3 & 1.4±0.2 & 0.9±0.2 & -10.5±3.3 & -14.2±3.1 & 12.0±7.2 \\
GEM & \uls{0.1±0.1} & \bbf{0.1±0.1} & \uls{0.0±0.2} & -2.6±1.8 & -3.7±2.0 & -7.3±2.7 \\
TWP & 0.9±0.3 & 0.8±0.1 & 0.3±0.2 & -12.5±3.4 & -16.2±3.0 & -24.8±3.5 \\
ERGNN & 1.6±0.4 & 0.7±0.2 & 0.4±0.2 & 34.7±7.3 & 24.9±5.9 & -5.0±5.7 \\
CGNN & \bbf{0.0±0.1} & 0.4±0.1 & \bbf{-0.1±0.2} & -5.1±1.5 & -9.0±2.4 & -15.8±2.6 \\
PackNet & 0.3±0.1 & \uls{0.3±0.1} & 0.3±0.2 & N/A & N/A & N/A \\
Piggyback & 1.4±0.3 & 1.0±0.1 & 0.3±0.2 & N/A & N/A & N/A \\
HAT & 0.8±0.1 & 0.5±0.2 & 0.1±0.2 & N/A & N/A & N/A \\

\bottomrule
\end{tabular}
}
\caption{Intransigence (INT, the lower, the better)}
\end{subtable}
\end{table*}

Under Task-IL, most graph CL methods perform better as the number of tasks increases in terms of all $3$ metrics, due to the decrease in the number of tasks considered in each task.
On the contrary, under Class-IL, the performance tends to degrade in terms of AP and AF as the number of tasks increases since the distribution shift, which each model needs to adapt to, occurs more frequently. 
Interestingly, we find that INT tends to change positively as the number of tasks increases because INT does not consider the forgetting of previous tasks.

\clearpage

\section{Efficiency of \ourfw}
\label{sec:app:efficiency}
To evaluate the efficiency of the BeGin framework and the scalability of implemented algorithms, we selected three scenarios for each problem: \products (NC, Class-IL), \collab (LP, Time-IL), and \ppa (GC, Domain-IL). Note that in these scenarios the largest datasets for each problem were used. For scalability measurement in NC and LP, the original graph was scaled up (by replication) at $1\times$, $2\times$, and $4\times$ scales to create small, medium, and large datasets
In GC, the number of graphs was scaled up (by replication) at $1\times$, $2\times$, and $4\times$ scales to create small, medium, and large datasets For CL methods, we employed Bare, MAS (regularization-based method), GEM (replay-based method), and PI-GNN (parameter-isolation-based method), as representative methods for each category.
We separately measured the overhead of the BeGin framework,\footnote{It includes the execution time for loading the target scenario and the datasets for each task and handling event functions (e.g., \texttt{processBeforeTraining()} and \texttt{processAfterTraining()}), except for \texttt{processTrainIteration()} and \texttt{processEvalIteration()} for training and evaluation.} in addition to the time required for training and evaluation with the number of epochs for each task fixed to 100.

The results are presented in Table~\ref{tab:efficiency}, and from the results, we can draw the following conclusions:
\begin{itemize}
    \item \textbf{Scalability to Large Graphs:} \ourfw successfully supported training and evaluation on graphs with up to hundreds of millions of edges.
    \item \textbf{Low Overhead of Begin Framework:} In all experiments, the overhead introduced by \ourfw was very small, accounting for no more than 2.83\%. 
    \item \textbf{Linear Scalability of Algorithms:} The execution time of most methods increased at a relatively slow rate as dataset size grows, except for GEM on the \collab (LP, Time-IL) scenario. This exception may be due to the time complexity of GEM's iterations, which involve solving quadratic programming problems using external libraries (e.g., cvxpy), where convergence times and the inherent computational complexity of the solver can prevent linear scalability.
\end{itemize}

\begin{table}[h]
    \caption{\textbf{Comparison of execution times using datasets of varying sizes}}
    \setlength{\tabcolsep}{4pt}
    \centering
    \scalebox{0.7}{
    \begin{tabular}{c|c|rlc|rlc}
    \toprule[1pt]
     \multicolumn{2}{c|}{\textbf{Methods}} & \multicolumn{3}{c|}{\textbf{Bare}} & \multicolumn{3}{c}{\textbf{MAS}} \\ \midrule
     \multirow{2}{*}{Scenario} & \multirow{2}{*}{\makecell{Size of \\ Dataset}} & \multicolumn{2}{c}{Execution Time} & Overhead from \ourfw & \multicolumn{2}{c}{Execution Time} & Overhead from \ourfw \\
     & & \multicolumn{2}{c}{(seconds)} & (seconds) & \multicolumn{2}{c}{(seconds)} & (seconds) \\ \midrule
    \multirow{3}{*}{\makecell{\products \\ (NC, Class-IL)}} & Small (1$\times$) & 2850s & \multirow{2}{*}{(1.98$\times$)} & 15.9s (0.56\%) & 2831s & \multirow{2}{*}{(1.99$\times$)} & 32.7s (1.15\%) \\
     & Medium (2$\times$) & 5638s & \multirow{2}{*}{(1.97$\times$)} & 27.6s (0.49\%) & 5628s & \multirow{2}{*}{(2.06$\times$)} & 61.5s (1.09\%) \\
     & Large (4$\times$) & 11112s & & 53.5s (0.48\%) & 11614s & & 125.9s (1.08\%) \\ \midrule
    \multirow{3}{*}{\makecell{\collab \\ (LP, Time-IL)}} & Small (1$\times$) & 4359s & \multirow{2}{*}{(1.47$\times$)} & 89.2s (2.05\%) & 4414s & \multirow{2}{*}{(1.50$\times$)} & 125.0s (2.83\%) \\
     & Medium (2$\times$) & 6393s & \multirow{2}{*}{(1.85$\times$)} & 134.1s (2.10\%) & 6625s & \multirow{2}{*}{(1.91$\times$)} & 187.7s (2.83\%) \\
     & Large (4$\times$) & 11849s & & 234.9s (1.98\%) & 12629s & & 333.0s (2.64\%) \\ \midrule
     \multirow{3}{*}{\makecell{\ppa \\ (GC, Domain-IL)}} & Small (1$\times$) & 10582s & \multirow{2}{*}{(1.21$\times$)} & 24.0s (0.23\%) & 10889s & \multirow{2}{*}{(1.20$\times$)} & 81.5s (0.74\%) \\
     & Medium (2$\times$) & 12824s & \multirow{2}{*}{(1.29$\times$)} & 26.2s (0.20\%) & 13019s & \multirow{2}{*}{(1.36$\times$)} & 98.2s (0.75\%) \\
     & Large (4$\times$) & 16490s & & 29.8s (0.18\%) & 17702s & & 132.0s (0.75\%) \\
    \bottomrule[1pt]
    \end{tabular}
    }
    \scalebox{0.7}{
    \begin{tabular}{c|c|rlc|rlc}
    \toprule[1pt]
     \multicolumn{2}{c|}{\textbf{Methods}} & \multicolumn{3}{c|}{\textbf{GEM}} & \multicolumn{3}{c}{\textbf{PI-GNN}} \\ \midrule
     \multirow{2}{*}{Scenario} & \multirow{2}{*}{\makecell{Size of \\ Dataset}} & \multicolumn{2}{c}{Execution Time} & Overhead from \ourfw & \multicolumn{2}{c}{Execution Time} & Overhead from \ourfw \\
     & & \multicolumn{2}{c}{(seconds)} & (seconds) & \multicolumn{2}{c}{(seconds)} & (seconds) \\ \midrule
    \multirow{3}{*}{\makecell{\products \\ (NC, Class-IL)}} & Small (1$\times$) & 7965s & \multirow{2}{*}{(1.95$\times$)} & 15.5s (0.20\%) & 16866s & \multirow{2}{*}{(1.94$\times$)} & 46.5s (0.16\%) \\
     & Medium (2$\times$) & 15501s & \multirow{2}{*}{(2.08$\times$)} & 27.3s (0.18\%) & 32692s & \multirow{2}{*}{(1.95$\times$)} & 66.2s (0.20\%) \\
     & Large (4$\times$) & 32217s & & 51.5s (0.16\%) & 63662s & & 110.6s (0.17\%) \\ \midrule
    \multirow{3}{*}{\makecell{\collab \\ (LP, Time-IL)}} & Small (1$\times$) & 22503s & \multirow{2}{*}{(2.77$\times$)} & 87.7s (0.39\%) & \multicolumn{3}{c}{\multirow{3}{*}{N/A}}\\
     & Medium (2$\times$) & 62302s & \multirow{2}{*}{(2.80$\times$)} & 135.9s (0.22\%) \\
     & Large (4$\times$) & 174730s & & 233.9s (0.13\%) \\ \midrule
    \multirow{3}{*}{\makecell{\ppa \\ (GC, Domain-IL)}} & Small (1$\times$) & 15827s & \multirow{2}{*}{(1.20$\times$)} & 24.3s (0.15\%) & \multicolumn{3}{c}{\multirow{3}{*}{N/A}}\\
     & Medium (2$\times$) & 18953s & \multirow{2}{*}{(1.71$\times$)} & 27.5s (0.15\%) \\
     & Large (4$\times$) & 32334s & & 29.6s (0.09\%) \\
     \bottomrule[1pt]
    \end{tabular}
    }
    \label{tab:efficiency}
\end{table}

\section{Application of self-supervised-learning to graph continual learning}
\label{app:ssl}
\ourfw supports the application of self-supervised learning (SSL) methods to graph CL methods.
Specifically, it provides implementations of four widely-used SSL methods: Deep Graph Infomax~\cite{velivckovic2018deep}, GraphCL~\cite{you2020graph}, InfoGraph~\cite{sun2019infograph}, and LightGCL~\cite{cai2023lightgcl}.
Users also have the option to implement and apply their own custom SSL methods. Tutorials \cite{code} are available on (a) how to use the pre-implemented SSL methods in \ourfw and (b) how to implement custom SSL methods.

In Tables~\ref{tab:app:ssl_node} and \ref{tab:app:ssl_graph}, we present experimental results on the impact of applying SSL methods. We observed both performance improvements and declines, indicating that the gains from SSL methods were inconsistent. We speculate that this inconsistency stems from the fact that the SSL methods used were not specifically designed for graph CL.
These findings suggest that developing SSL methods specialized for graph CL can be a promising research direction.

\begin{table}[ht]
    \centering
    \caption{\textbf{Performance on \arxiv under the Task-IL setting in terms of Average Performance (AP, the higher, the better) and Average Forgetting (AF, the lower, the better).}}
    \scalebox{0.75}{
    \centering
    \begin{tabular}{c|c|cccccc}
        \toprule[1pt]
        &  & \multicolumn{6}{c}{\textbf{Self-supervised Learning Method}} \\ 
        \cmidrule{3-8}
        \makecell{\textbf{Scenario} \\ (Problem, IL Setting)}&  \textbf{Method} & \multicolumn{2}{c}{\textbf{None}} & \multicolumn{2}{c}{\textbf{DGI~\cite{velivckovic2018deep}}} & \multicolumn{2}{c}{\textbf{GraphCL~\cite{you2020graph}}} \\ 
        \cmidrule{3-8}
        & & AP & AF & AP & AF & AP & AF \\ \midrule
        
        \multirow{4}{*}{\makecell{\arxiv \\ (NC, Task-IL)}} & Bare & 65.0±7.6 &  32.5±8.5 & 64.7±7.1 & 33.2±7.8 & 64.9±6.4 & 32.9±7.1 \\
         & MAS & 91.8±0.8 & 0.6±0.4 & 91.5±0.8 & 0.0±0.1 & 91.5±0.8 & 0.1±0.1 \\
         & GEM & 90.6±0.7 & 3.6±0.5 & 90.5±1.0 & 3.6±0.6  & 90.9±1.2 & 3.3±0.7 \\
         & Piggyback & 92.6±0.7 & 0.0±0.0 & 92.8±0.7 & 0.0±0.0 & 92.8±0.7 & 0.0±0.0 \\
        \bottomrule[1pt]
    \end{tabular}
    }
    \label{tab:app:ssl_node}
\end{table}

\begin{table}[ht]
    \centering
    \caption{\textbf{Performance on \aroma under the Class-IL setting in terms of Average Performance (AP, the higher, the better) and Average Forgetting (AF, the lower, the better).}}
    \scalebox{0.75}{
    \centering
    \begin{tabular}{c|c|cccccc}
        \toprule[1pt]
        &  & \multicolumn{6}{c}{\textbf{Self-supervised Learning Method}} \\ 
        \cmidrule{3-8}
        \makecell{\textbf{Scenario} \\ (Problem, IL Setting)}&  \textbf{Method} & \multicolumn{2}{c}{\textbf{None}} & \multicolumn{2}{c}{\textbf{InfoGraph~\cite{sun2019infograph}}} & \multicolumn{2}{c}{\textbf{GraphCL~\cite{you2020graph}}} \\ 
        \cmidrule{3-8}
        & & AP & AF & AP & AF & AP & AF \\ \midrule
        \multirow{3}{*}{\makecell{\aromaticity \\ (GC, Class-IL)}} & Bare & 6.2±1.4 & 69.4±2.6 & 6.3±1.4 & 71.3±3.7 & 6.5±1.3 & 71.4±2.8 \\
         & MAS & 5.8±1.1 & 64.7±4.0 & 5.5±0.9 & 62.1±3.7 & 6.0±0.8 & 64.8±5.4 \\
         & GEM & 6.5±1.3 & 33.7±9.6 & 6.4±1.1 & 69.3±3.4 & 6.5±1.2 & 69.4±3.9 \\
        \bottomrule[1pt]
    \end{tabular}
    }
    \label{tab:app:ssl_graph}
\end{table}

\begin{table}[ht]
    \centering
    \caption{\textbf{Performance on \gowalla under the Time-IL setting in terms of Average Performance (AP, the higher, the better) and Average Forgetting (AF, the lower, the better).}}
    \scalebox{0.75}{
    \centering
    \begin{tabular}{c|c|cccccccc}
        \toprule[1pt]
        &  & \multicolumn{8}{c}{\textbf{Self-supervised Learning Method}} \\ 
        \cmidrule{3-10}
        \makecell{\textbf{Scenario} \\ (Problem, IL Setting)}&  \textbf{Method} & \multicolumn{2}{c}{\textbf{None}} & \multicolumn{2}{c}{\textbf{DGI~\cite{velivckovic2018deep}}} & \multicolumn{2}{c}{\textbf{GraphCL~\cite{you2020graph}}} & \multicolumn{2}{c}{\textbf{LightGCL~\cite{cai2023lightgcl}}} \\ 
        \cmidrule{3-10}
        & & AP & AF & AP & AF & AP & AF & AP & AF \\ \midrule
        \multirow{3}{*}{\makecell{\gowalla \\ (LC, Time-IL)}} & Bare & 9.4±0.5 & 0.5±0.8 & 9.5±0.4 & 0.9±0.6 & 9.2±0.5 & 1.6±0.6 & 9.4±0.6 & 2.4±0.7 \\
         & MAS & 8.8±0.5 & 0.4±0.6 & 9.0±0.5 & 0.8±0.5 & 9.1±0.3 & 0.6±0.4 & 7.7±1.2 & 1.5±0.8 \\
         & GEM & 8.9±0.6 & 1.0±0.9 & 8.7±0.6 & 1.9±0.6 & 8.5±0.5 & 1.8±0.5 & 8.5±0.7 & 1.5±0.8 \\
        \bottomrule[1pt]
    \end{tabular}
    }
    \label{tab:app:ssl_link}
\end{table}

\section{Parameter-Isolation-Based Methods}
\label{sec:app:pim}
We extend parameter-isolation-based methods (spec., PackNet, Piggyback, and HAT), which have been used for independent data (e.g., images and text), to graph-related problems (spec., NC, LC, and GC).
The methods learn binary or real-valued masks for weighting parameters or model outputs for each task.
Note that, since they require knowing which task each query is on, they are applicable only to Task-IL.
Below, we describe how we apply the methods to graph CL.

\smallsection{PackNet~\citep{mallya2018packnet}.} It learns binary masks for network parameters as follows: 
\begin{enumerate}
    \item For the current task, pre-train all parameters that have been unmasked so far.
    \item Mask the set of pre-trained parameters whose magnitude is in top $(1-p) \times 100$\%, where $p$ is the pruning ratio.
    \item Re-initialize the unmasked ones to $0$ and re-train only the newly masked parameters for the current task.
    \item After the re-training, fix the masked parameters to their current values to prevent forgetting \item Repeat the above processes for the next task.
\end{enumerate}
We apply binary masks to both the backbone model (e.g., GCN) and fully-connected layer(s).

\smallsection{Piggyback~\citep{mallya2018piggyback}.} It requires a pre-trained network, which is fixed over tasks, and learns real-valued masks for each task.
For all node-level problems, we used Deep Graph Infomax~\citep{velivckovic2018deep} to pre-train the backbone graph neural network (GNN) layers (e.g., GCN).
For all link-level problems, we applied Deep Graph Infomax to pre-train the GCN layers and the 3-layer MLP. For the graph-level problems, we used InfoGraph~\citep{sun2019infograph} to pre-train the GCN layers and the 3-layer MLP.
Note that any unsupervised training method for graphs can be used instead for pre-training.

In the training phase, Piggyback learns real-valued masks, which are multiplied with the pre-trained  parameters, end-to-end for each task. 
In the test phase, it uses a binarizer to transform the mask entries above the threshold hyperparameter $\tau$ into $1$ and the others into $0$. Lastly, for each task, it applies the binarized masks for the  task to the pre-trained network to obtain the network to be used for task.
As in PackNet, we apply the above steps to both the backbone model (e.g., GCN) and fully-connected layer(s).

\smallsection{HAT~\citep{serra2018overcoming}.} It learns real-valued masks, which play a role similar to attention modules, for the outputs of each layer, while the above two methods apply binary masks to network parameters.
Specifically, we learn real-valued masks for each layer, and perform element-wise multiplication of them and the outputs of each layer.
For each task, we apply the above steps to both the backbone model (e.g., GCN) and fully-connected layer(s).

\begin{figure*}[t]
    \centering
    \begin{subfigure}[b]{0.242\textwidth}
         \centering
         \includegraphics[width=\textwidth]{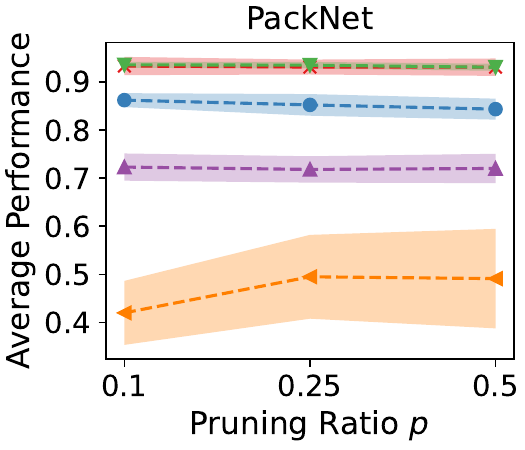}
         \caption{PackNet}
     \end{subfigure}
     \begin{subfigure}[b]{0.242\textwidth}
         \centering
         \includegraphics[width=\textwidth]{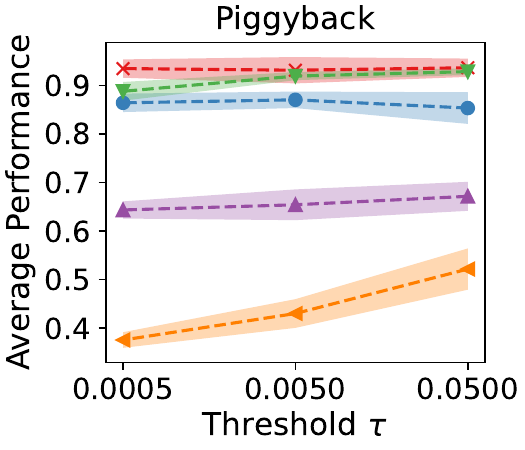}
         \caption{Piggyback}
     \end{subfigure}
     \hspace{5mm}
     \begin{subfigure}[b]{0.18\textwidth}
         \centering
         \includegraphics[width=\textwidth]{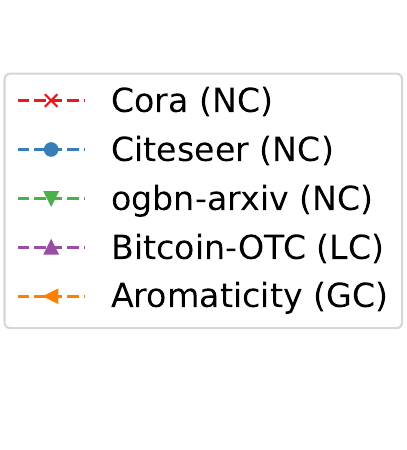}
     \end{subfigure}
     \caption{\textbf{Effect of hyperparameters of PackNet and Piggyback on Average Performance (AP).} We represent the changes in average performances with dotted lines, and the colored areas represent their standard deviations.} 
     \label{fig:sensitivity_pp}
\end{figure*}

\begin{figure*}[t]
    \centering
    \begin{subfigure}[b]{0.4\textwidth}
        \centering
        \includegraphics[width=\textwidth]{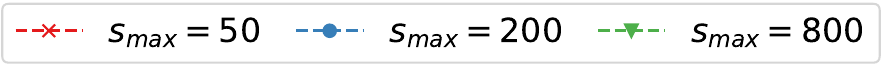}
    \end{subfigure}
    
    \begin{subfigure}[b]{0.242\textwidth}
         \centering
         \includegraphics[width=\textwidth]{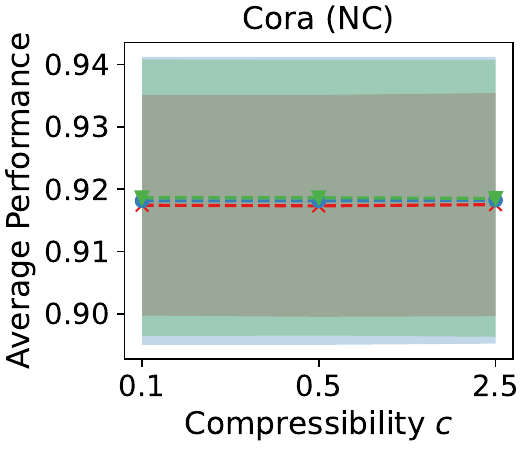}
         \caption{Cora (NC)}
     \end{subfigure}
     \begin{subfigure}[b]{0.242\textwidth}
         \centering
         \includegraphics[width=\textwidth]{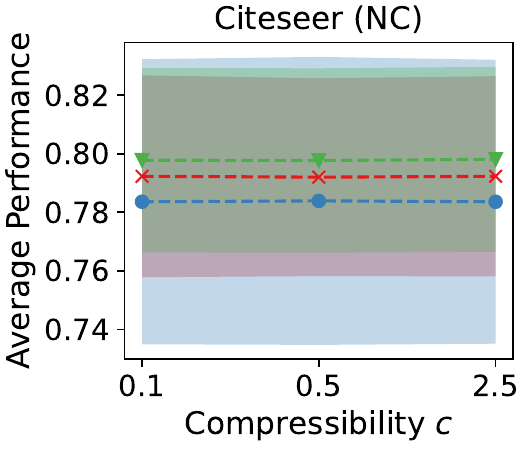}
         \caption{Citeseer (NC)}
     \end{subfigure}
     \begin{subfigure}[b]{0.242\textwidth}
         \centering
         \includegraphics[width=\textwidth]{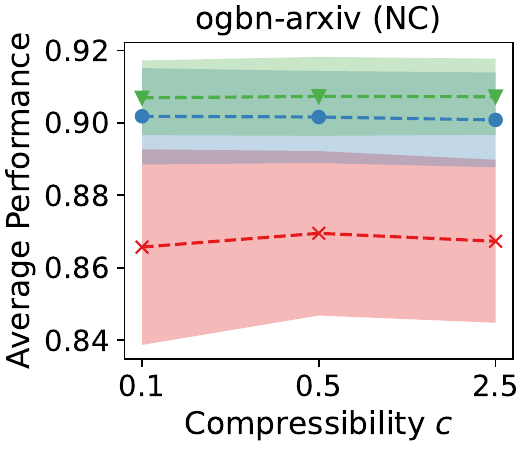}
         \caption{ogbn-arxiv (NC)}
     \end{subfigure}
     
     \begin{subfigure}[b]{0.242\textwidth}
         \centering
         \includegraphics[width=\textwidth]{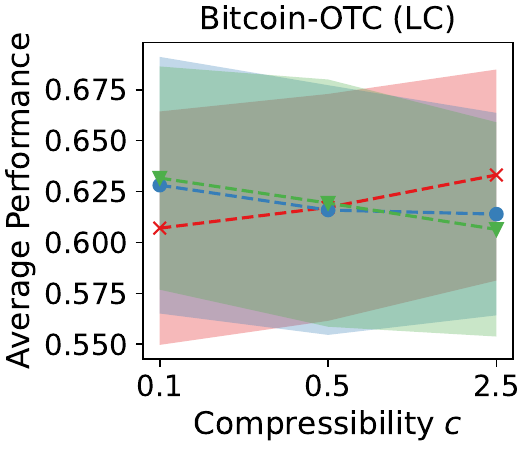}
         \caption{Bitcoin-OTC (LC)}
     \end{subfigure}
     \begin{subfigure}[b]{0.242\textwidth}
         \centering
         \includegraphics[width=\textwidth]{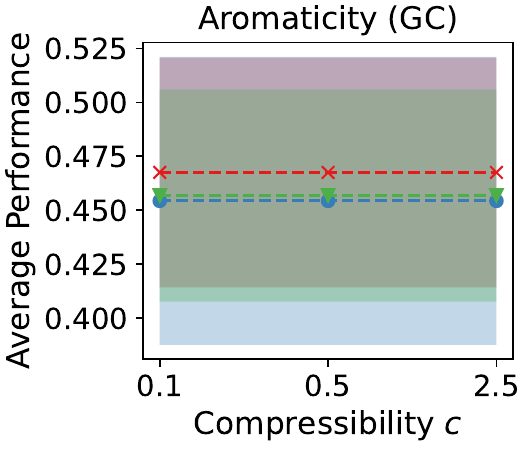}
         \caption{Aromaticity (GC)}
     \end{subfigure}
     \caption{\textbf{Effect of hyperparameters of HAT on Average Performance (AP).} The dotted lines represent the means, and the colored regions represent the standard deviations.}
     \label{fig:sensitivity_hat}
\end{figure*}

\smallsection{Effects of Hyperparameters.}
We performed extra experiments to investigate the effects of their hyperparameters on the performance of the parameter-isolation-based methods. 
For PackNet, we examined the effect of the pruning ratio $p$ while varying it among $\{0.1, 0.25, 0.5\}$. 
For piggyback, we examined the effect of the threshold $\tau$  while varying it among $\{\text{5e-2}, \text{5e-3}, \text{5e-4}\}$.
For HAT, we checked the effect of the stability parameter $s_{max}$ and the compressibility parameter $c$ while varying them among $\{50, 200, 800\}$ and $\{0.1, 0.5, 2.5\}$, respectively. 

In Figures~\ref{fig:sensitivity_pp}-\ref{fig:sensitivity_hat}, we report the AP scores for NC on \cora, \citeseer, and \arxiv; LC on \bitcoin, and GC on \aroma.
In Figure~\ref{fig:sensitivity_pp}, we show the effect of the hyperparameters of PackNet and Piggyback. 
In general, PackNet was not sensitive to the choice of pruning ratio $p$.
Specifically, the effect of the changes in the pruning ratio was significant only on \aroma, where increasing the pruning ratio from $0.1$ to $0.25$ greatly increased the accuracy.
Regarding Piggyback, its accuracy changed gradually with respect to the threshold $\tau$, and the tendencies vary with datasets.
In Figure~\ref{fig:sensitivity_hat}, we report the results regarding HAT.
Its accuracy depended heavily on the stability $s_{max}$, especially on \citeseer, \arxiv, and \aroma, while HAT was relatively insensitive to the choice of the compressibility parameter $c$.
The compressibility parameter $c$ significantly affected the accuracy of HAT only on \bitcoin.

\section{Detailed Comparison with Existing Studies on Graph CL}
\label{sec:app:detilaed_compare}

As summarized in Table \ref{tab:app:compare_datasets}, we newly used $15$ datasets that have not been used in the previous graph CL literature. Even for previously considered datasets, in most cases, we made better use of them under more incremental settings. 
For example, we used the \magdata dataset for node classification under Time-IL settings and used the \ppa dataset for graph classification under Domain-IL settings. 
Furthermore, commonly utilized datasets, such as \cora, \citeseer, \arxiv, \corafull, and \reddit, are mostly obtained from bibliographic data, and 
only limited combinations of the problem levels and the incremental settings (spec., NC for Task-/Class-/Time-IL) have been considered.
\ourfw provides $\numscean$ benchmark scenarios with the real-world graph datasets obtained from $11$ application domains (protein association, user-item interaction, citation, molecule, co-purchase, hyperlink, trading, co-authorship, image, traffic, and language) and covers all combinations of the problem levels and the incremental settings.

\begin{table}[h!]
\centering
\caption{\textbf{Comparison of datasets and incremental settings considered in each study.} ‘T’, ‘C’, ‘D’, and ‘Ti’ denote task-, class-, domain-, and time-incremental settings, respectively.
An asterisk (*) indicates a dataset not used in this study. A double asterisk (**) indicates that the incremental tasks were constructed without actual time information, due to the absence of the information.}
\label{tab:app:compare_datasets}
    \scalebox{0.7}{
        \setlength{\tabcolsep}{1pt}
        \renewcommand{\arraystretch}{1.3}
        \begin{tabular}{c|c|c|c|c|c|c|c|c|c|c|c|c|c|c|c|c}
            \toprule[1pt]
            Datasets & \cite{liu2021overcoming} & \cite{carta2021catastrophic} & \cite{galke2021lifelong} & \cite{cai2022multimodal} & \cite{sun2023self} & \cite{zhang2022hierarchical} & \cite{zhou2021overcoming} & \cite{wang2022lifelong} & \cite{zhang2022cglb} & \cite{kim2022dygrain} & \cite{tan2022graph} & \cite{rakaraddi2022reinforced} & \cite{han2020graph} & \cite{liu2023cat} & \cite{zhang2023continual} & Ours \\  \midrule \hline
            \multicolumn{17}{c}{Datasets Used for Node-level Problems} \\ \hline
            \makecell\cora &  &  &  &  & T & C & T & C & & & & T, C & & & Ti** & T, C\\ \hline
            \citeseer &  &  &  &  & T & C & T & C & & & & T, C &  & & Ti** & T, C\\ \hline
            \makecell{\actor* \cite{pei2020geom}} &  &  &  &  & T & C &  &  & & & & &  & & & \\ \hline
            \corafull & T &  &  &  &  &  &  &  & T, C & & & T, C &  & T, C& & T\\ \hline
            \makecell{\pubmed* \cite{sen2008collective}} &  &  &  &  &  &  &  & C & & Ti** & & &  & & & \\ \hline
            \makecell{\dblp* \cite{tang2008arnetminer}} &  &  & Ti &  &  &  &  &  & & & C & &  & & Ti & \\ \hline
            \arxiv &  &  &  &  & T & C &  & C & T, C & Ti & & & & T, C& Ti & T, C, Ti\\ \hline
            \makecell{\reddit \cite{hamilton2017inductive}} & T &  &  &  & T &  & T &  & T, C & Ti & C & &  & T, C & & \\ \hline
            \makecell{\pharmabio* \cite{melnychuk2019effects}} &  &  & Ti &  &  &  &  &  & & & & &  & & & \\ \hline
            \proteins &  &  &  &  &  &  &  &  & & & & &  & & & D\\ \hline
            \makecell{\articles* \cite{cai2022multimodal}} &  &  &  & T &  &  &  &  & & & & &  & & & \\ \hline
            \magdata &  &  &  &  &  &  &  &  & & & & &  & & & T, C, Ti\\ \hline
            \makecell{\ppi* \cite{zitnik2017predicting}} & T &  &  &  &  &  &  &  & & & & &  & & & \\ \hline
            \products &  &  &  &  &  & C &  &  & T, C & Ti** & & &  & T, C & & C\\ \hline
            \twitch &  &  &  &  &  &  &  &  & & & & &  & & & D \\ \hline
            \makecell{\amazon* \cite{he2016ups}} &  &  &  & T &  &  &  &  & & & & &  & & & \\ \hline
            \makecell{\amazoncomputer* \cite{mcauley2015image}} & T &  &  &  &  &  &  &  & & & & T, C &  & & Ti** & \\ \hline
            \makecell{\amazonclothing* \cite{mcauley2015inferring}} & &  &  &  &  &  &  &  & & & C & &  & & & \\ \hline
            \amazonphoto*\cite{shchur2018pitfalls} &  &  &  &  &  &  &  &  & & & & &  & & Ti** &  \\ \hline
            \elliptic*\cite{weber2019anti} &  &  &  &  &  &  &  &  & & & & &  & & Ti &  \\ \hline
            \paperm*\cite{wang2020microsoft} &  &  &  &  &  &  &  &  & & & & &  & & Ti &  \\ \hline
            
            \multicolumn{17}{c}{Datasets Used for Link-level Problems} \\ \hline
            \wikics &  &  &  &  &  &  &  &  & & & & & & & & D\\ \hline
            \bitcoin &  &  &  &  &  &  &  &  & & & & &  & & & T, C, Ti\\ \hline
            \collab &  &  &  &  &  &  &  &  & & & & & & & & Ti\\ \hline
            \facebook &  &  &  &  &  &  &  &  & & & & & & & & D\\ \hline
            \askubuntu &  &  &  &  &  &  &  &  & & & & & & & & Ti\\ \hline
            \gowalla &  &  &  &  &  &  &  &  & & & & & & & & Ti\\ \hline
            \movie &  &  &  &  &  &  &  &  & & & & & & & & Ti\\ \hline
            
            \multicolumn{17}{c}{Datasets Used for Graph-level Problems} \\ \hline
            \mnist &  & C &  &  &  &  &  &  & & & & & & & & T, C\\ \hline
            \cifar &  & C &  &  &  &  &  &  & & & & & & & & T, C\\ \hline
            \aromaticity &  &  &  &  &  &  &  &  & T, C & & & & & & & T, C\\ \hline
            \molhiv &  &  &  &  &  &  &  &  &  & & & & & & & D\\ \hline
            \ppa &  & C &  &  &  &  &  &  &  & & & & & & & D\\ \hline
            \makecell{\sider* \cite{wu2018moleculenet}} &  &  &  &  &  &  &  &  & T & & & & & & & \\ \hline
            \nyctaxi &  &  &  &  &  &  &  &  &  & & & & & & & Ti\\ \hline
            \sentiment &  &  &  &  &  &  &  &  &  & & & & & & & Ti\\ \hline
            \makecell{\tox* \cite{mayr2016deeptox,huang2016tox21challenge}} & T &  &  &  &  &  &  &  & T & & & & & & & \\ \hline 
            \makecell{\politifact* \cite{shu2020fakenewsnet}} &  &  &  &  &  &  &  &  &  & & & & D & & & \\ \hline
            \makecell{\gossipcop* \cite{shu2020fakenewsnet}} &  &  &  &  &  &  &  &  &  & & & & D & & & \\ \hline
            \zinc &  &  &  &  &  &  &  &  &  & & & & & & & D\\ \hline
            \aqsol &  &  &  &  &  &  &  &  &  & & & & & & & D\\
            \hline \hline
            \textbf{Total \# of Settings} & \textbf{5} & \textbf{3} & \textbf{2} & \textbf{2} & \textbf{5} & \textbf{5} & \textbf{3} & \textbf{4} & \textbf{12} & \textbf{4} & \textbf{3} & \textbf{8} & \textbf{2} & \textbf{8} & \textbf{8} & \textbf{\numscean}  \\
            \bottomrule[1pt]
        \end{tabular}
    }
\end{table}

\section{License of Assets}
\label{sec:app:detailsall}

\smallsection{Implementation.}
All assets of \method are available under the Apache license 2.0. All assets we utilized from DGL \citep{wang2019deep} and DGL-LifeSci \citep{li2021dgl} are also available under the Apache license 2.0. For the implementation of GEM, TWP, and ERGNN, we utilized the implementation from CGLB \citep{zhang2022cglb}, which is available under CC BY-NC 4.0. For the implementation of ContinualGNN, CaT, and PI-GNN, we utilized the implementation written by the author, whose license is not specified.

\smallsection{Benchmark Datasets.} The detailed license information for the considered datasets are as follows:
\begin{itemize}[leftmargin=*]
    \item The \cora, \citeseer, \corafull, \bitcoin, and \wikics datasets are publicly available.
     \wikics is under the MIT license, but the licenses for these datasets are not specified. For accessing the datasets, we used DGL \citep{wang2019deep}.
    \item The \arxiv, \magdata, and \collab datasets are publicly available under the ODC-BY license. We used OGB \citep{hu2020open} to access the datasets.
    \item The \products dataset is publicly available under the Amazon license. We used OGB \citep{hu2020open} to access the datasets.
    \item The \proteins, \molhiv, and \ppa datasets are publicly available under the CC-0 license and the MIT license, respectively. We used OGB \citep{hu2020open} to access the datasets.
    \item The \twitch and \facebook datasets are publicly available under the MIT license. The \askubuntu dataset is also publicly available, but the license is not specified. These three datasets are available at \url{http://snap.stanford.edu/data/index.html}.
    \item The \mnist and \cifar datasets are publicly available under the MIT license. We used PyG \citep{fey2019fast} to access the datasets.
    \item The \aroma dataset, which is a subset of the PubChem BioAssay dataset \citep{xiong2019pushing}, is publicly available under the CC BY-NC 3.0 license. We used DGL-LifeSci \citep{li2021dgl} to access the dataset.
    \item The raw data of \nyctaxi is publicly available, and the license is not specified. We transformed the Yellow Taxi Trip data collected from 2018 to 2021 into graphs, as described in Section \ref{sec:scenarios:examples}. The graphs are available in our framework, and some basic statistics are provided in Table \ref{tab:datasets}.
    \item The raw data of \sentiment is publicly available, and the license is not specified. We transformed the online posts with sentiment into graphs, as described in Section \ref{sec:scenarios:examples}. The graphs are available in our framework, and some basic statistics are provided in Table \ref{tab:datasets}.
    \item The \gowalla dataset is publicly available under the MIT license. The dataset is available at \url{https://github.com/xiangwang1223/neural_graph_collaborative_filtering/tree/master/Data/gowalla}.
    \item The raw data of \movie is publicly available at \url{https://grouplens.org/datasets/movielens/}, and the license is specified at \url{https://files.grouplens.org/datasets/movielens/ml-1m-README.txt}. Following the license, we do not redistribute the dataset but provide the implementation for processing the dataset (recall that we formed a bipartite graph between users and items, as described in Section \ref{sec:scenarios:examples}).
    \item The \zinc and \aqsol datasets are publicly available under the MIT license. These two datasets are available at \url{https://github.com/graphdeeplearning/benchmarking-gnns}.

\end{itemize}

\end{appendices}

\end{document}